%% file: main.tex
\documentclass[a4paper,fleqn]{cas-sc}

\usepackage[authoryear]{natbib}
\usepackage{url}
\usepackage{hyperref}
\usepackage{setspace} 
\usepackage{soul}

\usepackage{booktabs, multirow} 
\usepackage{soul}
\usepackage{changepage,threeparttable} 

\usepackage{physics}
\usepackage{amsmath}
\usepackage{tikz}
\usepackage{mathdots}
\usepackage{yhmath}
\usepackage{cancel}
\usepackage{color}
\usepackage{siunitx}
\usepackage{array}
\usepackage{amssymb}
\usepackage{gensymb}
\usepackage{tabularx}
\usepackage{extarrows}
\usetikzlibrary{fadings}
\usetikzlibrary{patterns}
\usetikzlibrary{shadows.blur}
\usetikzlibrary{shapes}

\def\Equal{\texttt{=}}

\begin{document}
\let\WriteBookmarks\relax
\def\floatpagepagefraction{1}
\def\textpagefraction{.001}

\shorttitle{Improving Chest X-Ray Report Generation by Leveraging Warm Starting}

\shortauthors{Aaron Nicolson et~al.}

\title [mode = title]{Improving Chest X-Ray Report Generation by Leveraging Warm Starting}                      
\author[1]{Aaron Nicolson}[
    type=editor, auid=000,bioid=1,orcid=0000-0002-7163-1809
    ]
\cormark[1]
\ead{aaron.nicolson@csiro.au}

\affiliation[1]{
organization={The Australian e-Health Research Centre, CSIRO Health and Biosecurity, Brisbane, Australia},
    }

\author[1]{Jason Dowling}[]

\author[1]{Bevan Koopman}[]


\cortext[cor1]{Corresponding author}

\begin{abstract}
Automatically generating a report from a patient's Chest X-Rays (CXRs) is a promising solution to reducing clinical workload and improving patient care. However, current CXR report generators---which are predominantly encoder-to-decoder models---lack the diagnostic accuracy to be deployed in a clinical setting. To improve CXR report generation, we investigate warm starting the encoder and decoder with recent open-source computer vision and natural language processing checkpoints, such as the Vision Transformer (ViT) and PubMedBERT. To this end, each checkpoint is evaluated on the MIMIC-CXR and IU X-Ray datasets. Our experimental investigation demonstrates that the Convolutional vision Transformer (CvT) ImageNet-21K and the Distilled Generative Pre-trained Transformer 2 (DistilGPT2) checkpoints are best for warm starting the encoder and decoder, respectively. Compared to the state-of-the-art ($\mathcal{M}^2$ Transformer Progressive), CvT2DistilGPT2 attained an improvement of 8.3\% for CE F-1, 1.8\% for BLEU-4, 1.6\% for ROUGE-L, and 1.0\% for METEOR. The reports generated by CvT2DistilGPT2 have a higher similarity to radiologist reports than previous approaches. This indicates that leveraging warm starting improves CXR report generation. Code and checkpoints for CvT2DistilGPT2 are available at \url{https://github.com/aehrc/cvt2distilgpt2}.
\end{abstract}

\begin{keywords}
Chest X-ray report generation \sep Image captioning \sep Multi-modal learning \sep warm starting
\end{keywords}

\maketitle
\doublespacing
\sloppy

\section{Introduction}

\begin{figure*}
    \centering
    \includegraphics[scale=1.0]{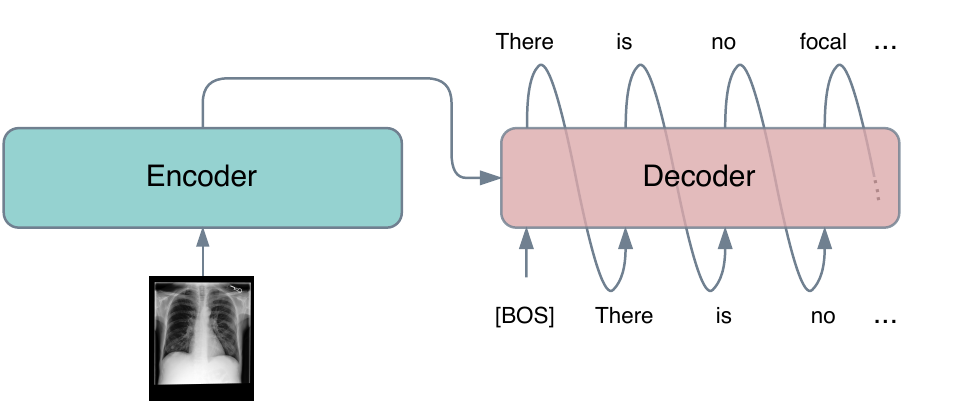}
    \caption{\label{fig:encoder_to_decoder} Encoder-to-decoder model for CXR report generation. The encoder extracts visual features from the CXR. Following this, the decoder autoregressively generates each subword of the radiology report based on the previously generated subwords and the visual features from the encoder. The example is DICOM \texttt{02aa804e-bde0afdd-112c0b34-7bc16630-4e384014} from study \texttt{50414267} of patient \texttt{10000032} from MIMIC-CXR. \texttt{[BOS]} is the beginning-of-sentence special token. }
\end{figure*}

Chest X-Ray (CXR) report generation is the task of automatically generating a radiology report from a given patient's CXR. It has the potential to improve radiologist workflows, reduce the burden of radiology reporting, and improve patient outcomes \citep{Thrall_2018}. The most popular method of CXR report generation is with a deep learning model, specifically, an encoder-to-decoder model as shown in Figure \ref{fig:encoder_to_decoder} \citep{Pavlopoulos_2021}. First, the encoder extracts visual features from a given CXR. Next, the decoder autoregressively generates each word (or subword) of the radiology report based on the previously generated words and the visual features. While current CXR report generation methods are promising, a significant improvement in diagnostic accuracy is required before clinical consideration. One cause is that the publicly available datasets used to develop CXR report generators (e.g., MIMIC-CXR \citep{Johnson_2019} and IU X-Ray \citep{DemnerFushman_2015}) are relatively smaller and of lesser quality than general-domain image captioning datasets \citep{chen2015microsoft}. For example, CXR report generators perform poorly on  underrepresented abnormalities within MIMIC-CXR \citep{Liu_2019}.


A potential solution that has not been thoroughly investigated for CXR report generation is warm starting. Warm starting refers to the initialisation of a models parameters with those from a pre-trained model, that is, an identical model that has been trained on some task that is related to the task of concern (where the related task is otherwise known as a pre-training task). By warm starting a model, the transfer of knowledge from the pre-training task to the task of concern can provide a significant performance boost. Warm starting is particularly effective when the dataset for the task of concern is lacking in size or quality, the domain of the pre-training task is similar to the task of concern, and the pre-training dataset is of considerable size or quality \citep{9134370}.  

When warm starting, the collection of learned parameters of a pre-trained model from a specific point during its training is used, otherwise known as a checkpoint. Checkpoints are saved at set intervals during model training, and the checkpoint that is selected for further use (in this case, warm starting) is usually the one that attains the best score for a chosen metric. For an encoder-to-decoder model, either the encoder, decoder, or both can be warm started with different encoder-only or decoder-only checkpoints. This gives rise to a vast number of different encoder and decoder checkpoint combinations that can be considered. 

There are many publicly available checkpoints that can be used for warm starting, with two main types frequently available. There are checkpoints that have been trained on a large amount of general-domain data. There are also those trained in a specific domain on a smaller amount of data than their general-domain counterparts, yet larger than the data available for the task of concern. For example, BERT is a general-domain checkpoint trained on English Wikipedia and BookCorpus \citep{Devlin_2019}, and ClinicalBERT is a domain-specific checkpoint whose last stage of pre-training is on MIMIC-III \citep{Alsentzer_2019}. Together, English Wikipedia and BookCorpus form a significantly larger dataset than MIMIC-III, yet the domain of the Intensive Care Unit (ICU) Electronic Health Records (EHRs) from MIMIC-III is closer to radiology reports. This makes it difficult to discern which compromise is appropriate when selecting a checkpoint for a specific task such as CXR report generation.

In this work, we study the impact of warm starting on CXR report generation with the aim of generating reports that are more similar to radiologists' reports than previous approaches. This is motivated by the performance gains that can be attained through warm starting, especially when there are deficiencies with the dataset for a task \citep{9134370}. Given the deficiencies associated with MIMIC-CXR and IU X-Ray, transferring knowledge from a pre-trained model via warm starting may be beneficial. We do this by evaluating various general-domain and domain-specific checkpoints for the encoder and decoder on the MIMIC-CXR and IU X-Ray datasets (described in Section \ref{sec:dataset}) using several objective metrics (described in Section \ref{sec:metrics}). From this evaluation, we determine which checkpoints are most suitable for warm starting the encoder and decoder of a CXR report generator. Using these checkpoints, we aim to generate reports that are more similar to radiologists' reports than previous approaches in the literature. We also aim to better understand what influences the effectiveness of warm starting, such as the domain and size of the pre-training data and the model architecture of the checkpoint. This investigation also enables us to give insightful recommendations on how to further improve CXR report generation.



\section{Background} \label{sec:background}

The chest radiograph or CXR is an indispensable tool for diagnosing diseases of the cardiovascular and respiratory systems \citep{Kelly_2012}. While it is the most commonly performed radiologic examination, the CXR is difficult to interpret \citep{Kanne_2005}. Even though CXRs are analysed by clinicians of all types, radiologists demonstrate the highest diagnostic accuracy \citep{Satia_2013}. Alarmingly, the workload of radiologists has increased significantly over the last couple of decades---mostly due to increases in cross-sectional imaging and understaffing \citep{McDonald_2015, Liu_2017}. As a consequence, burnout is common amongst radiologists, with fatigue leading to a reduction in their diagnostic accuracy \citep{Harolds_2016, Krupinski_2010}. Radiologists also exhibit interobserver and intraobserver variability between their diagnoses \citep{Balabanova_2005}. Currently, radiologists communicate their findings to referring clinicians through a report. However, there are inconsistencies in reporting styles, with structured reports recommended \citep{ESR_2011}. Moreover, failing to report in a clear and concise manner---which is exacerbated by fatigue---can also lead to sub-optimal patient care \citep{Siegal_2017, Vosshenrich_2021}. 



Deep learning is an important tool for medical image analysis, with recent applications to tasks such as segmentation \citep{Wang_2022}, COVID-19 diagnosis \citep{Li_2021}, and disease classification \citep{li_novel_2023}. It has been applied to a range of modalities, for example, radiography, ultrasonography, computed tomography, magnetic resonance, and radiomic data. An emerging solution for CXR interpretation is the use of deep learning to automatically generate a report from a given CXR \citep{Pavlopoulos_2021}. CXR report generation has become the most popular medical image captioning task with more datasets than any other modality or anatomical region \citep{Ayesha_2021}. Medical image captioning is a domain-specific version of general-domain image captioning---a topic that has garnered a high amount of attention over the last couple of years \citep{Yang_2022, Ji_2021}. 

CXR report generation has the potential to improve radiologist workflows; by providing a pre-filled radiology report ready for modification, the burden of radiology reporting could be reduced. Such a system could also increase the diagnostic accuracy of clinicians who have a lower diagnostic confidence \citep{Thrall_2018, Alexander_2020}. A hosted system would also be able to serve the radiology demands of an entire health system, as it possesses the ability to remotely interpret multiple CXRs simultaneously in seconds. Moreover, the system would not suffer from intraobserver variability or fatigue---which could result in more consistent reporting \citep{Leeuwen_2021}. While current CXR report generation approaches are promising, a significant improvement in diagnostic accuracy is needed before adoption into a clinical setting \citep{Kelly_2019}.

As mentioned previously, the encoder-to-decoder model is the most popular approach to automatic CXR report generation \citep{Pavlopoulos_2021}. Visual features from the CXR are extracted with the encoder, were the visual features are a high-level representation of regions in the CXR. During generation, the decoder is conditioned on the visual features via its cross-attention modules and on the previous words of the report via its self-attention modules. The encoder is typically a Convolutional Neural Network (CNN), such as a Residual or Densely-connected Network (ResNet or DenseNet, respectively) \citep{He_2016, Huang_2017}. A CNN includes multiple layers of two-dimensional convolutional kernels that have an inductive bias towards local spatial features---a characteristic that makes CNNs suitable for Computer Vision (CV) tasks. For the decoder, a Transformer decoder is most commonly employed \citep{Vaswani_2017}. A Transformer consists of layers that have multiple attention heads, where attention produces an output based on which inputs it deems important. For Natural Language Processing (NLP), each input corresponds to a word. Along with the encoder-to-decoder model, several deep learning techniques have been investigated for CXR report generation. 

For CXR interpretation, warm starting the encoder with a general-domain CV checkpoint---typically a ResNet or DenseNet ImageNet-1K checkpoint---provides a significant performance boost over random parameter initialisation \citep{Ke_2021, Russakovsky_2015, Pavlopoulos_2021}. More recently, DistilGPT2---a general-domain NLP checkpoint---was used to warm start the decoder \citep{Alfarghaly_2021}. However, a direct comparison to a randomly initialised decoder was not provided. While warm starting the encoder has become standard practice, and warm starting the decoder is beginning to receive traction, there exists a plethora of general-domain and domain-specific CV and NLP checkpoints that have yet to be explored for CXR report generation.

General-domain CV checkpoints have received a considerable amount of attention as of late \citep{Kolesnikov_2020}. Various pre-training tasks, such as distillation and self-supervised learning, have produced checkpoints that provide a significant performance boost on general-domain tasks \citep{Bao_2021, Touvron_2021}. Along with these, several prominent CV models and their checkpoints have not been investigated for CXR report generation. ImageNet-1K checkpoints for EfficientNet are one such example; EfficientNet is a CNN that can outperform ResNets and DenseNets while consuming drastically fewer parameters \citep{Tan_2019}. Another promising checkpoint is the Vision Transformer (ViT) and its improvements \citep{Dosovitskiy_2020, Wu_2021, ElNouby_2021}. ViT is a Transformer that has been trained on large general-domain image classification datasets, where the set of words as input are replaced with a set of patches from an image. Transformers possess several benefits over CNNs, such as no pooling, a full-receptive field at each layer, and robustness to occlusion \citep{Naseer_2021}. More recent checkpoints that use distillation have outperformed EfficientNet on ImageNet-1K, namely the 
Data-efficient image Transformer (DeiT) \citep{Touvron_2021}.

While DistilGPT2 has been investigated for CXR report generation, several other prominent NLP checkpoints have not. General-domain NLP checkpoints formed from large corpora, such as BERT and GPT2, have been shown to boost performance in Natural Language Understanding (NLU) \citep{Devlin_2019} and Natural Language Generation (NLG) \citep{Radford_2019} tasks, respectively. NLU focuses on comprehending natural language through grammar and context while NLG focuses on constructing natural language based on a given input. As BERT is an NLU checkpoint, one would assume that it is less suitable than GPT2 for warm starting a decoder tasked with NLG. However, \citet[{\footnotesize BERT}{\normalsize 2}{\footnotesize BERT} vs. {\footnotesize BERT}{\normalsize 2}{\footnotesize GPT2}]{Rothe_2020} disproved this by demonstrating that BERT is more apt than GPT2 for warm starting the decoder of a sequence-to-sequence model. Following BERT, several NLU checkpoints were developed from large biomedical natural language corpora. Warm starting with a biomedical NLU checkpoint instead of BERT was shown to improve performance in biomedical NLU tasks \citep{Lee_2019}. As highlighted by \citet{Kaur_2021}, warm starting the decoder of a CXR report generator with a biomedical NLU checkpoint is a promising approach that warrants investigation. Hence, an investigation determining if NLG and NLU checkpoints can be effectively fine-tuned to model not only CXR reports but also visual features of CXRs is warranted.

\section{Related work}

\subsection{CXR report generation} \label{sec:cxr_methods}
In this section, we summarise both foundational and recent automatic CXR report generation approaches in the literature. For a more exhaustive review of CXR report generation, we refer the reader to the survey conducted by \citet{Kaur_2021}. As highlighted by \citet{Pavlopoulos_2021}, the MIMIC-CXR and IU X-Ray datasets are consistently used in the literature to evaluate CXR report generation (both are described in Subsection \ref{sec:dataset}). The following approaches are evaluated using at least one or both of these datasets.

\citet{Wang_2018} was the first to use an encoder-to-decoder model for CXR report generation---a {\small ResNet-50}{\large 2}{\small RNN} with attention, where RNN refers to a Recurrent Neural Network.\footnote{Henceforth, the naming convention for encoder-to-decoder models will be the name of the encoder, followed by {\normalsize 2}, followed by the name of the decoder. For example, a model with a CNN encoder and an RNN decoder will be named {\footnotesize CNN}{\normalsize 2}{\footnotesize RNN}.} The authors employed multi-task learning of CXR report generation (IU X-Ray) and multi-label abnormality classification (ChestX-ray14 abnormality labels). The ResNet was warm started using an ImageNet-1K checkpoint. The proposed approach outperformed a general-domain image captioning approach, where several NLG metrics were used for evaluation. However, many false negative abnormality predictions were evident in the generated reports \cite[Figure 4]{Wang_2018}. 

While NLG metrics capture the similarity between the predicted and ground-truth reports, they do not always capture diagnostic accuracy \citep{Pavlopoulos_2021}. Motivated by this, \citet{Liu_2019} proposed the Clinical Efficacy (CE) metrics. The CE metrics make use of the CheXpert labeler---a tool developed to extract 14 observations from the reports of the CheXpert dataset (12 of which are abnormalities) \citep{Irvin_2019}. For each observation, the CheXpert labeler predicts whether each observation was mentioned as positive, negative, or uncertain, or if it was not mentioned. For the CE metric, observations are first extracted from the generated and ground-truth reports using the CheXpert labeler. Following this, the precision, recall, and F-1 score between the observations of the generated and ground-truth reports are calculated to give the scores of the CE metric. \citet{Liu_2019} used a reward derived from the CE metric for Self-Critical Sequence Training (SCST), in place of an NLG metric. SCST enables the decoder to be fine-tuned with its own outputs as input---something that is not possible with standard training schema \citep{Rennie_2017}. This is achieved by using a reinforcement learning algorithm that uses the score between the generated and ground-truth reports as the reward. This was able to outperform the approach by \citet{Wang_2018} for all tested NLG metrics. Moreover, using SCST was shown to improve the diagnostic accuracy of the generated reports.

To embed expert knowledge, \citet{Zhang_2020} incorporated a Graph Convolutional Network (GCN)---whose nodes consisted of abnormalities---into the encoder. This was able to outperform the aforementioned CXR report generation approaches on multiple NLG metrics. The authors also developed a diagnostic accuracy metric based on the graph CNN---which showed that the graph CNN improves diagnostic accuracy.

More recently, Lovelace and Mortazavi employed a CNN followed by a series of Transformer layers as the encoder and a Transformer decoder. To improve diagnostic accuracy, the CheXpert labeler was used to extract observations from the generated report in a differentiable manner, where the loss between the observations of the generated and ground-truth reports was added to the training loss. This outperformed multiple approaches that employed an RNN as the decoder, where multiple NLG metrics and the CE metrics were used for evaluation \citep{Lovelace_2020}. 

Chen \textit{et al.} proposed a `memory-driven' Transformer decoder (R2Gen) \citep{Chen_2020}, and later built upon this by proposing the `Cross-modal Memory Network' (CMN) \citep{Chen_2021}. Using multiple NLG metrics and the CE metric, it was determined that the proposed approaches were able to outperform several previous CXR report generation approaches in the literature, each of which used an RNN as the decoder \citep{Jing_2019, Li_2018a, Jing_2018}.

\citet{Alfarghaly_2021} employed CheXNet \citep{Rajpurkar_2017a}---a DenseNet-121 ChestX-ray14 checkpoint---as the encoder and a Transformer as the decoder. The decoder was warm started with DistilGPT2, a general-domain NLG checkpoint. However, cross-attention was not used with each layer of the decoder, rather the embeddings of each predicted abnormality of CheXNet were fed as input to DistilGPT2. This method outperformed multiple approaches that utilised an RNN as the decoder for multiple NLG metrics.

\citet{Liu_2021a} proposed the Contrastive Attention model which compares the visual features of the current CXR to a pool of CXRs that have no abnormalities, where a ResNet-50 warm started with a CheXpert checkpoint was used to extract the visual features. Contrastive Attention comprises two modules, `aggregate attention' and `difference attention'. Aggregate attention finds the normal CXRs from the pool that are closest to the CXR in question. Difference attention involves two steps; for the first step, common features between the current CXR and the pool of normal CXRs are found. For the second step, the common features are subtracted from the features of the current CXR to capture `contrastive information'. This was able to outperform R2Gen, where multiple NLG metrics and the CE metrics were used for evaluation.

With the aim to imitate the interpretation process of radiologists, \citet{Liu_2021b} proposed the \textit{a Posterior}-and-\textit{Prior} Knowledge Exploring-and-Distilling approach (PPKED). Along with a CXR, the model consumes embeddings of the most common abnormalities found in the reports from the training set. It also consumes encoded reports from CXRs of the training set that have similar visual features to the current CXR. Finally, the model consumes the embedding from a knowledge graph that represents the most common abnormalities found in the training set. This outperformed R2Gen, where multiple NLG metrics and the CE metrics were used for evaluation.

\citet{Nooralahzadeh_2021} proposed an approach that generated a report from detected abnormalities in the CXR. First, a DenseNet-121 extracted visual features from a CXR, where a pre-trained CheXpert checkpoint was used for warm starting. Next, an encoder-to-decoder model called the meshed Transformer with memory ($\mathcal{M}^2$ Transformer) \citep{Cornia_2020} predicted the abnormalities from the visual features. The labels for the $\mathcal{M}^2$ Transformer were extracted from the CXR using the GCN of \citet{Zhang_2020}. These were fed to BART to generate the report, where BART is a sequence-to-sequence encoder-to-decoder model \citep{Lewis_2020}. The final method, named $\mathcal{M}^2$ Transformer Progressive, outperformed R2Gen, where multiple NLG metrics and the CE metrics were used for evaluation.

\subsection{CV checkpoints}

As highlighted in the previous subsection, the encoder is typically warm started with a ResNet or DenseNet  ImageNet-1K or CXR checkpoint. These are summarised in this section, along with more recent CV models and their associated checkpoints.

Residual and dense aggregations of layer outputs have been found to benefit training. Residual aggregations simplify the landscape of the loss function and prevent the vanishing and exploding gradient problems \citep{He_2016}. This allows the training of very deep CNNs, called ResNets. The dense aggregations of DenseNet offer direct feature re-usage, as deeper layers have access to the outputs of shallower layers \citep{Huang_2017}. Following this, \citet{Tan_2019} focused on efficiently scaling the depth, width, and input image size of CNNs, forming EfficientNets. As a result, EfficientNets are able to significantly outperform ResNets and DenseNets on ImageNet-1K, while remaining parameter efficient. An EfficientNet as the encoder, warm started or not, has not been investigated for CXR report generation. Domain-specific CXR checkpoints for CNNs also exist in the literature. \citet{Rajpurkar_2017a} proposed CheXNet, a DenseNet-121 checkpoint for multi-label classification of 14 abnormalities. The checkpoint was formed by additionally training an ImageNet-1K checkpoint on ChestX-Ray14.

More recently, Transformer encoders have been investigated for CV. \citet{Dosovitskiy_2020} proposed the Vision Transformer (ViT)---a Transformer checkpoint pre-trained on large general-domain image classification datasets which takes as input a set of patches from an image. ViT possesses several appealing features for CV, including no pooling, a full-receptive field, and robustness to occlusion \citep{Naseer_2021}. However, ViT does not posses the same inductive bias that makes CNNs an attractive option for medical CV tasks---a bias towards local spatial features. While the self-attention weights of a ViT head are able to model the relationship between patches, they do not model the relationship between pixels. This may be detrimental for CXR report generation, as many anatomical features of the chest---such as the pulmonary arteries---are represented in fine detail by the CXR. Moreover, ViT was only able to outperform ResNet when warm starting with checkpoints that have been trained using 30 million or more images \cite[Figure 4, ViT-B/32 vs. ResNet50x1 (BiT)]{Dosovitskiy_2020}. This lead to Dosovitskiy \textit{et al.} concluding that ViT does not generalise well when trained on insufficient amounts of data.

There are multiple improvements to ViT in the literature that  either use a self-supervised pre-training task or modify the Transformer to manually inject an inductive bias towards local spatial features. The Convolutional vision Transformer (CvT) replaces the linear layers of each self-attention head with two-dimensional convolutional layers, thus introducing an inductive bias to local spatial features into each head. This enabled CvT to outperform ViT on ImageNet-1K \citep{Wu_2021}. The Data-efficient image Transformer (DeiT) is an ImageNet-1K checkpoint that incorporated knowledge distillation into its set of pre-training tasks. Knowledge distillation involves two models; a smaller model---called the student---and a larger model---called the teacher. The student is trained to replicate the categorical distribution of the teacher. For DeiT, the teacher was a recent CNN. On ImageNet-1K, DeiT was able to outperform ViT and EfficientNet \citep{Touvron_2021}. 

The Cross-Covariance image Transformer (XCiT) utilises a transposed version of the self-attention mechanism. This results in the attention weights modelling the relationship between feature channels rather than patches. To model the relationship between local patches, each layer of XCiT employs a two-dimensional convolutional kernel. XCiT was able to outperform both EfficientNet and DeiT on ImageNet-1K \citep{ElNouby_2021}. Inspired by BERT, \citet{Bao_2021} used a self-supervised task, namely Masked Image Modelling (MIM), to form a Bidirectional Encoder representation from an image Transformer (BEiT). For MIM, the objective for each randomly masked patch is to predict its corresponding discrete visual token. The token for each patch was produced by a discrete Variational Auto-Encoder (VAE). BEiT was able to outperform both ViT and DeiT on ImageNet-1K.

\subsection{NLP checkpoints}

As detailed in Subsection \ref{sec:cxr_methods}, the Transformer is the standard decoder used by recent CXR report generation methods. In this subsection, we summarise the Transformer and its checkpoints. \citet{Vaswani_2017} proposed the Transformer---a sequence-to-sequence model for NLG. Its layers employ multiple scaled dot-product attention heads, each of which model the complex dependencies between its inputs. Due to this, Transformers more efficiently model long-range dependencies between subword tokens than RNNs and Temporal Convolutional Networks (TCNs) \cite[Table 1]{Vaswani_2017}. 

Building upon this, checkpoints for the encoder and decoder of the Transformer were formed with self-supervised pre-training tasks and extremely large, unlabelled corpora. Masked Language Modelling (MLM) and Next Sentence Prediction (NSP) are self-supervised pre-training tasks that were used to form Bidirectional Encoder Representations from Transformers (BERT)---a general-domain NLU checkpoint created from BookCorpus \citep{Zhu_2015} and English Wikipedia. As the encoder is non-causal, an output for BERT depends on all input tokens during MLM and NSP, and not just previous tokens. This enabled BERT to outperform previous Transformer checkpoints on multiple NLU tasks \citep{Devlin_2019}. The second version of the Generative Pre-trained Transformer (GPT2) is a checkpoint for the decoder of the Transformer. Using language modelling as the pre-training task and the WebText corpus \citep{Radford_2019}, GPT2 achieved state-of-the-art performance on several zero-shot NLG tasks \citep{Radford_2019}.

\citet{Sanh_2019} proposed DistilBERT, a distilled version of BERT that consumes 40\% fewer parameters while retaining 97\% of its performance. Knowledge distillation was incorporated into the training task, where DistilBERT was the student and BERT was the teacher. The authors also produced DistilGPT2 in a similar fashion.

\begin{figure}
\centering
\input{tex/natural_language_domain}
\caption{\label{fig:nl_domains}Non-proportional Venn diagram of the vocabulary of each natural language domain. As CXR reports are included in Intensive Care Unit (ICU) Electronic Health Records (EHRs), the vocabulary of CXR reports is a subset of the vocabulary of ICU EHRs. The vocabulary of ICU EHRs is a subset of the vocabulary of biomedical natural language, which is a subset of the vocabulary of the general domain. An example of a corpus belonging to each domain is also given.}
\end{figure}
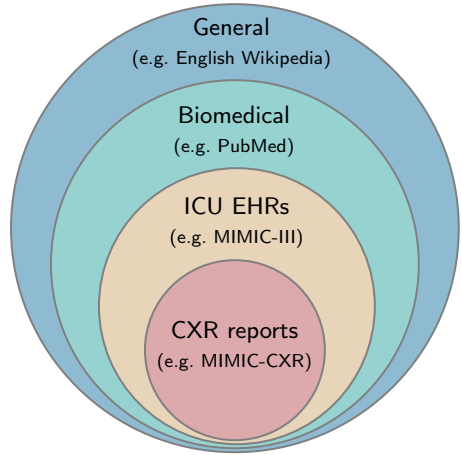

\begin{table}
\centering
\caption{Datasets sized for pre-training. The number of examples for Semantic Scholar are those used to pre-train SciBERT \citep{Beltagy_2019}.}\label{tab:dataset_sizes}
\scriptsize
\begin{tabular}{lrr}\toprule
\textbf{Dataset} &\textbf{No. of examples} \\\midrule
ImageNet-21K \citep{Deng_2009} &14M images \\
ImageNet-1K \citep{Russakovsky_2015} &1.3M images \\
CheXpert \citep{Irvin_2019} &224K images \\
BookCorpus \citep{Zhu_2015} &0.8B words \\
English Wikepedia &2.5B words \\
WebText  \citep{Radford_2019}
 &8M documents \\
Semantic Scholar \citep{Fricke_2018} & 1.14M papers \\
PubMed &4.5B words \\
PMC &13.5B words \\
MIMIC-III \citep{Johnson_2016} &2M notes \\
\bottomrule
\end{tabular}
\end{table}

The self-supervised tasks used to form BERT (namely MLM and NSP) have been used to form domain-specific NLU checkpoints. \citet{Beltagy_2019} formed a scientific NLU checkpoint from 1.14M documents of Semantic Scholar---a corpus of scientific publications. The resulting checkpoint, called SciBERT, outperformed BERT on several scientific NLU tasks. SciBERT also featured a domain-specific vocabulary constructed from Semantic Scholar. Multiple domain-specific checkpoints have been formed from biomedical corpora, particularly PubMed and PubMed Central (PMC). As seen in Table \ref{tab:dataset_sizes}, a benefit of PubMed and PMC is that they are both larger in size than the corpora used to train BERT and GPT2 (BookCorpus, English Wikipedia, and WebText). BioBERT is one such example, which outperformed BERT on several biomedical NLU tasks. However, BioBERT does not have a domain-specific vocabulary, instead it is inherited from BERT. This may be less than optimal, as medical terms not captured by BERT's vocabulary would be represented by multiple subwords \citep{Lee_2019}. PubMedBERT improves upon BioBERT by having a domain-specific vocabulary built from PubMed and PMC. The authors also demonstrate that training from scratch rather than additionally training a general-domain checkpoint such as BERT is best for biomedical NLU tasks---if there is sufficient data \citep{Gu_2020}.

There also exists multiple checkpoints formed from MIMIC-III---an Intensive Care Unit (ICU) Electronic Health Record (EHR) corpus. As seen in Figure \ref{fig:nl_domains}, the vocabulary of EHRs is a subset of the vocabulary of biomedical natural language. ClinicalBERT is an EHR NLU checkpoint formed by further training BioBERT with MLM and NSP on MIMIC-III \citep{Alsentzer_2019}. A similar checkpoint is BlueBERT, which is formed in two stages: BERT is further trained with MLM and NSP on PubMed. The resulting checkpoint is further trained with MLM and NSP on MIMIC-III, finally forming BlueBERT \citep{Peng_2019}. Both ClinicalBERT and BlueBERT were able to outperform BioBERT on several EHR NLU tasks. However, ClinicalBERT and BlueBERT do not have domain-specific vocabularies and instead rely on the vocabulary of BERT. Moreover, MIMIC-III is considerably smaller in size than the other corpora listed in Table \ref{tab:dataset_sizes}, raising concerns as to whether ClinicalBERT and BlueBERT can be successfully fine-tuned to CXR report generation.

\section{Contributions}

Other than determining which checkpoints are most suitable for warm starting the encoder and decoder of a CXR report generator, we aim to answer the following Research Questions (RQs):

\begin{description}
    \item[\textbf{RQ1}:] Are Transformer encoder CV checkpoints better than CNN checkpoints for warm starting the encoder?
    \item[\textbf{RQ2}:] Can an NLP checkpoint be effectively fine-tuned to model not only natural language but also visual features?
    \item[\textbf{RQ3}:] Are NLU checkpoints (e.g., BERT) better for warm starting the decoder than NLG checkpoints (e.g., GPT2)?
    \item[\textbf{RQ4}:] Are domain-specific checkpoints better for warm starting than general-domain checkpoints?
\end{description}

Our study differs from previous studies as follows:
\begin{itemize}
    \item We investigate the impact of publicly available  checkpoints on CXR report generation.
    \item We investigate CV models and checkpoints not previously considered for CXR report generation, namely, EfficientNet and Transformer encoder CV models.
    \item We investigate NLP checkpoints, specifically, NLU checkpoints such as BERT, and NLG checkpoints such as GPT2.
    \item We investigate domain-specific CV and NLP checkpoints.
    \item We present a case study of the final model that includes the attention weights of its cross-attention heads---to reveal what the model attends to when generating a report.
    \item We present a fine-grained evaluation on individual abnormalities, something that has been lacking in recent studies. 
\end{itemize}

\section{Datasets} \label{sec:dataset}

The Medical Information Mart for Intensive Care CXR dataset (\textbf{MIMIC-CXR}) consists of $377{,}110$ CXRs in both DICOM and JPEG formats, and $227{,}835$ English radiology reports associated with $64{,}588$ patients. The reports and CXRs were automatically de-identified \citep{Johnson_2019, Johnson_2019a}. \citet{DemnerFushman_2015} released a dataset called \textbf{IU X-Ray} that consisted of $3{,}955$ English radiology reports and $7{,}470$ CXRs in both DICOM and PNG formats, where both the reports and CXRs were de-identified automatically. Each report is associated with a single patient. We refer the reader to the survey conducted by \citet[Subsection 2.1]{Pavlopoulos_2021} for a more detailed analysis of MIMIC-CXR and IU X-Ray. 

\begin{figure*}
    \centering
    \includegraphics[scale=0.75]{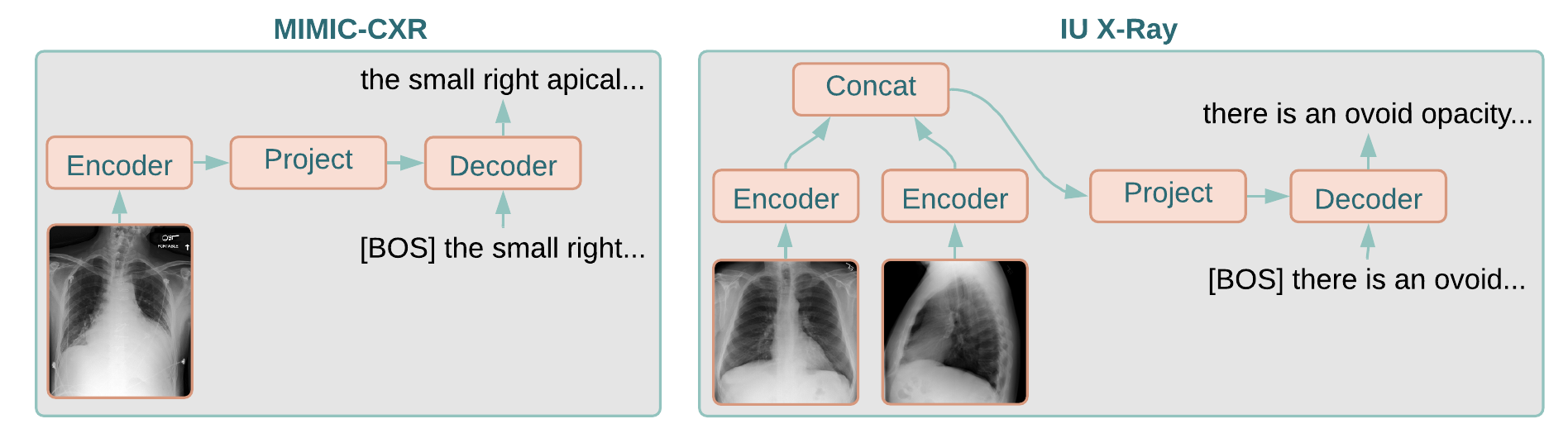}
    \caption{CXR report generation framework for (left) MIMIC-CXR and (right) IU X-Ray as in \citet{Chen_2020}. \texttt{[BOS]} is the beginning-of-sentence special token.}
    \label{fig:tasks}
\end{figure*}

In order to compare to previous studies, we adopt the dataset splits and labels of \citet{Chen_2020} for both MIMIC-CXR and IU X-Ray.\footnote{The MIMIC-CXR and IU X-Ray subsets used by \citet{Chen_2020} are available at: \url{https://github.com/cuhksz-nlp/R2Gen}.} For MIMIC-CXR, the splits are formed from $276{,}778$ of the CXRs. A portion of the reports were associated with multiple CXRs, meaning that they were the label for multiple examples. The subset splits are detailed in Table \ref{tab:dataset_splits}; the $276{,}778$ CXRs are split into $270{,}790$, $2{,}130$, and $3{,}858$ for training, validation, and testing, respectively. For IU X-Ray, the subsets are formed from $2{,}955$ of the reports, where each report is associated with a frontal and lateral view. This means that the model consumes two CXRs per example for IU X-Ray. The $2{,}955$ reports are split into $2{,}069$, $296$, and $590$ for training, validation, and testing, respectively. A training example for IU X-Ray is shown in Figure \ref{fig:tasks} (right); the frontal and  lateral views are given as input to the encoder and the label is the ground-truth report produced by a radiologist. For both MIMIC-CXR and IU X-Ray, each patient is included in only one of the subsets.

\begin{table*}
\centering
\caption{Subset splits of MIMIC-CXR and IU X-Ray from \citet{Chen_2020}.}\label{tab:dataset_splits}
\begin{tabular}{@{}llllcc@{}}
\toprule
\multirow{2}{*}{Dataset}& \multicolumn{3}{c}{No. of examples}& \multirow{2}{*}{Pixel depth} & \multirow{2}{*}{Views per example} \\ \cmidrule(lr){2-4}
& Training& Validation & Test&& \\ \midrule
MIMIC-CXR (splits from \citet{Chen_2020}) & $270{,}790$ & $2{,}130$& $3{,}858$ & 8-bit& 1 \\
IU X-Ray (splits from \citet{Chen_2020}) & $2{,}069$ & 296& 590 & 8-bit& 2 \\ \bottomrule
\end{tabular}
\end{table*}

\section{Problem formulation}

Given a set of $N$ CXRs $\mathcal{X}= \left\{ \pmb{X}_0, \dots, \pmb{X}_{N-1} \right\}$ where $\pmb{X}_i \in \mathbb{R}^{C \times W \times H}$ and $C$, $W$, and $H$ denote the number of channels, the width, and the height of each CXR, respectively, the aim is to generate a report ($\hat{R}$) whose target is the ground-truth report $R$. The encoder-to-decoder model generates $\hat{R}$ in multiple stages; first, the encoder $E$ produces a set of visual features from each of the CXRs $\mathcal{V}=\left\{\pmb{V}_0, \dots, \pmb{V}_{N-1} \right\}$, where the encoder processes each CXR independently ($E : \pmb{X}_i \rightarrow \pmb{V}_i$), $\pmb{V}_i \in \mathbb{R}^{S \times F}$, $S$ is the number of spatial positions, and $F$ is the number of features for each spatial position. Here, the visual features correspond to the last hidden state of the encoder. For some of the CV models, particularly the CNNs, the spatial positions are distributed over two axes---which are flattened to give $S$. Moreover, for some CV models, the spatial position and feature axes of the last hidden state are opposite (i.e. $\pmb{V}_i \in \mathbb{R}^{F \times S}$) and must be transposed. Next, the visual features for each CXR are concatenated along the spatial position axis: $\pmb{V}_{concat}=\texttt{concat} \left(\pmb{V}_0, \dots, \pmb{V}_{N-1} \right) \in \mathbb{R}^{D \times F}$, where \texttt{concat} is the concatenation operation and $D=S \times N$. Next, the visual features are projected to the hidden state size of the decoder $H$ using learned projection matrix $\pmb{P}\in \mathbb{R}^{F \times H}$: $\pmb{V}_{project}=\pmb{V}_{concat} \cdot \pmb{P} \in \mathbb{R}^{D \times H}$.

\input{tex/table_encoders}
\input{tex/table_decoders}

The projected visual features are fed to the decoder via a randomly initialised multi-head cross-attention module, which is inserted between the masked multi-head self-attention module and the feedforward neural network module of each decoder layer \cite[Section 3.1, Decoder]{Vaswani_2017}. The decoder then generates the report from the projected visual features in an autoregressive fashion ($D : \pmb{V}_{project} \rightarrow \hat{R}$). As shown in Figure \ref{fig:tasks}, $N=1$ for MIMIC-CXR and $N=2$ for IU X-Ray. This means that the concatenation operation is not required for MIMIC-CXR. For IU X-Ray, the two CXRs for an example correspond to a frontal and lateral view, whereas for MIMIC-CXR, the single CXR for an example is either a frontal or a lateral view, as in \citet{Chen_2020}.

\section{Methodology} \label{sec:methodology}

\subsection{Checkpoints}

\begin{figure}
    \centering
    \includegraphics[scale=0.75]{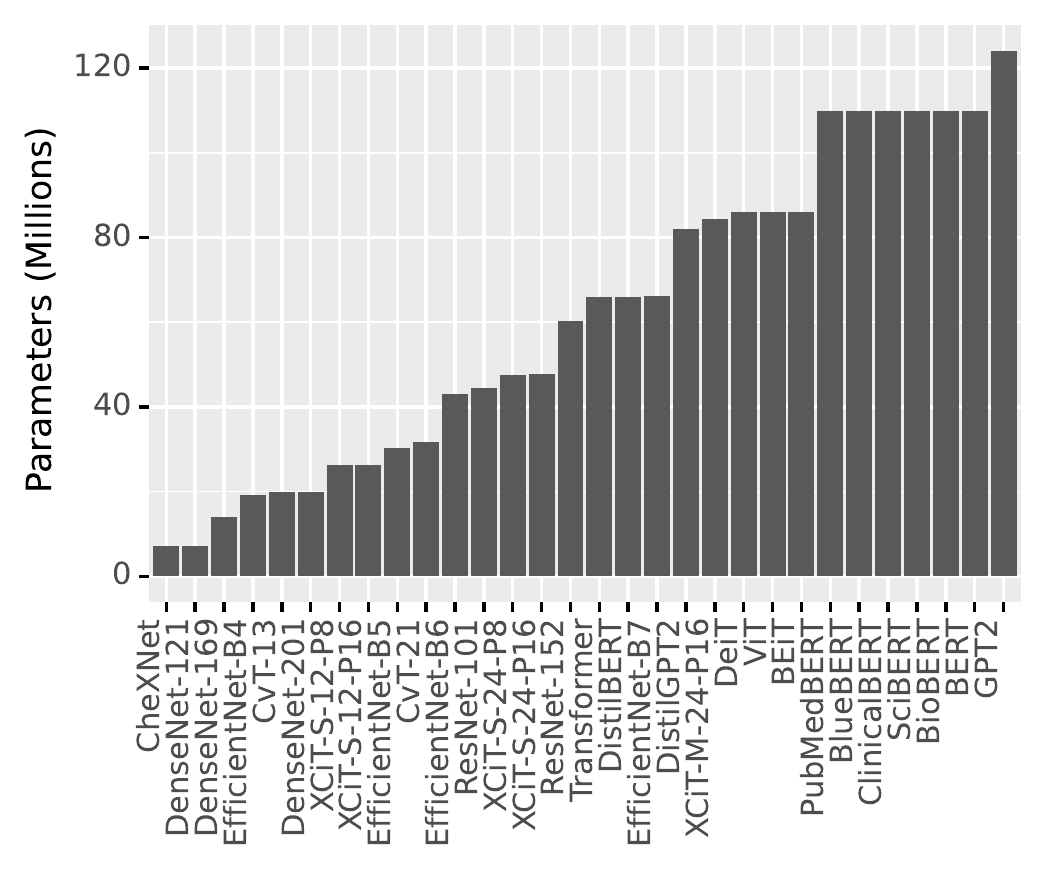}
    \caption{The number of parameters for each checkpoint.}
    \label{fig:num_params}
\end{figure}

In this study, we investigate the publicly available CV checkpoints described in Table \ref{tab:encoders} and the NLP checkpoints described in Table \ref{tab:decoders} for warm starting CXR report generation. The hyperparameters for each model were determined by those from their publicly available checkpoint. The training data that formed each checkpoint, the vocabulary of each NLP checkpoint, and the image width of each CV checkpoint is given. Note that the image height is equal to the image width for each CV checkpoint. The configuration of each CV model is also indicated. For each decoder checkpoint, if available, we use the uncased version, else, the cased version. This was due to the lowercase format of the ground-truth reports, as described in Subsection \ref{sec:report_pre-processing}. Finally, the number of parameters for each checkpoint are given in Figure \ref{fig:num_params}.

\subsection{Evaluation Metrics} \label{sec:metrics}

As in previous CXR report generation studies, we employ several NLG metrics and the CE metrics to evaluate the generated reports \citep{Pavlopoulos_2021}. The NLG metrics are word overlap measures that compute a similarity score based on the number of words shared between the generated and ground-truth reports. The Bi-Lingual Evaluation Understudy (\textbf{BLEU-\textit{n}}) measure computes the word \textit{n}-gram overlap between the generated and ground-truth reports, for example, BLEU-3 considers trigrams \citep{Papineni_2002}. The Metric for Evaluation of Translation with Explicit ORdering (\textbf{METEOR}) builds upon BLEU-1 by instead computing the $F_{\beta}$ score between unigrams (where recall is weighted higher than precision). METEOR also employs stemming and synonymy matching \citep{Banerjee_2005}.

\begin{figure*}
\centering
\scalebox{.79}{
\input{tex/cvt2distilgpt2}
}
\caption{High-level view of the architecture of \label{fig:cvt2distilgpt2}{\small CvT-21}{\large 2}{\small DistilGPT2}. See Figure \ref{fig:encoder_to_decoder} for a high-level depiction of the CXR report generation process with an encoder-to-decoder model. Here, it is configured for MIMIC-CXR with the \citet{Chen_2020} splits. $Q$, $K$, and $V$ are the queries, keys, and values, respectively, for multi-head attention \citep{Vaswani_2017}. * indicates that the linear layers for $Q$, $K$, and $V$ are replaced with the convolutional layers depicted below the multi-head attention module. \texttt{[BOS]} is the beginning-of-sentence special token. $N_l$ is the number of layers for each stage, where $N_l=1$, $N_l=4$, and $N_l=16$ for the first, second, and third stage, respectively. The head for DistilGPT2 is the same used for language modelling. Subwords produced by DistilGPT2 are separated by a vertical bar.}
\end{figure*}
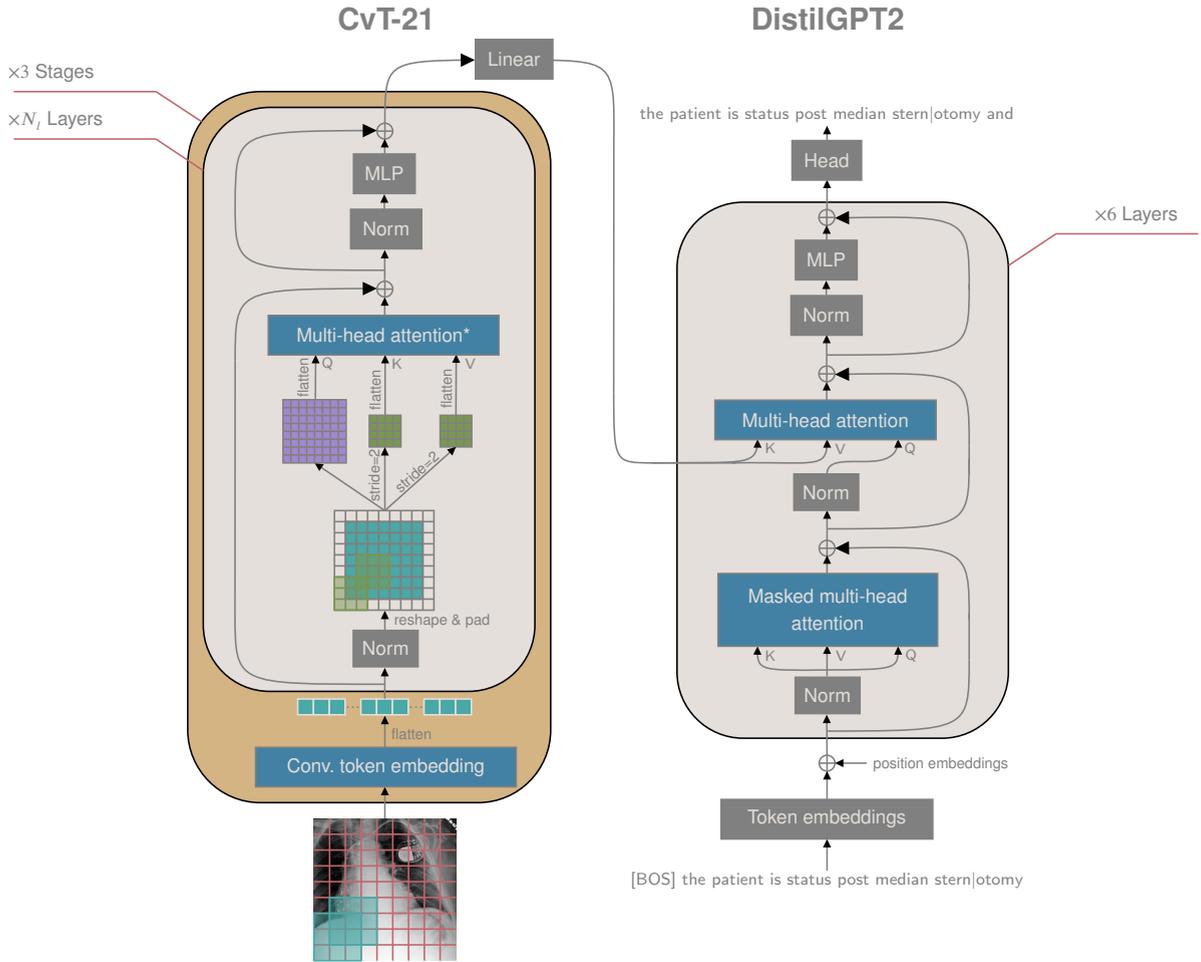

The Recall-Oriented Understudy for Gisting Evaluation with Longest common subsequence-based statistics (\textbf{ROUGE-L}) is the harmonic mean of ROUGE-L Recall and Precision. ROUGE-L Recall is ratio of the length of the longest common \textit{n}-gram shared by the generated and ground-truth reports, to the number of words in the ground-truth report. ROUGE-L Precision is identical to ROUGE-L Recall, except that the denominator is the number of words in the generated report \citep{Lin_2004}. The Consensus-based Image Description Evaluation (\textbf{CIDEr}) measure computes the cosine similarity between Term Frequency-Inverse Document Frequency (TF-IDF) \textit{n}-grams of the generated and ground-truth reports. Cosine similarities are calculated for 1 to 4-grams and their average is returned as the final score. By using TF-IDFs, terms that are infrequent in the corpus are rewarded, while terms that are common (e.g., stopwords) are penalised \citep{Vedantam_2015}.

It should be noted that the aforementioned word overlap measures do not necessarily capture diagnostic accuracy \citep{Babar_2021}. To more effectively evaluate this, we use the CE metrics that was previously employed to evaluate CXR report generators \citep{Liu_2019, Chen_2020}. In place of the CheXpert labeler, we use CheXbert---a BERT-based approach to extracting the 14 CheXpert observations from a given report. CheXbert demonstrated a statistically significant improvement in performance over the CheXpert labeler (CheXbert had a macro-averaged F-1 score improvement of 0.055 over the CheXpert labeler), while being 120 times faster (when a GPU is available) \citep{Smit_2020}. The CE classification scores are calculated as follows; CheXbert first determines the class of each of the 14 observations from the generated and ground-truth reports as either \textit{positive}, \textit{negative}, \textit{uncertain}, or \textit{no mention}.\footnote{The 14 observations include \textit{enlarged cardiomediastinum}, \textit{cardiomegaly}, \textit{lung opacity}, \textit{lung lesion}, \textit{edema}, \textit{consolidation}, \textit{pneumonia}, \textit{atelectasis}, \textit{pneumothorax}, \textit{pleural effusion}, \textit{pleural other}, \textit{fracture}, \textit{support devices}, and \textit{no finding}.} Next, the multi-class classification task for each observation is converted into a binary classification task; observations that are \textit{positive} are considered positive results, and observations that are \textit{negative}, \textit{uncertain}, or \textit{no mention} are all considered negative results. Following this, true positives, true negatives, and false negatives are found by comparing the observations of the generated reports to that of the ground-truth reports. From this, the example-based precision, recall, and F-1 scores are found over the 14 observations, while the label-based precision, recall, and F-1 scores are found for each of the 14 observations \cite[Section 7]{Sorower_2010}.


\subsection{CXR pre-processing and augmentation}
\label{sec:cxr_pre-processing}

Following \citet{Chen_2020}, we adopted the CXRs in JPEG and PNG formats for MIMIC-CXR and IU X-Ray, respectively, resulting in an 8-bit pixel depth and three identical channels ($C=3$). This was required as each CV checkpoint was configured to take three channels as input (even CheXNet). The following pre-processing and augmentation steps were then applied to each CXR; first, a given CXR was resized using bilinear interpolation so that its smallest side had a length of $W+64$, where $W$ is given in Table \ref{tab:encoders}, and its largest side was set such that it maintained the aspect ratio. Next, the resized CXR was cropped to a size of $\mathbb{R}^{3 \times W \times H}$, where $W=H$. The crop location was random during training and centered during testing. After this and only during training, the CXR was rotated around its centre where the angle of rotation was sampled from $\mathcal{U}{[{-5^{\circ}, 5^{\circ}}]}$. Finally, the CXR was standardised using the mean and standard deviation of each channel provided with each CV checkpoint.

\subsection{Report pre-processing and generation}
\label{sec:report_pre-processing}

In order to compare to previous CXR report generators, we formatted the ground-truth reports identically to \citet{Chen_2020}. This was achieved by allowing a maximum of 60 words per report (words after the $60^{th}$ word were removed), changing upper-cased letters to lowercase, removing special characters, and replacing words that occurred less than three times in the corpus with a special unknown token. During testing, the maximum amount of subwords that the decoder could generate was set to 128, as each word could be represented by multiple subwords. Beam search with a beam size of four was used during testing when generating the reports. During validation a beam size of one was used (i.e., greedy search).

\subsection{Fine-tuning} \label{sec:pre_training_&_fine_tuning}

Teacher forcing was used for fine-tuning \citep{Williams_1989}. Each model was implemented in PyTorch version 1.9.0 and trained with $4 \times$NVIDIA P100 16GB GPUs with automatic mixed precision. To select the best epoch for a model, we use the highest CIDEr validation score. Due to phenomena such as mild overparameterisation, not every random initialisation of a models parameters will lead to a global minimum during gradient descent \citep{Simsek_2021}. This leads to large performance variability between training runs. To account for this, we performed multiple training runs for each model. Finally, only models fine-tuned on the MIMIC-CXR training set were tested on the MIMIC-CXR test set and models fine-tuned on the IU X-Ray training set were tested on the IU X-Ray test set. The following configuration was used to fine-tune each model with teacher forcing:
\begin{itemize}
    \item Categorical cross-entropy as the loss function.
    \item \textit{AdamW} optimiser for gradient descent optimisation \citep{Loshchilov_2019}. An initial learning rate of $1e-5$ and $1e-4$ for the encoder and all other parameters, respectively, following \citet{Chen_2020}. All other hyperparameters for \textit{AdamW} were set to their defaults.
    \item A mini-batch size of 16.
    \item Early stopping with a patience of 10 epochs and a minimum delta of $1e-4$.
    \item The validation CIDEr score was the monitored metric for early stopping.
\end{itemize}

\input{tex/table_mimic_cxr}
\input{tex/table_example_ce}

\input{tex/table_iu_x_ray}

\subsection{Statistical analysis}

The Confidence Intervals (CIs) in Tables \ref{tab:mimic_cxr}, \ref{tab:example_ce_metrics}, \ref{tab:iu_x_ray}, and \ref{tab:label_ce_metrics} were found with bootstrapping; the test set was resampled with replacement $1{,}000$ times, where the size of each sample was equal to the size of the respective test set (as indicated in Table \ref{tab:dataset_sizes}) \citep{Efron_1979}. The $95\%$ CIs were then calculated from the mean scores of each sample. In Subsections \ref{sec:encoder_comparison}, \ref{sec:decoder_comparison}, and \ref{sec:num_examples} we perform statistical tests on the NLG metric scores where the checkpoint type was the factor and the scores for each example of the MIMIC-CXR test set and for each of the training runs were the dependent variables. For each test, we used a \textit{p}-value of 0.05. First, a Levene's test revealed that the variances of the scores were not homogeneous. This lead us to using a one-way Welch's ANOVA to determine if there was a significant difference between the scores of the checkpoints. If a significant difference existed, Games-Howell tests were used to perform a post hoc analysis.

\section{Results and discussion} \label{sec:r&d}

In this section, we first compare the final model shown in Figure \ref{fig:cvt2distilgpt2}, whose encoder and decoder is warm started with the CvT-21 ImageNet-21K and DistilGPT2 checkpoints, respectively, (i.e., {\small CvT-21}{\large 2}{\small DistilGPT2}) to current CXR report generators in the literature (Subsection \ref{sec:comparison}). Next, we evaluate the diagnostic performance of {\small CvT-21}{\large 2}{\small DistilGPT2} on 14 observations (Subsection \ref{sec:observations}). Following this,  the case study in Subsection \ref{sec:case_study} provides insight as to how {\small CvT-21}{\large 2}{\small DistilGPT2} interprets a CXR when generating a report. In Subsection \ref{sec:encoder_comparison}, we compare the CV checkpoints and determine which is best for warm starting the encoder. Moreover, we compare the NLP checkpoints in Subsection \ref{sec:decoder_comparison} and determine which is best for warm starting the decoder. We also answer RQ1-RQ4 in Subsections \ref{sec:encoder_comparison} and \ref{sec:decoder_comparison}. Finally, we provide the reader with the limitations of this investigation and several future recommendations.

\subsection{Comparison to current methods} \label{sec:comparison}

Here, we compare {\small CvT-21}{\large 2}{\small DistilGPT2} to other CXR report generators in the literature that were evaluated on the labels of \citet{Chen_2020}.\footnote{Scores for the other CXR report generators in the literature were taken from their respective articles.} The NLG metric scores on the MIMIC-CXR test set are shown in Table \ref{tab:mimic_cxr}. {\small CvT-21}{\large 2}{\small DistilGPT2} attained the highest mean scores for all NLG metrics. This indicates that the reports generated by {\small CvT-21}{\large 2}{\small DistilGPT2} are more similar to the reports produced by radiologists than those of previous approaches.

For the example-based CE scores in Table \ref{tab:example_ce_metrics}, $\mathcal{M}^2$ Transformer Progressive had the highest recall---but also the lowest precision. The authors of  $\mathcal{M}^2$ Transformer Progressive speculated that the high false positive rate was due to its generated reports having a longer length on average than the ground-truth reports \cite[Section 4]{Nooralahzadeh_2021}. {\small CvT-21}{\large 2}{\small DistilGPT2} was able to attain the highest precision and the second highest recall---leading to the highest F-1 score. This indicates that {\small CvT-21}{\large 2}{\small DistilGPT2} is more diagnostically accurate than previous approaches. 

The NLG metric scores on the IU X-Ray test set are shown in Table \ref{tab:iu_x_ray}. {\small CvT-21}{\large 2}{\small DistilGPT2} attained the highest mean scores for BLEU-3, BLEU-4, METEOR, and CIDEr. However, Contrastive Attention attained the highest mean BLEU-1 and ROUGE-L scores, while PPKED and $\mathcal{M}^2$ Transformer Progressive both attained the highest mean BLEU-2 score. This could indicate that the training set size of IU X-Ray is too small for {\small CvT-21}{\large 2}{\small DistilGPT2}. Alternatively, concatenating the visual features of the two CXRs---as shown in Figure \ref{fig:tasks} (right)---may not be the best multi-source combination technique, in fact, \citet{Libovicky_2018} found it to be the worst technique for a Transformer decoder.

\subsection{Performance on different observations} \label{sec:observations}

Next, to get an indication of the diagnostic accuracy of {\small CvT-21}{\large 2}{\small DistilGPT2} for different abnormalities, we analyse its performance on each of the 14 CheXpert observations. Scores for the label-based CE metrics are given in Table \ref{tab:label_ce_metrics}, along with the macro- and micro-averaged scores over the observations. {\small CvT-21}{\large 2}{\small DistilGPT2} attained the highest precision and recall for \textit{support devices}---which is expected as it was one of the most frequent observations in the training set. {\small CvT-21}{\large 2}{\small DistilGPT2} performed well for \textit{pleural effusion} and \textit{cardiomegaly}, of which both are frequently observed in the training set. For \textit{no finding}, {\small CvT-21}{\large 2}{\small DistilGPT2} demonstrated high precision but poor recall. Oppositely, it had a high recall and low precision for \textit{lung opacity}. The performance of {\small CvT-21}{\large 2}{\small DistilGPT2} for \textit{consolidation}, \textit{enlarged cardiomediastinum}, \textit{lung lesion}, \textit{fracture}, and \textit{pleural other} was poor---likely due to the infrequency of these observations in the training set. This was also the case with previous CXR report generators, for example, the approach by \citet[Table 3]{Liu_2019} performed poorly on the rarer abnormalities of MIMIC-CXR. This suggests that the class imbalance of the observations in the MIMIC-CXR training set leads to poor performance. Interestingly, recall was higher for all observations except \textit{no finding} and \textit{support devices}, as reflected by the macro- and micro-averaged scores for precision and recall. This shows that {\small CvT-21}{\large 2}{\small DistilGPT2} has a higher false positive rate but a lower false negative rate.

\input{tex/table_label_ce}

\subsection{Case study} \label{sec:case_study}

\begin{figure*}
\centering
\scalebox{.85}{
\input{tex/case_study}
}
\caption{\label{fig:case_study} Case study of CXR \textit{a0578edb-12a640ca-1ddab351-089c4d4c-00bb6f19} from study \textit{s54265960} of patient \textit{p18615099} from the MIMIC-CXR test set. The `ground-truth' report was produced by a radiologist, while the `generated' report was produced by {\small CvT-21}{\large 2}{\small DistilGPT2}. The CXR displayed has been pre-processed for testing. Subwords produced by DistilGPT2 are separated by a vertical bar. Each cross-attention weight matrix is min-max normalised and then scaled element wise using $\exp \left(1- a^{-1} \right)$ for ease of interpretation (where $a$ is an attention weight). A beam size of one (i.e., a greedy search) was used to produce the report. One cross-attention head from each layer of DistilGPT2 was selected for analysis.}
\end{figure*}
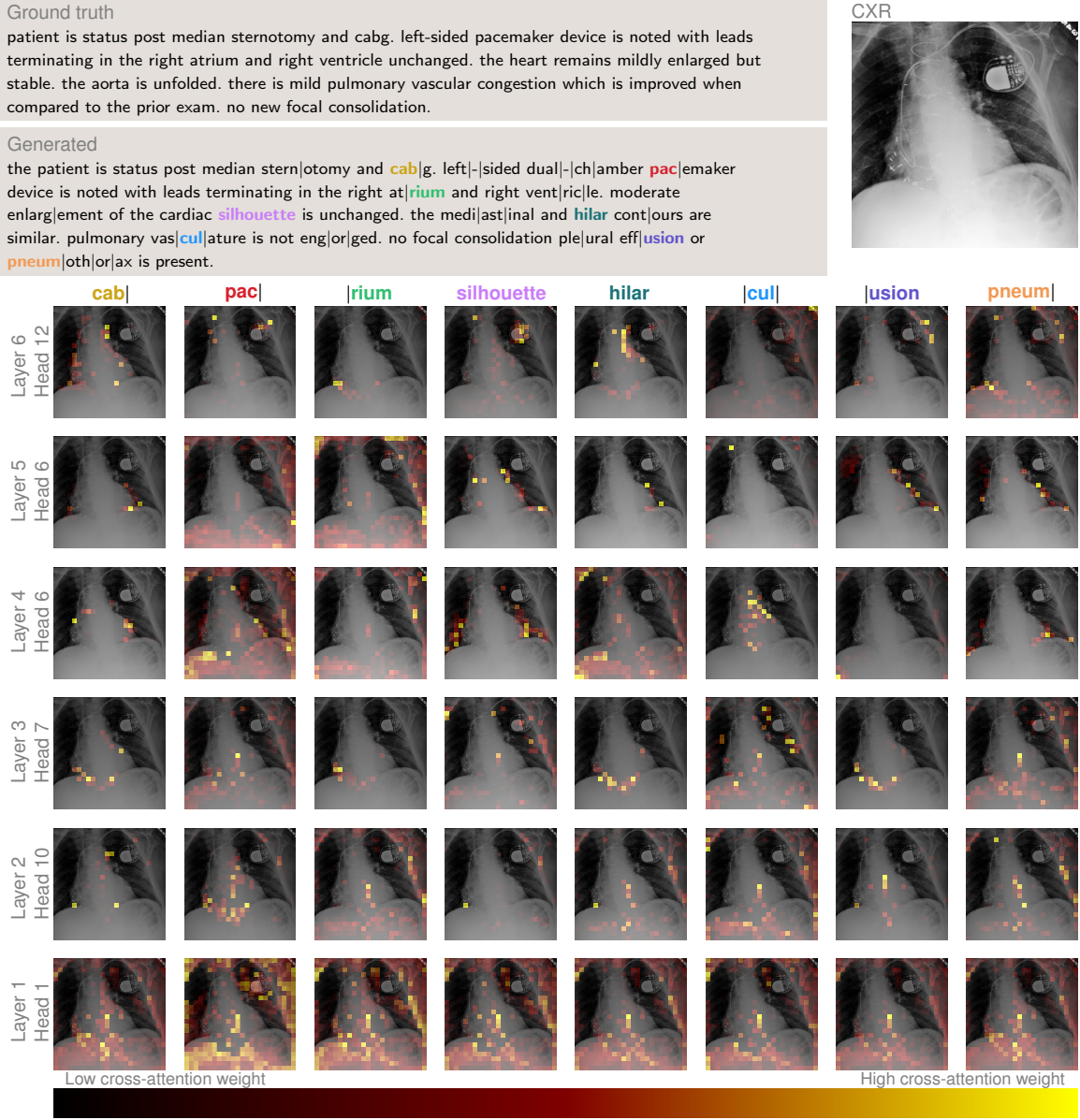

Here, we observe the attention weights of a subset of the cross-attention heads of {\small CvT-21}{\large 2}{\small DistilGPT2} as they collectively generate a word (or subword) for the case study in Figure \ref{fig:case_study} (where subwords are separated by a vertical bar). The cross-attention layers in the decoder are the means of conditioning the report generation process on the visual features of the CXR, where each layer comprises multiple cross-attention heads. Cross-attention heads are parallel units within the layer that compute an output based on the relationship between its two inputs, in this case, the visual features and the previously generated words. Here, the visual features are a high-level representation of the regions of the CXR. For a cross-attention head, a high-level representation of both the visual features and words are compared via their dot product, giving the attention weights. A higher weight indicates a higher similarity between the CXR region and the word. Through this, we can get an indication of what words and CXR regions the model has learnt to correctly or incorrectly relate to one another. This case study was selected to highlight both strengths and weaknesses of the model and of the current paradigm of CXR report generators. It should be noted that higher layers (e.g., Layer 6) are considered higher-level representations of lower layers (e.g., Layer 1)---which typically model more primitive relationships. Because we are interested in the anatomical aspects of the CXR, we direct the readers attention to the higher layers.

Comparing the generated and ground-truth reports, {\small CvT-21}{\large 2}{\small DistilGPT2} correctly predicted that the patient has had Coronary Artery Bypass Graft (CABG) surgery via a median sternotomy incision, although it is unclear from both the generated and ground-truth reports which of the coronary arteries were bypassed. It can be seen that the heads attended to the vascular clips when generating the subword `cab|'---a key indicator of CABG. Next, {\small CvT-21}{\large 2}{\small DistilGPT2} correctly identified the left-side pacemaker with its leads terminating in the right atrium and ventricle. Note that in the ground-truth report, it was mentioned that the positioning of the leads remained `unchanged'---something that is impossible for {\small CvT-21}{\large 2}{\small DistilGPT2} to infer as it has not observed the previous study. It can be seen that Head 12 of Layer 6 and Head 1 of Layer 1 attend to the pacemaker when generating the subword `pac|'. However, not one of the presented heads attend to the lead in the right atrium when generating the subword `|rium'.

{\small CvT-21}{\large 2}{\small DistilGPT2} indicated mild cardiomegaly with the phrase ``a moderate enlargement of the cardiac silhouette''. However, {\small CvT-21}{\large 2}{\small DistilGPT2} specified that the mild cardiomegaly was unchanged, referring to a previous, unobserved study. This is because {\small CvT-21}{\large 2}{\small DistilGPT2} has learnt to often describe abnormalities in the context of disease progression due to the ground-truth reports of the training set frequently referring to previous studies. When generating the word `silhouette', Head 6 of Layers 4 and 5 attend to regions of the cardiac silhouette. Moreover, {\small CvT-21}{\large 2}{\small DistilGPT2} missed the widened and decreased curvature of the aortic arch indicating that the aorta is unfolded. {\small CvT-21}{\large 2}{\small DistilGPT2} also mentioned that the mediastinal and hilar contours are similar, which is not a valid interpretation due to the word \textit{similar}. Not one of the heads in Figure \ref{fig:case_study} attends to the right or left hilar points when generating the word `hilar'. 

{\small CvT-21}{\large 2}{\small DistilGPT2} also incorrectly predicted no pulmonary vascular congestion, while it is reported as mild and an improvement over a previous study in the ground-truth report. Head 7 of Layer 3 does pay some attention to the pulmonary vessels when generating the subword `|cul|'. However, it is difficult to determine an increased prominence of the pulmonary vessels without a prior normal CXR of the patient. Lastly, {\small CvT-21}{\large 2}{\small DistilGPT2} mentions which abnormalities were not present in the CXR; no focal consolidation, pleural effusion, or pneumothorax. Some of the presented heads attend to regions of the pleura when generating the subword `|usion', especially Head 6 of Layer 5. When looking for signs of pneumothorax, the heads did not attend to the position of the trachea, however, some attention was paid to each of the hemidiaphragms.

\begin{figure*}
    \centering
    \includegraphics[scale=0.9]{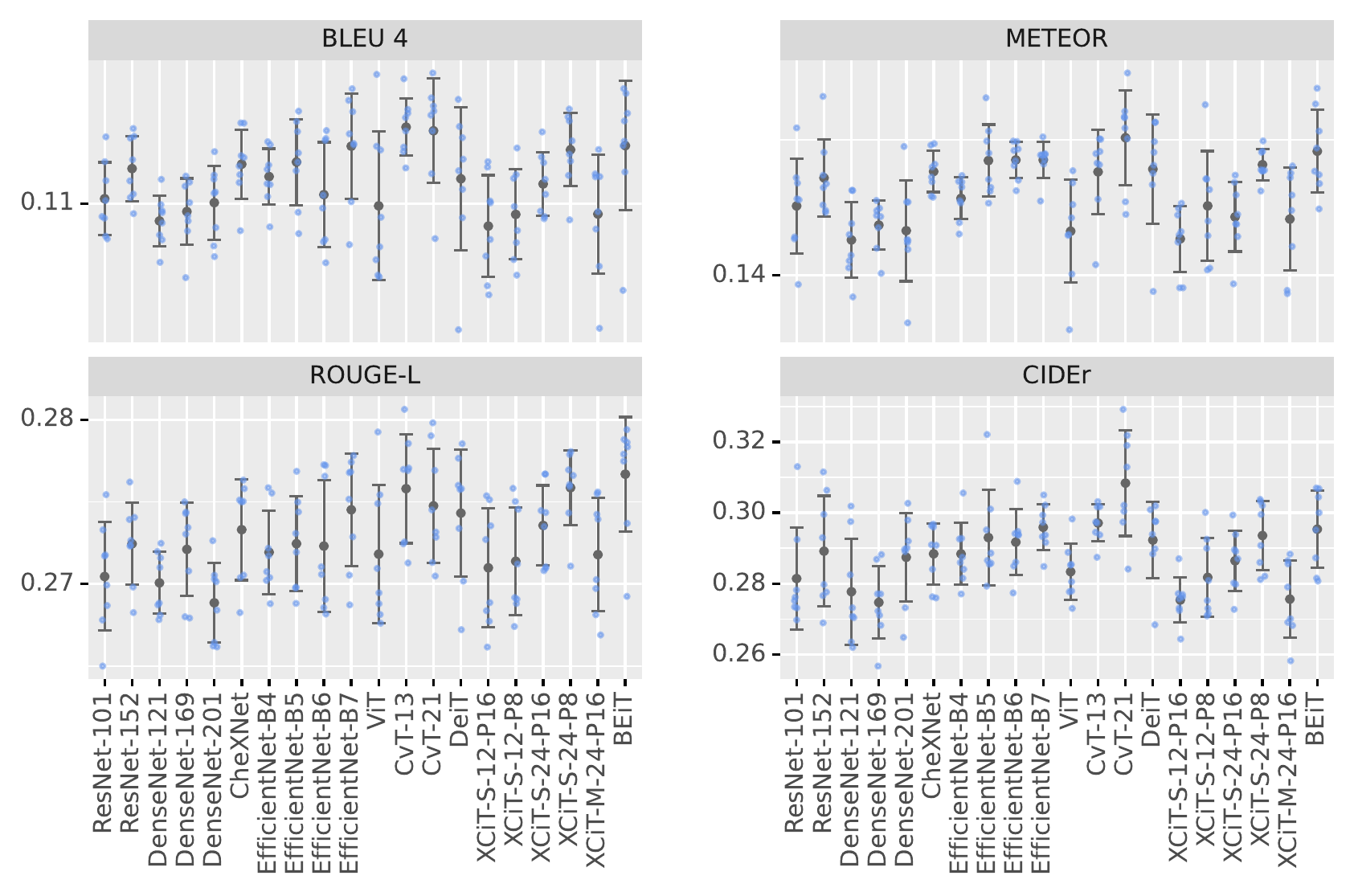}
    \caption{NLG metric scores for each CV checkpoint when warm starting the encoder. DistilGPT2 was used to warm start the decoder. The black dots and the error bars indicate the mean and standard deviation over the training runs, respectively.}
    \label{fig:encoder_scores}
\end{figure*}

\begin{figure*}
    \centering
    \includegraphics[scale=0.9]{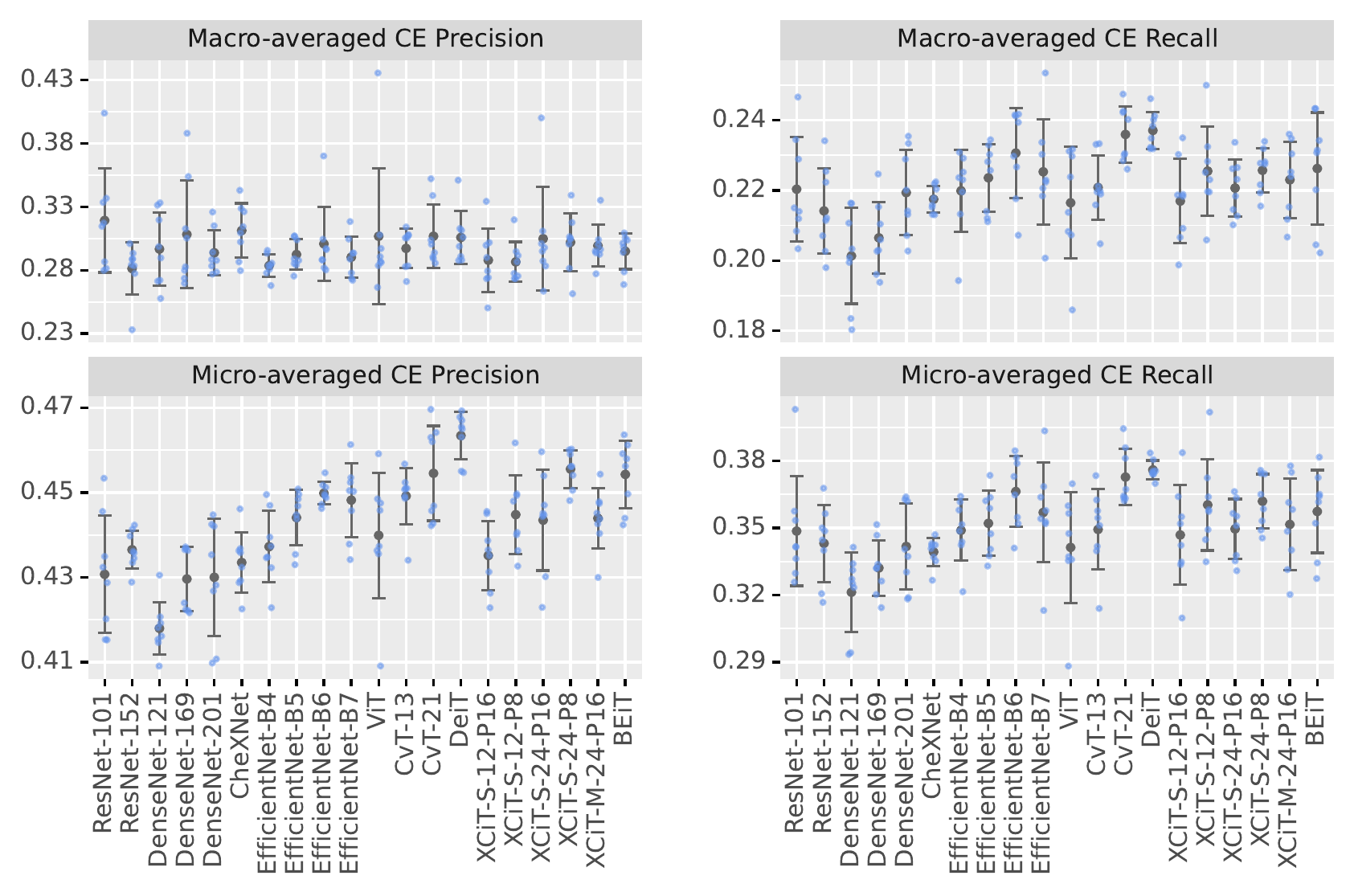}
    \caption{Label-based CE scores for each CV checkpoint when warm starting the encoder. DistilGPT2 was used to warm start the decoder. The black dots and the error bars indicate the mean and standard deviation over the training runs, respectively.}
    \label{fig:encoder_cls_scores}
\end{figure*}

\subsection{Best checkpoint for warm starting the encoder} \label{sec:encoder_comparison}

In this subsection, we determine which CV checkpoint is best for warm starting the encoder. To succeed, the pre-training task of the CV checkpoint must be transferable to that of extracting salient visual features from CXRs. To make this investigation more tractable, the following methodology was adopted; for training, we use the 50K subset of MIMIC-CXR's training set described in Subsection \ref{sec:num_examples}; we make the assumption that the best CV checkpoint is independent of the NLP checkpoint used to warm start the decoder. Following this, the decoder of each encoder-to-decoder model is warm started with DistilGPT2. Eight training runs are performed for each CV checkpoint to account for the variability introduced by randomly initialising the parameters of $\pmb{P}$ and the cross-attention modules of the decoder. The scores for the NLG metrics are shown in Figure \ref{fig:encoder_scores}, while the label-based CE metrics are shown in Figure \ref{fig:encoder_cls_scores}. A one-way Welch's ANOVA revealed a statistically significant difference between the NLG metric scores of each CV checkpoint, showing that they have an effect on performance. 

CvT-13 attained the highest mean BLEU-4 score; CvT-21 the highest mean METEOR and CIDEr scores; BEiT the highest mean ROUGE-L score; ResNet-101 the highest macro-averaged precision; DeiT the highest macro-averaged recall and micro-averaged precision and recall. Between the two best performing checkpoints, namely, CvT-21 and DeiT, CvT-21 attained higher mean BLEU-4, METEOR, ROUGE-L, CIDEr, and macro-averaged precision scores, while DeiT achieved a higher macro-averaged recall and micro-averaged precision and recall. However, Games-Howell tests revealed no significant difference between their NLG metric scores. As CvT-21 and DeiT cannot be separated based on their performance, we instead select based on their parameter efficiency. Hence, we select CvT-21 over DeiT as it consumes drastically fewer parameters (31.6M vs. 86M).

Comparing the CNN ImageNet-1K checkpoints, EfficientNet (B6 and B7) attained the highest scores for each metric (except for macro-averaged precision). In fact, Games-Howell tests revealed a significant difference between the METEOR scores of EfficientNet-B7 and each of the DenseNets. This is likely due to the improvements proposed by \citet{Tan_2019} to scale the depth, width, and input image size of the network to more efficiently use the networks parameters. Next, we compare the DenseNet-121 ImageNet-1K checkpoint to its domain-specific equivalent, namely CheXNet. CheXNet demonstrates a marked improvement, attaining higher mean scores for all metrics with Games-Howell tests revealing a statistically significant difference between their NLG metric scores. 

\textbf{Answer to RQ4 (for CV checkpoints)}: CheXNet demonstrates that there is a clear advantage to warm starting the encoder with a domain-specific CV checkpoint over a general-domain CV checkpoint.

ViT lacked performance, attaining lower mean scores for all metrics (except macro-averaged precision) than EfficientNet-B5, B6, and B7. Moreover, Games-Howell tests confirmed a significant difference between the METEOR scores of EfficientNet-B5, B6, and B7 versus ViT. Its hard to pinpoint the cause of the performance difference as many variables exist, such as the number of parameters and the training dataset. Another reason could be due to ViTs inability to model intra-patch visual features with its self-attention weights. The performance of DeiT was on par with EfficientNet, with DeiT attaining a higher macro- and micro-averaged precision and recall than EfficientNet-B7, whereas EfficientNet-B7 attained higher mean BLEU-4, METEOR, ROUGE-L, and CIDEr scores. Additionally, Games-Howell tests revealed no significant difference between the NLG metric scores of the two. The performance of BEiT was also on par with EfficientNet, with BEiT attaining higher mean BLEU-4, METEOR, ROUGE-L, and micro-averaged precision than EfficientNet-B7, whereas EfficientNet-B7 attained higher CIDEr, macro-averaged precision and recall, and micro-averaged recall. Moreover, Games-Howell tests revealed a significant difference between only the ROUGE-L scores of BEiT and EfficientNet-B4, B5, and B6 (in favour of BEiT), but not for B7. While having a similar architecture to ViT, the pre-training tasks of DeiT and BEiT enable them to perform comparatively to EfficientNet.
 
Finally, we analyse adaptations of the Transformer, namely XCiT and CvT. XCiT-S-24-P8 attained higher mean ROUGE-L and CIDEr scores than the EfficientNets, as well as a higher macro- and micro-averaged precision. However, Games-Howell tests only revealed significant difference for ROUGE-L between EfficientNet (B4, B5, and B6), but not for B7 (in favour of XCiT-S-24-P8). Moreover, EfficientNet-B6 and B7 attained higher mean BLEU-4 and METEOR scores as well as a higher macro- and micro-averaged recall. CvT-21 attained higher mean scores than the EfficientNets for all metrics, indicating that CvT-21 is the only Transformer-based CV checkpoint that can consistently outperform EfficientNet. However, Games-Howell tests revealed no statistically significant difference between the NLG metric scores of EfficientNet-B7 and CvT-21. 

\textbf{Answer to RQ1}: Currently, it seems that Transformers require pre-training with a task such as MIM or distillation to perform comparatively to CNNs. However, the performance of CvT suggests that incorporating convolutional layers into the Transformer alleviates it from having to learn the inductive bias of local spatial feature processing, allowing it to outperform EfficientNet.


\subsection{Best checkpoint for warm starting the decoder} \label{sec:decoder_comparison}

\begin{figure}
    \centering
    \includegraphics[scale=0.9]{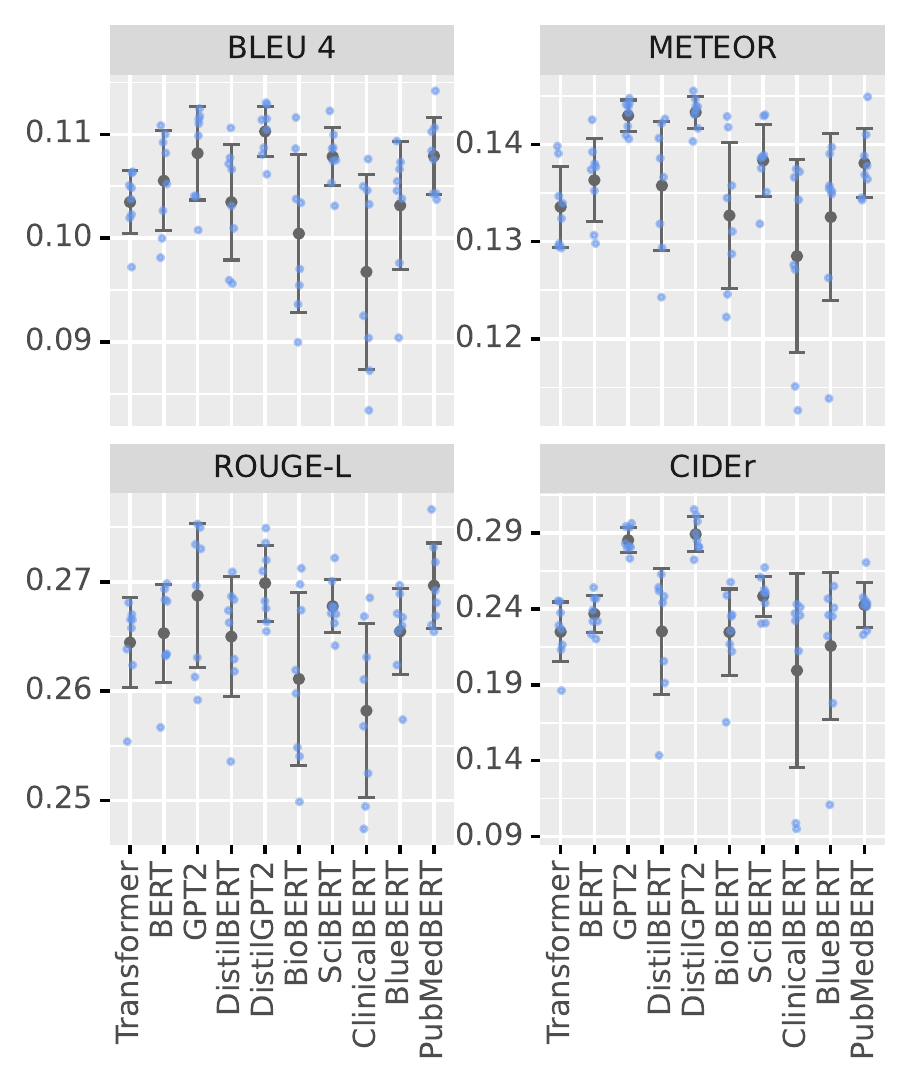}
    \caption{NLG metric scores for each NLP checkpoint when warm starting the decoder. A ResNet-101 ImageNet-1K checkpoint was used to warm start the encoder. The black dots and the error bars indicate the mean and standard deviation over the training runs, respectively.}
    \label{fig:decoder_scores}
\end{figure}

\begin{figure}
    \centering
    \includegraphics[scale=0.9]{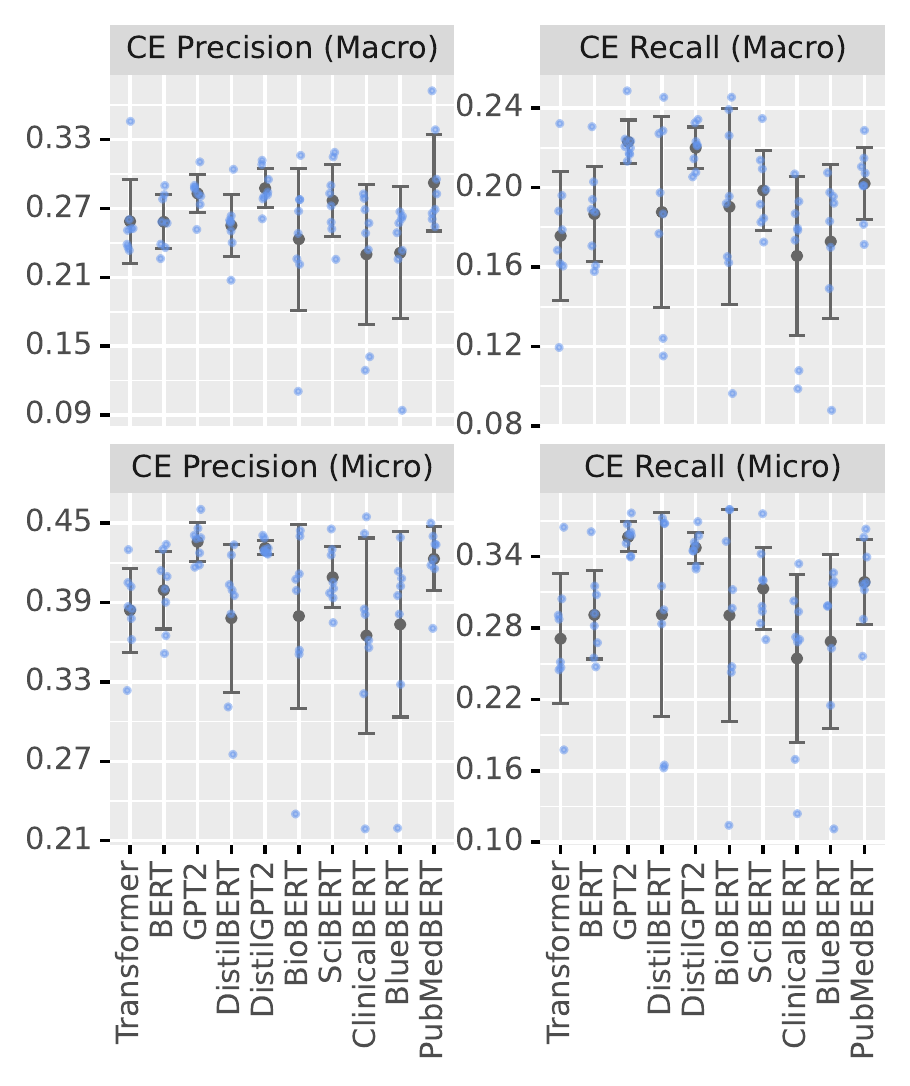}
    \caption{Label-based CE scores for each NLP checkpoint when warm starting the decoder. A ResNet-101 ImageNet-1K checkpoint was used to warm start the encoder. The black dots and the error bars indicate the mean and standard deviation over the training runs, respectively.}
    \label{fig:decoder_cls_scores}
\end{figure}

This subsection is identical in methodology to the previous subsection, except that now we determine which NLP checkpoint is best for warm starting the decoder. A ResNet-101 ImageNet-1K checkpoint was employed to warm start the encoder of each encoder-to-decoder model following the assumption that the best NLP checkpoint is independent of the CV checkpoint used to warm start the encoder. To succeed, the pre-training task of an NLP checkpoint must be transferable to that of generating reports from visual features. The scores for the NLG metrics are shown in Figure \ref{fig:decoder_scores}, while the label-based CE metrics are shown in Figure \ref{fig:decoder_cls_scores}. A one-way Welch's ANOVA revealed a statistically significant difference between the NLG metric scores of each checkpoint, showing that they have an effect on performance. 

DistilGPT2 attained the highest mean BLEU-4, METEOR, ROUGE-L, and CIDEr scores while GPT2 attained the highest macro-averaged recall and micro-averaged precision and recall. Furthermore, Games-Howell tests revealed a statistically significant difference between the scores of both GPT2 and DistilGPT2 versus the remaining checkpoints for both METEOR and CIDEr, indicating that GPT2 and DistilGPT2 are the most suitable checkpoints for warm starting the decoder. Comparing GPT2 to DistilGPT2, Games-Howell tests revealed no significant difference between their NLG metric scores, showing that distilling GPT2 to 66\% of its parameters has no effect on performance.\footnote{Games-Howell tests also revealed no significant difference between the NLG metric scores of BERT and DistilBERT, again demonstrating that distillation does not impact performance.} Due to its parameter efficiency, we select DistilGPT2 over GPT2 as the best checkpoint for warm starting the decoder.

\textbf{Answer to RQ2}: GPT2 demonstrated that an NLP checkpoint can be effectively fine-tuned to model not only natural language, but also visual features (i.e., a pre-trained NLP checkpoint is able to outperform a randomly initialised decoder).

Comparing BERT and GPT2, Games-Howell tests revealed a significant difference between their METEOR, ROUGE-L, and CIDEr scores (in favour of GPT2). GPT2 also attained higher label-based CE scores than BERT. This finding is opposite to that of \citet[{\footnotesize BERT}{\normalsize 2}{\footnotesize BERT} vs. {\footnotesize BERT}{\normalsize 2}{\footnotesize GPT2}]{Rothe_2020}; however, the encoder in their case was an NLP checkpoint rather than a CV checkpoint. Some of the main differences between BERT and GPT2 include their number of parameters (110M for BERT vs. 124M for GPT2), their training data (BookCorpus and English Wikipedia for BERT vs. WebText for GPT2), their vocabulary size and source (30K formed from BookCorpus and English Wikipedia for BERT vs. 50K formed from WebText for GPT2), and their token embeddings (BERT uses WordPiece \citep{Devlin_2019} while GPT2 uses byte-pair encodings \citep{Radford_2019}). However, we hypothesise that these differences are relatively minor and have only a slight effect on the performance difference. Instead, we speculate that their pre-training tasks are the main cause of the performance difference and suggest that GTP2's language modelling pre-training task is better than BERT's MLM and NSP pre-training tasks for CXR report generation.

\textbf{Answer to RQ3}: GPT2 is able to outperform BERT. Our hypothesis as to why this is the case is due to their different pre-training tasks (MLM and NSP vs. language modelling).

Amongst the domain-specific NLU checkpoints, PubMedBERT performed best. Games-Howell tests revealed a significant difference between the METEOR and ROUGE-L scores of BERT and PubMedBERT (in favour of PubMedBERT). PubMedBERT also attained higher label-based CE metric scores. This indicates that a biomedical NLU checkpoint is better than a general-domain NLU checkpoint for warm starting the decoder. Opposite to PubMedBERT, BioBERT lacked performance, with Games-Howell tests revealing a significant difference between all of the NLG metric scores of BioBERT and PubMedBERT (in favour of PubMedBERT). Additionally, PubMedBERT achieved higher label-based CE metric scores. Unlike BioBERT, PubMedBERT was pre-trained from scratch on PubMed and PMC and has a domain-specific vocabulary, both of which are cited as the reason why PubMedBERT outperforms BioBERT on biomedical NLU tasks \citep{Gu_2020}.  

For the EHR NLU checkpoints, BlueBERT was outperformed by both BERT and PubMedBERT, attaining lower mean scores for each metric (except for ROUGE-L, where BlueBERT attained a higher mean ROUGE-L score than BERT).\footnote{BlueBERT outperformed ClinicalBERT, attaining higher mean scores for each metric, with Games-Howell tests revealing a significant difference between their NLG metric scores.} Furthermore, Games-Howell tests revealed a significant difference between all the NLG metric scores of BlueBERT and PubMedBERT, and a significant difference between the METEOR and CIDEr scores of BlueBERT and BERT. Several factors could be causing the deficit in performance to the general-domain and biomedical NLU checkpoints. Unlike BERT or PubMedBERT, ClinicalBERT and BlueBERT are not pre-trained from scratch, rather, each involves three stages of pre-training, as shown in Table \ref{tab:decoders}. Moreover, MIMIC-III is significantly smaller than the datasets used to pre-train BERT and PubMedBERT (Table \ref{tab:dataset_sizes}). Finally, both ClinicalBERT and BlueBERT do not have domain-specific vocabularies. 

\textbf{Answer to RQ4 (for NLP checkpoints)}: While PubMedBERT outperformed BERT, the EHR NLU checkpoints---which are closer in domain to CXR reports---did not. This is more likely due to factors other than the choice of domain, such as the size of the pre-training datasets and the use of domain-specific vocabularies.

\begin{figure}
    \centering
    \includegraphics[scale=0.9]{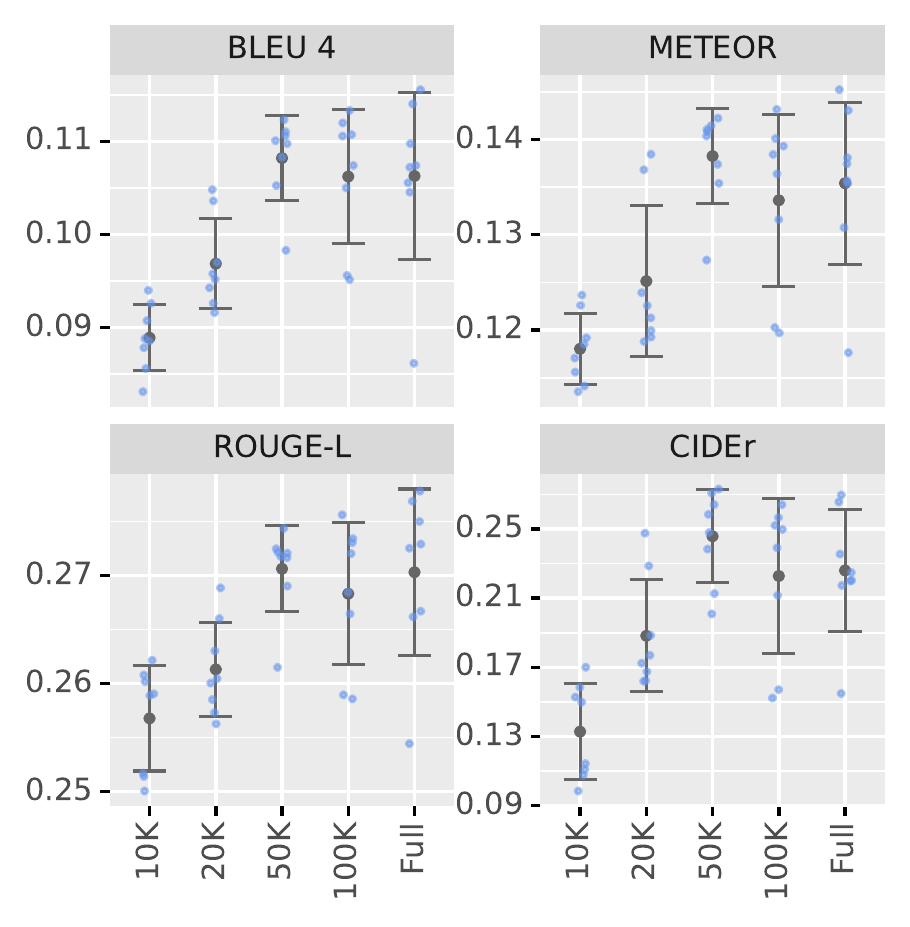}
    \caption{Different subset sizes of MIMIC-CXR's training set. {\small ResNet-101}{\large 2}{\small DistilBERT} was used for the comparison. The black dots and the error bars indicate the mean and standard deviation over the training runs, respectively.}
    \label{fig:num_training_examples}
\end{figure}

\subsection{No. of training examples} \label{sec:num_examples}

In order to make the experiments in Subsections \ref{sec:encoder_comparison} and \ref{sec:decoder_comparison} tractable, we sought to find a subset size of the MIMIC-CXR training set that was large enough that it produced similar results to the full training set, yet small enough to reduce training time. Hence, we investigate different subset sizes of the MIMIC-CXR training set, as presented in Figure \ref{fig:num_training_examples}. For each size, each training run used the same random sample (without replacement) of the MIMIC-CXR training set. We selected {\small ResNet-101}{\large 2}{\small DistilBERT} as the model. Eight training runs were conducted for each training size, where only teacher forcing was employed during fine-tuning. A one-way Welch's ANOVA revealed a significant difference between the scores of each NLG metric over the different sizes. Games-Howell tests revealed a significant difference between the scores for 10K and 20K and the full training set. This was not the case with 50K and 100K and the full training set. Hence, we use the 50K subset of MIMIC-CXR's training set in Subsections \ref{sec:encoder_comparison} and \ref{sec:decoder_comparison}, as it is the smallest tested size with no statistically significant difference to the full training set.

\section{Limitations and future recommendations} \label{sec:future}

There are several limitations of our work which lead to recommendations we propose for future investigation: 
\begin{enumerate}
    \item We did not consider every combination of encoder and decoder checkpoint, instead, we assumed that the choice of encoder does not impact the decoder and vice versa. This assumption may be wrong and could be considered for future investigation.
    \item We did not consider CXRs in DICOM format. The formats that we used for the CXRs (JPEG and PNG) meant that the pixel depth was 8-bit, rather than the 12-bits available from the DICOM format. Reducing the quantisation error could improve performance.
    \item Downsampling each CXR to $384\times 384$ increases the risk of missing fine details. Hence, the downsampled CXRs should be closer in resolution to the original CXRs.
    \item With MIMIC-CXR, reports are typically generated from only a single CXR of a study. However, studies with multiple CXRs often have multiple views. Certain abnormalities are easier to detect when interpreting both frontal and lateral views simultaneously. A CXR report generator that can accommodate a variable number of CXRs may improve performance.
    \item Radiologists frequently reference previous studies of a patient in the ground-truth reports. We hypothesise that observing both the generated reports and the visual features of previous studies will further improve CXR report generation. This will be especially important for monitoring disease progression.
    \item Motivated by Section \ref{sec:observations}, we hypothesise that an under or over sampling technique could reduce the impact of the class imbalance of the MIMIC-CXR training set.
    \item There also exists more recent CV checkpoints which are not yet open-source that may outperform CvT. For example, a ViT checkpoint was proposed that had the image patches as input replaced with the output of a small number of stacked stride-two 3$\times$3 convolutions ($\textrm{ViT}_{c}$). By injecting this inductive bias towards local spatial feature processing, $\textrm{ViT}_{c}$ is able to outperform ViT and EfficientNet \citep{Xiao_2021}.
    \item As demonstrated by CheXNet outperforming the DenseNet-121 ImageNet-21K checkpoint, a domain-specific CvT-21 checkpoint may be more apt for warm starting the encoder than the CvT-21 ImageNet-21K checkpoint. 
     \item The recent rise of Large Language Models (LLMs) has produced Transformer decoder checkpoints with considerable capabilities. An investigation into LLMs as the decoder for CXR report generation should be considered.
    \item This study indicates that the following modifications to DistilGPT2 checkpoint should be investigated: i) pre-training on PubMed and PMC from scratch before distillation, and ii) using a domain-specific vocabulary.
    \item An important consideration is the level of diagnostic accuracy that a CXR report generator must attain before retrospective or prospective clinical trials are deemed appropriate. Understanding the diagnostic accuracy of humans will help establish this, for example, a study by \citet[Table 1]{Satia_2013} found that the diagnostic accuracy of clinicians ranged from 66\% to 83\% depending on expertise. As demonstrated in Table \ref{tab:label_ce_metrics}, current CXR report generators demonstrate poor diagnostic accuracy for multiple abnormalities, indicating that further research and development is required before reaching such a threshold.
    \item We were not able to include Receiver Operating Characteristics (ROC) curves. In future work, we will consider ROC curves to provide a more thorough investigation.
    \item Metrics that evaluate the clinical decisions inferred from the generated reports should be considered---as clinical decisions ultimately dictate patient care.  
\end{enumerate}

\section{Conclusion} \label{sec:conclusion}

In this study, we investigate warm starting the encoder and decoder of a CXR report generator with recent publicly available CV and NLP checkpoints. 
Our investigation led us to the CvT-21 ImageNet-21K checkpoint---which possesses the advantages of both CNNs and Transformers---as the best CV checkpoint for warm starting the encoder. 
Moreover, we find that DistilGPT2---a distilled general-domain NLG checkpoint---is best for warm starting the decoder. 
The results indicate that the reports generated by {\small CvT-21}{\large 2}{\small DistilGPT2} are more diagnostically accurate and have a higher similarity to radiologist reports than previous approaches. 
Compared to $\mathcal{M}^2$ Transformer Progressive, {\small CvT-21}{\large 2}{\small DistilGPT2} attained an improvement of 8.3\% for CE F-1, 1.8\% for BLEU-4, 1.6\% for ROUGE-L, and 1.0\% for METEOR.

Our investigation also reveals several important findings about warm starting the encoder and decoder of a CXR report generator.
The first is that a Transformer-based CV checkpoint that incorporates convolutional layers is better than a CNN checkpoint for warm starting the encoder.
Moreover, an NLP checkpoint can be effectively fine-tuned to model not only natural language but also visual features. 
Furthermore, we find that GPT2 is better for warm starting the decoder than BERT. 
Finally, we found that domain-specific checkpoints are better for warm starting than general-domain checkpoints---if the size and quality of the pre-training dataset is sufficient.  
In general, the best checkpoint for a task depends on multiple variables; for example, the pre-training task, the size and quality of the dataset, the vocabulary, and the model architecture.

Our results indicate that leveraging warm starting improves CXR report generation. The future outlook on CXR report generation is promising; a CXR report generator that has been clinically validated through retrospective and prospective trails---that also meets regulatory requirements---could have a significant impact on radiology. 
Automatic CXR report generation could provide more consistent and reliable reporting, as well as cheaper running costs. 
It could also reduce the burden placed on overworked radiologists. 
A secondary use for such a technology could be to conduct a retrospective analysis on previous reports, which in turn could be used for other tasks such as question answering or population research. {\small CvT-21}{\large 2}{\small DistilGPT2} and its MIMIC-CXR checkpoint are available at \url{https://github.com/aehrc/cvt2distilgpt2}.

\section*{Acknowledgments}
This work was partially funded by CSIRO’s Machine Learning and Artificial Intelligence Future
Science Platform.

\bibliographystyle{cas-model2-names}
\bibliography{main.bib}

\end{document}

%% file: tex/natural_language_domain.tex
\tikzset{every picture/.style={line width=0.75pt}} 

\begin{tikzpicture}[x=0.75pt,y=0.75pt,yscale=-1,xscale=1]

\draw  [fill={rgb, 255:red, 143; green, 187; blue, 210 }  ,fill opacity=1 ] (44,158) .. controls (44,96.14) and (94.14,46) .. (156,46) .. controls (217.86,46) and (268,96.14) .. (268,158) .. controls (268,219.86) and (217.86,270) .. (156,270) .. controls (94.14,270) and (44,219.86) .. (44,158) -- cycle ;
\draw  [fill={rgb, 255:red, 149; green, 210; blue, 208 }  ,fill opacity=1 ] (64,176) .. controls (64,125.19) and (105.19,84) .. (156,84) .. controls (206.81,84) and (248,125.19) .. (248,176) .. controls (248,226.81) and (206.81,268) .. (156,268) .. controls (105.19,268) and (64,226.81) .. (64,176) -- cycle ;
\draw  [fill={rgb, 255:red, 231; green, 212; blue, 185 }  ,fill opacity=1 ] (88,197) .. controls (88,158.89) and (118.89,128) .. (157,128) .. controls (195.11,128) and (226,158.89) .. (226,197) .. controls (226,235.11) and (195.11,266) .. (157,266) .. controls (118.89,266) and (88,235.11) .. (88,197) -- cycle ;
\draw  [fill={rgb, 255:red, 220; green, 170; blue, 172 }  ,fill opacity=1 ] (111,219) .. controls (111,194.15) and (131.15,174) .. (156,174) .. controls (180.85,174) and (201,194.15) .. (201,219) .. controls (201,243.85) and (180.85,264) .. (156,264) .. controls (131.15,264) and (111,243.85) .. (111,219) -- cycle ;

\draw (101,51) node [anchor=north west][inner sep=0.75pt]   [align=left] {\begin{minipage}[lt]{77.69pt}\setlength\topsep{0pt}
\begin{center}
General\\{\scriptsize (e.g. English Wikipedia)}
\end{center}

\end{minipage}};
\draw (119,95) node [anchor=north west][inner sep=0.75pt]   [align=left] {\begin{minipage}[lt]{52.6pt}\setlength\topsep{0pt}
\begin{center}
Biomedical\\{\scriptsize (e.g. PubMed)}
\end{center}

\end{minipage}};
\draw (111,203) node [anchor=north west][inner sep=0.75pt]   [align=left] {\begin{minipage}[lt]{65pt}\setlength\topsep{0pt}
\begin{center}
CXR reports\\{\scriptsize (e.g. MIMIC-CXR)}
\end{center}

\end{minipage}};
\draw (120.03,140.95) node [anchor=north west][inner sep=0.75pt]  [align=left] {\begin{minipage}[lt]{53pt}\setlength\topsep{0pt}
\begin{center}
ICU EHRs\\{\scriptsize (e.g. MIMIC-III)}
\end{center}

\end{minipage}};

\end{tikzpicture}

%% file: tex/table_encoders.tex
\begin{table*}
\centering
\caption{CV checkpoints for warm-starting the encoder. Configurations for the same model are separated by ampersands and commas. The symbol $\to$ indicates that a checkpoint was transferred to a new domain. The configurations for ResNet, DenseNet, and CvT indicate the number of layers; the configurations for EfficientNet indicate the size of the model with B4 being the smallest and B7 being the largest; the configurations for ViT, DeiT, and BEiT indicate the patch size, where distillation is the pre-training task for DeiT; the configurations for XCiT are determined as follows: the letter and number combination before the forward slash indicates the size of the model, $\Upsilon$ indicates that distillation was used as a pre-training task, and the number between the forward slash and $\Upsilon$ indicates the width and height of the patch size.}\label{tab:encoders}
\scriptsize
\begin{tabular}{lllll}\toprule
Model &Configuration/s &Image width ($W$) &Pre-training data \\\midrule
ResNet &101 \& 152 &224 \& 224 &ImageNet-1K \\
DenseNet &121, 169, \& 201 &224, 224, \& 224 &ImageNet-1K \\
CheXNet &DenseNet-121 &224 &ImageNet-1K $\to$ CheXpert \\
EfficientNet & B4, B5, B6, \& B7 &380, 456, 528, \& 600 &ImageNet-1K \\
$\rm ViT_{BASE}$ &16x16 &384 &ImageNet-21K $\to$ ImageNet-1K \\
CvT &13 \& 21 &384 \& 384 &ImageNet-21K \\

$\rm DeiT_{BASE}$ & 16x16 with distillation & 384 & ImageNet-1K \\

XCiT  &S12/16$\Upsilon$, S12/8$\Upsilon$, S24/16$\Upsilon$, S24/8$\Upsilon$, \& M24/16$\Upsilon$ &384, 384, 384, 384, \& 384 &ImageNet-1K \\

$\rm BEiT_{BASE}$ & 16x16 & 384 & ImageNet-21K $\to$ ImageNet-1K \\

\bottomrule
\end{tabular}
\end{table*}

%% file: tex/table_decoders.tex
\begin{table*}
\centering
\caption{NLP checkpoints for warm-starting the decoder. The symbol $\to$ indicates that a checkpoint was transferred to a new domain. The Transformer is identical in configuration to DistilBERT, except that its parameters are randomly initialised rather than warm-started. PubMed and PubMed Central (PMC) are available at \url{https://pubmed.ncbi.nlm.nih.gov/} and \url{https://www.ncbi.nlm.nih.gov/pmc/}, respectively.}\label{tab:decoders}
\scriptsize
\begin{tabular}{lllll}\toprule
Model &Cased/uncased & Pre-training data &Vocabulary \\\midrule
Transformer &Uncased &– & 30K BookCorpus + English Wikipedia \\
GPT2 & Cased & WebText &50k WebText \\
$\rm BERT_{BASE}$ &Uncased &BookCorpus + English Wikipedia &30K BookCorpus + English Wikipedia \\
$\rm DistilBERT_{BASE}$ &Uncased & BookCorpus + English Wikipedia &30K BookCorpus + English Wikipedia \\
DistilGPT2 & Cased & WebText &50k WebText \\
$\rm BioBERT_{BASE}$ &Cased &$\rm BERT_{BASE}$ $\to$ PubMed + PMC &30K BookCorpus + English Wikipedia \\
$\rm SciBERT_{BASE}$ &Uncased & Semantic Scholar &30K Semantic Scholar \\
$\rm ClinicalBERT_{BASE}$  & Cased &$\rm BioBERT_{BASE}$ $\to$ MIMIC-III &30K BookCorpus + English Wikipedia \\
$\rm BlueBERT_{BASE}$ & Cased &$\rm BERT_{BASE}$ $\to$ PubMed $\to$ MIMIC-III &30K BookCorpus + English Wikipedia \\

$\rm PubMedBERT_{BASE}$ &Uncased &PubMed + PMC &30K PubMed + PMC \\

\bottomrule
\end{tabular}
\end{table*}

%% file: tex/cvt2distilgpt2.tex
\tikzset{every picture/.style={line width=0.75pt}} 

\begin{tikzpicture}[x=0.75pt,y=0.75pt,yscale=-1,xscale=1]

\draw  [color={rgb, 255:red, 0; green, 0; blue, 0 }  ,draw opacity=1 ][fill={rgb, 255:red, 212; green, 180; blue, 131 }  ,fill opacity=1 ] (290.16,425.97) .. controls (290.16,400.58) and (310.74,380) .. (336.13,380) -- (474.03,380) .. controls (499.42,380) and (520,400.58) .. (520,425.97) -- (520,784.03) .. controls (520,809.42) and (499.42,830) .. (474.03,830) -- (336.13,830) .. controls (310.74,830) and (290.16,809.42) .. (290.16,784.03) -- cycle ;
\draw  [color={rgb, 255:red, 0; green, 0; blue, 0 }  ,draw opacity=1 ][fill={rgb, 255:red, 228; green, 223; blue, 218 }  ,fill opacity=1 ] (300,432) .. controls (300,408.8) and (318.8,390) .. (342,390) -- (468,390) .. controls (491.2,390) and (510,408.8) .. (510,432) -- (510,717.59) .. controls (510,740.79) and (491.2,759.59) .. (468,759.59) -- (342,759.59) .. controls (318.8,759.59) and (300,740.79) .. (300,717.59) -- cycle ;
\draw (415,885) node  {\includegraphics[width=67.5pt,height=67.5pt]{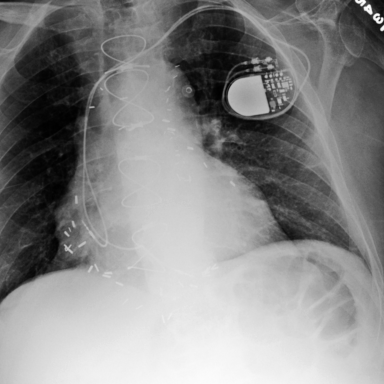}};
\draw  [draw opacity=0][line width=0.75]  (370,840) -- (460,840) -- (460,930) -- (370,930) -- cycle ; \draw  [color={rgb, 255:red, 193; green, 102; blue, 107 }  ,draw opacity=1 ][line width=0.75]  (380,840) -- (380,930)(390,840) -- (390,930)(400,840) -- (400,930)(410,840) -- (410,930)(420,840) -- (420,930)(430,840) -- (430,930)(440,840) -- (440,930)(450,840) -- (450,930) ; \draw  [color={rgb, 255:red, 193; green, 102; blue, 107 }  ,draw opacity=1 ][line width=0.75]  (370,850) -- (460,850)(370,860) -- (460,860)(370,870) -- (460,870)(370,880) -- (460,880)(370,890) -- (460,890)(370,900) -- (460,900)(370,910) -- (460,910)(370,920) -- (460,920) ; \draw  [color={rgb, 255:red, 193; green, 102; blue, 107 }  ,draw opacity=1 ][line width=0.75]   ;
\draw  [color={rgb, 255:red, 72; green, 169; blue, 166 }  ,draw opacity=1 ][fill={rgb, 255:red, 72; green, 169; blue, 166 }  ,fill opacity=0.49 ][line width=0.75]  (370,900) -- (400,900) -- (400,930) -- (370,930) -- cycle ;
\draw  [color={rgb, 255:red, 72; green, 169; blue, 166 }  ,draw opacity=1 ][fill={rgb, 255:red, 72; green, 169; blue, 166 }  ,fill opacity=0.49 ][line width=0.75]  (380,890) -- (410,890) -- (410,920) -- (380,920) -- cycle ;
\draw    (415,795) -- (415,778) ;
\draw [shift={(415,775)}, rotate = 90] [fill={rgb, 255:red, 0; green, 0; blue, 0 }  ][line width=0.08]  [draw opacity=0] (5.36,-2.57) -- (0,0) -- (5.36,2.57) -- cycle    ;
\draw    (415,840) -- (415,823) ;
\draw [shift={(415,820)}, rotate = 90] [fill={rgb, 255:red, 0; green, 0; blue, 0 }  ][line width=0.08]  [draw opacity=0] (5.36,-2.57) -- (0,0) -- (5.36,2.57) -- cycle    ;
\draw    (415,645.2) -- (372.98,616.68) ;
\draw [shift={(370.5,615)}, rotate = 34.16] [fill={rgb, 255:red, 0; green, 0; blue, 0 }  ][line width=0.08]  [draw opacity=0] (5.36,-2.57) -- (0,0) -- (5.36,2.57) -- cycle    ;
\draw  [draw opacity=0][fill={rgb, 255:red, 161; green, 136; blue, 206 }  ,fill opacity=1 ] (350.5,575) -- (390.5,575) -- (390.5,615) -- (350.5,615) -- cycle ; \draw   (355.5,575) -- (355.5,615)(360.5,575) -- (360.5,615)(365.5,575) -- (365.5,615)(370.5,575) -- (370.5,615)(375.5,575) -- (375.5,615)(380.5,575) -- (380.5,615)(385.5,575) -- (385.5,615) ; \draw   (350.5,580) -- (390.5,580)(350.5,585) -- (390.5,585)(350.5,590) -- (390.5,590)(350.5,595) -- (390.5,595)(350.5,600) -- (390.5,600)(350.5,605) -- (390.5,605)(350.5,610) -- (390.5,610) ; \draw   (350.5,575) -- (390.5,575) -- (390.5,615) -- (350.5,615) -- cycle ;
\draw  [draw opacity=0][fill={rgb, 255:red, 121; green, 151; blue, 81 }  ,fill opacity=1 ] (405.16,584.86) -- (425.16,584.86) -- (425.16,604.86) -- (405.16,604.86) -- cycle ; \draw   (410.16,584.86) -- (410.16,604.86)(415.16,584.86) -- (415.16,604.86)(420.16,584.86) -- (420.16,604.86) ; \draw   (405.16,589.86) -- (425.16,589.86)(405.16,594.86) -- (425.16,594.86)(405.16,599.86) -- (425.16,599.86) ; \draw   (405.16,584.86) -- (425.16,584.86) -- (425.16,604.86) -- (405.16,604.86) -- cycle ;
\draw  [draw opacity=0][fill={rgb, 255:red, 121; green, 151; blue, 81 }  ,fill opacity=1 ] (450,584.86) -- (470,584.86) -- (470,604.86) -- (450,604.86) -- cycle ; \draw   (455,584.86) -- (455,604.86)(460,584.86) -- (460,604.86)(465,584.86) -- (465,604.86) ; \draw   (450,589.86) -- (470,589.86)(450,594.86) -- (470,594.86)(450,599.86) -- (470,599.86) ; \draw   (450,584.86) -- (470,584.86) -- (470,604.86) -- (450,604.86) -- cycle ;
\draw    (415,645) -- (415,608) ;
\draw [shift={(415,605)}, rotate = 90] [fill={rgb, 255:red, 0; green, 0; blue, 0 }  ][line width=0.08]  [draw opacity=0] (5.36,-2.57) -- (0,0) -- (5.36,2.57) -- cycle    ;
\draw    (415,645.2) -- (457.77,606.86) ;
\draw [shift={(460,604.86)}, rotate = 138.13] [fill={rgb, 255:red, 0; green, 0; blue, 0 }  ][line width=0.08]  [draw opacity=0] (5.36,-2.57) -- (0,0) -- (5.36,2.57) -- cycle    ;
\draw  [color={rgb, 255:red, 228; green, 223; blue, 218 }  ,draw opacity=1 ][fill={rgb, 255:red, 72; green, 169; blue, 166 }  ,fill opacity=1 ] (359.8,764.4) -- (369.8,764.4) -- (369.8,774.4) -- (359.8,774.4) -- cycle ;
\draw  [color={rgb, 255:red, 228; green, 223; blue, 218 }  ,draw opacity=1 ][fill={rgb, 255:red, 72; green, 169; blue, 166 }  ,fill opacity=1 ] (369.8,764.4) -- (379.8,764.4) -- (379.8,774.4) -- (369.8,774.4) -- cycle ;
\draw  [color={rgb, 255:red, 228; green, 223; blue, 218 }  ,draw opacity=1 ][fill={rgb, 255:red, 72; green, 169; blue, 166 }  ,fill opacity=1 ] (379.8,764.4) -- (389.8,764.4) -- (389.8,774.4) -- (379.8,774.4) -- cycle ;
\draw  [color={rgb, 255:red, 228; green, 223; blue, 218 }  ,draw opacity=1 ][fill={rgb, 255:red, 72; green, 169; blue, 166 }  ,fill opacity=1 ] (399.8,764.4) -- (409.8,764.4) -- (409.8,774.4) -- (399.8,774.4) -- cycle ;
\draw  [color={rgb, 255:red, 228; green, 223; blue, 218 }  ,draw opacity=1 ][fill={rgb, 255:red, 72; green, 169; blue, 166 }  ,fill opacity=1 ] (409.8,764.4) -- (419.8,764.4) -- (419.8,774.4) -- (409.8,774.4) -- cycle ;
\draw  [color={rgb, 255:red, 228; green, 223; blue, 218 }  ,draw opacity=1 ][fill={rgb, 255:red, 72; green, 169; blue, 166 }  ,fill opacity=1 ] (419.8,764.4) -- (429.8,764.4) -- (429.8,774.4) -- (419.8,774.4) -- cycle ;
\draw  [color={rgb, 255:red, 228; green, 223; blue, 218 }  ,draw opacity=1 ][fill={rgb, 255:red, 72; green, 169; blue, 166 }  ,fill opacity=1 ] (439.8,764.4) -- (449.8,764.4) -- (449.8,774.4) -- (439.8,774.4) -- cycle ;
\draw  [color={rgb, 255:red, 228; green, 223; blue, 218 }  ,draw opacity=1 ][fill={rgb, 255:red, 72; green, 169; blue, 166 }  ,fill opacity=1 ] (449.8,764.4) -- (459.8,764.4) -- (459.8,774.4) -- (449.8,774.4) -- cycle ;
\draw  [color={rgb, 255:red, 228; green, 223; blue, 218 }  ,draw opacity=1 ][fill={rgb, 255:red, 72; green, 169; blue, 166 }  ,fill opacity=1 ] (459.8,764.4) -- (469.8,764.4) -- (469.8,774.4) -- (459.8,774.4) -- cycle ;
\draw    (414.96,764.18) -- (414.96,747.68) ;
\draw [shift={(414.96,744.68)}, rotate = 90] [fill={rgb, 255:red, 0; green, 0; blue, 0 }  ][line width=0.08]  [draw opacity=0] (5.36,-2.57) -- (0,0) -- (5.36,2.57) -- cycle    ;
\draw    (415,721) -- (415,711.5) ;
\draw [shift={(415,708.5)}, rotate = 90] [fill={rgb, 255:red, 0; green, 0; blue, 0 }  ][line width=0.08]  [draw opacity=0] (5.36,-2.57) -- (0,0) -- (5.36,2.57) -- cycle    ;
\draw  [draw opacity=0][fill={rgb, 255:red, 72; green, 169; blue, 166 }  ,fill opacity=1 ] (390.08,652.18) -- (439.08,652.18) -- (439.08,701.18) -- (390.08,701.18) -- cycle ; \draw   (397.08,652.18) -- (397.08,701.18)(404.08,652.18) -- (404.08,701.18)(411.08,652.18) -- (411.08,701.18)(418.08,652.18) -- (418.08,701.18)(425.08,652.18) -- (425.08,701.18)(432.08,652.18) -- (432.08,701.18) ; \draw   (390.08,659.18) -- (439.08,659.18)(390.08,666.18) -- (439.08,666.18)(390.08,673.18) -- (439.08,673.18)(390.08,680.18) -- (439.08,680.18)(390.08,687.18) -- (439.08,687.18)(390.08,694.18) -- (439.08,694.18) ; \draw   (390.08,652.18) -- (439.08,652.18) -- (439.08,701.18) -- (390.08,701.18) -- cycle ;
\draw    (371,575) -- (371.16,549.62) ;
\draw [shift={(371.17,546.62)}, rotate = 90.35] [fill={rgb, 255:red, 0; green, 0; blue, 0 }  ][line width=0.08]  [draw opacity=0] (5.36,-2.57) -- (0,0) -- (5.36,2.57) -- cycle    ;
\draw    (415.16,584.86) -- (415.17,549.62) ;
\draw [shift={(415.17,546.62)}, rotate = 90.02] [fill={rgb, 255:red, 0; green, 0; blue, 0 }  ][line width=0.08]  [draw opacity=0] (5.36,-2.57) -- (0,0) -- (5.36,2.57) -- cycle    ;
\draw    (460,584.86) -- (460.16,549.29) ;
\draw [shift={(460.17,546.29)}, rotate = 90.26] [fill={rgb, 255:red, 0; green, 0; blue, 0 }  ][line width=0.08]  [draw opacity=0] (5.36,-2.57) -- (0,0) -- (5.36,2.57) -- cycle    ;
\draw   (410,405) .. controls (410,402.24) and (412.24,400) .. (415,400) .. controls (417.76,400) and (420,402.24) .. (420,405) .. controls (420,407.76) and (417.76,410) .. (415,410) .. controls (412.24,410) and (410,407.76) .. (410,405) -- cycle ; \draw   (410,405) -- (420,405) ; \draw   (415,400) -- (415,410) ;
\draw    (414.92,499.48) -- (414.92,483.23) ;
\draw [shift={(414.92,480.23)}, rotate = 90] [fill={rgb, 255:red, 0; green, 0; blue, 0 }  ][line width=0.08]  [draw opacity=0] (5.36,-2.57) -- (0,0) -- (5.36,2.57) -- cycle    ;
\draw   (410,505) .. controls (410,502.24) and (412.24,500) .. (415,500) .. controls (417.76,500) and (420,502.24) .. (420,505) .. controls (420,507.76) and (417.76,510) .. (415,510) .. controls (412.24,510) and (410,507.76) .. (410,505) -- cycle ; \draw   (410,505) -- (420,505) ; \draw   (415,500) -- (415,510) ;
\draw    (320.18,549.9) .. controls (320.28,505.29) and (320.43,504.9) .. (407.27,504.65) ;
\draw [shift={(409.93,504.65)}, rotate = 179.84] [fill={rgb, 255:red, 0; green, 0; blue, 0 }  ][line width=0.08]  [draw opacity=0] (8.93,-4.29) -- (0,0) -- (8.93,4.29) -- cycle    ;
\draw    (414.95,755.07) .. controls (320.2,755.07) and (320.2,755.57) .. (320.08,710.43) ;
\draw    (320.18,549.9) -- (320.08,710.43) ;
\draw    (414.67,454.23) -- (414.67,447.98) ;
\draw [shift={(414.67,444.98)}, rotate = 90] [fill={rgb, 255:red, 0; green, 0; blue, 0 }  ][line width=0.08]  [draw opacity=0] (5.36,-2.57) -- (0,0) -- (5.36,2.57) -- cycle    ;
\draw    (414.95,493.07) .. controls (320.2,493.07) and (320.03,493.34) .. (320.18,449.9) ;
\draw    (320.18,449.9) .. controls (320.77,405.54) and (320.44,404.91) .. (407.27,404.65) ;
\draw [shift={(409.93,404.65)}, rotate = 179.84] [fill={rgb, 255:red, 0; green, 0; blue, 0 }  ][line width=0.08]  [draw opacity=0] (8.93,-4.29) -- (0,0) -- (8.93,4.29) -- cycle    ;
\draw    (414.92,419.33) -- (414.92,412.98) ;
\draw [shift={(414.92,409.98)}, rotate = 90] [fill={rgb, 255:red, 0; green, 0; blue, 0 }  ][line width=0.08]  [draw opacity=0] (5.36,-2.57) -- (0,0) -- (5.36,2.57) -- cycle    ;
\draw    (414.92,521.65) -- (414.92,513.15) ;
\draw [shift={(414.92,510.15)}, rotate = 90] [fill={rgb, 255:red, 0; green, 0; blue, 0 }  ][line width=0.08]  [draw opacity=0] (5.36,-2.57) -- (0,0) -- (5.36,2.57) -- cycle    ;
\draw  [color={rgb, 255:red, 0; green, 0; blue, 0 }  ,draw opacity=1 ][fill={rgb, 255:red, 228; green, 223; blue, 218 }  ,fill opacity=1 ] (600,491.97) .. controls (600,468.79) and (618.79,450) .. (641.97,450) -- (767.87,450) .. controls (791.05,450) and (809.84,468.79) .. (809.84,491.97) -- (809.84,747.27) .. controls (809.84,770.44) and (791.05,789.23) .. (767.87,789.23) -- (641.97,789.23) .. controls (618.79,789.23) and (600,770.44) .. (600,747.27) -- cycle ;
\draw    (695,800) -- (694.83,777.32) ;
\draw [shift={(694.8,774.32)}, rotate = 89.56] [fill={rgb, 255:red, 0; green, 0; blue, 0 }  ][line width=0.08]  [draw opacity=0] (5.36,-2.57) -- (0,0) -- (5.36,2.57) -- cycle    ;
\draw    (695,750) -- (695,733.5) ;
\draw [shift={(695,730.5)}, rotate = 90] [fill={rgb, 255:red, 0; green, 0; blue, 0 }  ][line width=0.08]  [draw opacity=0] (5.36,-2.57) -- (0,0) -- (5.36,2.57) -- cycle    ;
\draw   (689.84,459.64) .. controls (689.84,456.88) and (692.08,454.64) .. (694.84,454.64) .. controls (697.6,454.64) and (699.84,456.88) .. (699.84,459.64) .. controls (699.84,462.4) and (697.6,464.64) .. (694.84,464.64) .. controls (692.08,464.64) and (689.84,462.4) .. (689.84,459.64) -- cycle ; \draw   (689.84,459.64) -- (699.84,459.64) ; \draw   (694.84,454.64) -- (694.84,464.64) ;
\draw    (694.76,553.12) -- (694.76,536.87) ;
\draw [shift={(694.76,533.87)}, rotate = 90] [fill={rgb, 255:red, 0; green, 0; blue, 0 }  ][line width=0.08]  [draw opacity=0] (5.36,-2.57) -- (0,0) -- (5.36,2.57) -- cycle    ;
\draw   (689.84,558.64) .. controls (689.84,555.88) and (692.08,553.64) .. (694.84,553.64) .. controls (697.6,553.64) and (699.84,555.88) .. (699.84,558.64) .. controls (699.84,561.4) and (697.6,563.64) .. (694.84,563.64) .. controls (692.08,563.64) and (689.84,561.4) .. (689.84,558.64) -- cycle ; \draw   (689.84,558.64) -- (699.84,558.64) ; \draw   (694.84,553.64) -- (694.84,563.64) ;
\draw    (694.8,784.71) .. controls (784.53,784.59) and (785.03,784.59) .. (785,740) ;
\draw    (694.51,508.87) -- (694.51,502.62) ;
\draw [shift={(694.51,499.62)}, rotate = 90] [fill={rgb, 255:red, 0; green, 0; blue, 0 }  ][line width=0.08]  [draw opacity=0] (5.36,-2.57) -- (0,0) -- (5.36,2.57) -- cycle    ;
\draw    (694.76,473.97) -- (694.76,467.62) ;
\draw [shift={(694.76,464.62)}, rotate = 90] [fill={rgb, 255:red, 0; green, 0; blue, 0 }  ][line width=0.08]  [draw opacity=0] (5.36,-2.57) -- (0,0) -- (5.36,2.57) -- cycle    ;
\draw    (694.76,575.29) -- (694.76,566.79) ;
\draw [shift={(694.76,563.79)}, rotate = 90] [fill={rgb, 255:red, 0; green, 0; blue, 0 }  ][line width=0.08]  [draw opacity=0] (5.36,-2.57) -- (0,0) -- (5.36,2.57) -- cycle    ;
\draw   (690,669) .. controls (690,666.24) and (692.24,664) .. (695,664) .. controls (697.76,664) and (700,666.24) .. (700,669) .. controls (700,671.76) and (697.76,674) .. (695,674) .. controls (692.24,674) and (690,671.76) .. (690,669) -- cycle ; \draw   (690,669) -- (700,669) ; \draw   (695,664) -- (695,674) ;
\draw    (415,400) .. controls (415.2,362.5) and (415.2,360.48) .. (469.48,360.02) ;
\draw [shift={(472,360)}, rotate = 179.57] [fill={rgb, 255:red, 0; green, 0; blue, 0 }  ][line width=0.08]  [draw opacity=0] (8.93,-4.29) -- (0,0) -- (8.93,4.29) -- cycle    ;
\draw    (502,360) .. controls (560.2,360.26) and (560.45,359.76) .. (560,410) ;
\draw    (559.9,570.54) .. controls (560.06,615.27) and (560.03,614.94) .. (585,615) ;
\draw    (560,410) -- (559.9,570.54) ;
\draw [color={rgb, 255:red, 193; green, 102; blue, 107 }  ,draw opacity=1 ]   (270,410) -- (275.64,413.76) -- (300,430) ;
\draw [color={rgb, 255:red, 193; green, 102; blue, 107 }  ,draw opacity=1 ]   (180,410) -- (270,410) ;
\draw [color={rgb, 255:red, 193; green, 102; blue, 107 }  ,draw opacity=1 ]   (270,380) -- (299,399) ;
\draw [color={rgb, 255:red, 193; green, 102; blue, 107 }  ,draw opacity=1 ]   (180,380) -- (270,380) ;
\draw    (695,746) .. controls (653.49,746.56) and (650.88,746.37) .. (650.96,733.8) ;
\draw [shift={(651,731)}, rotate = 91.12] [fill={rgb, 255:red, 0; green, 0; blue, 0 }  ][line width=0.08]  [draw opacity=0] (5.36,-2.57) -- (0,0) -- (5.36,2.57) -- cycle    ;
\draw    (695,746) .. controls (736.94,745.86) and (739.88,745.63) .. (740,733.86) ;
\draw [shift={(740,731)}, rotate = 89.6] [fill={rgb, 255:red, 0; green, 0; blue, 0 }  ][line width=0.08]  [draw opacity=0] (5.36,-2.57) -- (0,0) -- (5.36,2.57) -- cycle    ;
\draw    (695,685) -- (695,676.5) ;
\draw [shift={(695,673.5)}, rotate = 90] [fill={rgb, 255:red, 0; green, 0; blue, 0 }  ][line width=0.08]  [draw opacity=0] (5.36,-2.57) -- (0,0) -- (5.36,2.57) -- cycle    ;
\draw    (785,714) .. controls (785.03,667.89) and (784.54,667.43) .. (702.51,668.95) ;
\draw [shift={(700,669)}, rotate = 358.93] [fill={rgb, 255:red, 0; green, 0; blue, 0 }  ][line width=0.08]  [draw opacity=0] (8.93,-4.29) -- (0,0) -- (8.93,4.29) -- cycle    ;
\draw    (785,710) -- (785,740) ;
\draw    (695,664.25) -- (695,648) ;
\draw [shift={(695,645)}, rotate = 90] [fill={rgb, 255:red, 0; green, 0; blue, 0 }  ][line width=0.08]  [draw opacity=0] (5.36,-2.57) -- (0,0) -- (5.36,2.57) -- cycle    ;
\draw    (585,615) .. controls (639.78,614.31) and (649.92,617.14) .. (650.91,602.93) ;
\draw [shift={(651,600)}, rotate = 89.91] [fill={rgb, 255:red, 0; green, 0; blue, 0 }  ][line width=0.08]  [draw opacity=0] (5.36,-2.57) -- (0,0) -- (5.36,2.57) -- cycle    ;
\draw    (635,615) .. controls (686.09,614.06) and (694.51,619.2) .. (694.99,602.78) ;
\draw [shift={(695,600)}, rotate = 89.2] [fill={rgb, 255:red, 0; green, 0; blue, 0 }  ][line width=0.08]  [draw opacity=0] (5.36,-2.57) -- (0,0) -- (5.36,2.57) -- cycle    ;
\draw    (694.99,622.07) .. controls (694.98,604.9) and (736.68,624.49) .. (739.82,602.92) ;
\draw [shift={(740,600)}, rotate = 89.49] [fill={rgb, 255:red, 0; green, 0; blue, 0 }  ][line width=0.08]  [draw opacity=0] (5.36,-2.57) -- (0,0) -- (5.36,2.57) -- cycle    ;
\draw    (584,615) -- (637,615) ;
\draw    (694.8,656.71) .. controls (784.53,656.59) and (785.03,656.59) .. (785,612) ;
\draw    (695,455) -- (695,438.75) ;
\draw [shift={(695,435.75)}, rotate = 90] [fill={rgb, 255:red, 0; green, 0; blue, 0 }  ][line width=0.08]  [draw opacity=0] (5.36,-2.57) -- (0,0) -- (5.36,2.57) -- cycle    ;
\draw    (695,411) -- (695,404.65) ;
\draw [shift={(695,401.65)}, rotate = 90] [fill={rgb, 255:red, 0; green, 0; blue, 0 }  ][line width=0.08]  [draw opacity=0] (5.36,-2.57) -- (0,0) -- (5.36,2.57) -- cycle    ;
\draw    (785,604) .. controls (785.03,557.89) and (784.54,557.43) .. (702.51,558.95) ;
\draw [shift={(700,559)}, rotate = 358.93] [fill={rgb, 255:red, 0; green, 0; blue, 0 }  ][line width=0.08]  [draw opacity=0] (8.93,-4.29) -- (0,0) -- (8.93,4.29) -- cycle    ;
\draw    (785,604) -- (785,622) ;
\draw    (694.8,546.71) .. controls (784.53,546.59) and (785.03,546.59) .. (785,502) ;
\draw    (785,505) .. controls (785.03,458.89) and (784.54,458.43) .. (702.51,459.95) ;
\draw [shift={(700,460)}, rotate = 358.93] [fill={rgb, 255:red, 0; green, 0; blue, 0 }  ][line width=0.08]  [draw opacity=0] (8.93,-4.29) -- (0,0) -- (8.93,4.29) -- cycle    ;
\draw   (690,805) .. controls (690,802.24) and (692.24,800) .. (695,800) .. controls (697.76,800) and (700,802.24) .. (700,805) .. controls (700,807.76) and (697.76,810) .. (695,810) .. controls (692.24,810) and (690,807.76) .. (690,805) -- cycle ; \draw   (690,805) -- (700,805) ; \draw   (695,800) -- (695,810) ;
\draw    (720,805) -- (703,805) ;
\draw [shift={(700,805)}, rotate = 360] [fill={rgb, 255:red, 0; green, 0; blue, 0 }  ][line width=0.08]  [draw opacity=0] (5.36,-2.57) -- (0,0) -- (5.36,2.57) -- cycle    ;
\draw    (695,830) -- (695,813) ;
\draw [shift={(695,810)}, rotate = 90] [fill={rgb, 255:red, 0; green, 0; blue, 0 }  ][line width=0.08]  [draw opacity=0] (5.36,-2.57) -- (0,0) -- (5.36,2.57) -- cycle    ;
\draw    (695,873) -- (695,856) ;
\draw [shift={(695,853)}, rotate = 90] [fill={rgb, 255:red, 0; green, 0; blue, 0 }  ][line width=0.08]  [draw opacity=0] (5.36,-2.57) -- (0,0) -- (5.36,2.57) -- cycle    ;
\draw [color={rgb, 255:red, 193; green, 102; blue, 107 }  ,draw opacity=1 ]   (840,470) -- (930,470) ;
\draw [color={rgb, 255:red, 193; green, 102; blue, 107 }  ,draw opacity=1 ]   (810,490) -- (840,470) ;
\draw  [draw opacity=0] (383.08,645.18) -- (446.08,645.18) -- (446.08,708.18) -- (383.08,708.18) -- cycle ; \draw   (390.08,645.18) -- (390.08,708.18)(397.08,645.18) -- (397.08,708.18)(404.08,645.18) -- (404.08,708.18)(411.08,645.18) -- (411.08,708.18)(418.08,645.18) -- (418.08,708.18)(425.08,645.18) -- (425.08,708.18)(432.08,645.18) -- (432.08,708.18)(439.08,645.18) -- (439.08,708.18) ; \draw   (383.08,652.18) -- (446.08,652.18)(383.08,659.18) -- (446.08,659.18)(383.08,666.18) -- (446.08,666.18)(383.08,673.18) -- (446.08,673.18)(383.08,680.18) -- (446.08,680.18)(383.08,687.18) -- (446.08,687.18)(383.08,694.18) -- (446.08,694.18)(383.08,701.18) -- (446.08,701.18) ; \draw   (383.08,645.18) -- (446.08,645.18) -- (446.08,708.18) -- (383.08,708.18) -- cycle ;
\draw  [color={rgb, 255:red, 121; green, 151; blue, 81 }  ,draw opacity=1 ][fill={rgb, 255:red, 121; green, 151; blue, 81 }  ,fill opacity=0.5 ][line width=0.75]  (397.08,673.18) -- (418.08,673.18) -- (418.08,694.18) -- (397.08,694.18) -- cycle ;
\draw  [color={rgb, 255:red, 121; green, 151; blue, 81 }  ,draw opacity=1 ][fill={rgb, 255:red, 121; green, 151; blue, 81 }  ,fill opacity=0.5 ][line width=0.75]  (383.08,687.18) -- (404.08,687.18) -- (404.08,708.18) -- (383.08,708.18) -- cycle ;

\draw  [fill={rgb, 255:red, 128; green, 128; blue, 128 }  ,fill opacity=1 ]  (472.34,346.99) -- (521.34,346.99) -- (521.34,371.99) -- (472.34,371.99) -- cycle  ;
\draw (496.84,359.49) node   [align=left] {{\fontfamily{phv}\selectfont \textcolor[rgb]{0.89,0.87,0.85}{Linear}}};
\draw  [fill={rgb, 255:red, 66; green, 129; blue, 164 }  ,fill opacity=1 ]  (333.15,794.99) -- (498.15,794.99) -- (498.15,819.99) -- (333.15,819.99) -- cycle  ;
\draw (415.65,807.49) node  [color={rgb, 255:red, 228; green, 223; blue, 218 }  ,opacity=1 ] [align=left] {{\fontfamily{phv}\selectfont Conv. token embedding}};
\draw (416,334) node  [font=\footnotesize] [align=left] {{\fontfamily{phv}\selectfont \textbf{{\Large CvT-21}}}};
\draw (435.28,620.74) node  [font=\scriptsize,rotate=-318.16] [align=left] {{\fontfamily{phv}\selectfont stride=2}};
\draw  [fill={rgb, 255:red, 66; green, 129; blue, 164 }  ,fill opacity=1 ]  (341.19,521.64) -- (487.19,521.64) -- (487.19,546.64) -- (341.19,546.64) -- cycle  ;
\draw (414.19,534.14) node  [color={rgb, 255:red, 228; green, 223; blue, 218 }  ,opacity=1 ] [align=left] {{\fontfamily{phv}\selectfont Multi-head attention*}};
\draw (363.09,560.81) node  [font=\scriptsize,rotate=-270] [align=left] {{\fontfamily{phv}\selectfont flatten}};
\draw (394.79,769.89) node   [align=left] {{\footnotesize ...}};
\draw (434.79,769.89) node   [align=left] {{\footnotesize ...}};
\draw (451.25,715) node  [font=\scriptsize] [align=left] {{\fontfamily{phv}\selectfont reshape \& pad}};
\draw  [fill={rgb, 255:red, 128; green, 128; blue, 128 }  ,fill opacity=1 ]  (393.33,454.49) -- (438.33,454.49) -- (438.33,479.49) -- (393.33,479.49) -- cycle  ;
\draw (415.83,466.99) node  [color={rgb, 255:red, 228; green, 223; blue, 218 }  ,opacity=1 ] [align=left] {{\fontfamily{phv}\selectfont Norm}};
\draw  [fill={rgb, 255:red, 128; green, 128; blue, 128 }  ,fill opacity=1 ]  (395.16,419.49) -- (434.16,419.49) -- (434.16,444.49) -- (395.16,444.49) -- cycle  ;
\draw (414.66,431.99) node  [color={rgb, 255:red, 228; green, 223; blue, 218 }  ,opacity=1 ] [align=left] {{\fontfamily{phv}\selectfont MLP}};
\draw  [fill={rgb, 255:red, 128; green, 128; blue, 128 }  ,fill opacity=1 ]  (395,721) -- (436,721) -- (436,744) -- (395,744) -- cycle  ;
\draw (415.5,732.5) node  [font=\small,color={rgb, 255:red, 228; green, 223; blue, 218 }  ,opacity=1 ] [align=left] {{\fontfamily{phv}\selectfont Norm}};
\draw (431.94,786.63) node  [font=\scriptsize,rotate=-359.54] [align=left] {{\fontfamily{phv}\selectfont flatten}};
\draw (453.99,567.79) node  [font=\scriptsize,rotate=-270] [align=left] {{\fontfamily{phv}\selectfont flatten}};
\draw (175,361) node [anchor=north west][inner sep=0.75pt]   [align=left] {$\displaystyle \times 3${\fontfamily{phv}\selectfont  \ Stages }};
\draw (175,391) node [anchor=north west][inner sep=0.75pt]   [align=left] {$\displaystyle \times N_{l}${\fontfamily{phv}\selectfont  \ Layers}};
\draw (408.95,568.61) node  [font=\scriptsize,rotate=-270] [align=left] {{\fontfamily{phv}\selectfont flatten}};
\draw (408.28,622.74) node  [font=\scriptsize,rotate=-270] [align=left] {{\fontfamily{phv}\selectfont stride=2}};
\draw  [fill={rgb, 255:red, 66; green, 129; blue, 164 }  ,fill opacity=1 ]  (626,685) -- (765,685) -- (765,731) -- (626,731) -- cycle  ;
\draw (695.5,708) node  [color={rgb, 255:red, 228; green, 223; blue, 218 }  ,opacity=1 ] [align=left] {\begin{minipage}[lt]{91.71pt}\setlength\topsep{0pt}

\begin{center}
{\fontfamily{phv}\selectfont Masked multi-head }
\end{center}

\begin{center}
{\fontfamily{phv}\selectfont attention}
\end{center}

\end{minipage}};
\draw  [fill={rgb, 255:red, 128; green, 128; blue, 128 }  ,fill opacity=1 ]  (674,622) -- (715,622) -- (715,645) -- (674,645) -- cycle  ;
\draw (694.5,633.5) node  [font=\small,color={rgb, 255:red, 228; green, 223; blue, 218 }  ,opacity=1 ] [align=left] {{\fontfamily{phv}\selectfont Norm}};
\draw  [fill={rgb, 255:red, 128; green, 128; blue, 128 }  ,fill opacity=1 ]  (672.84,411) -- (716.84,411) -- (716.84,436) -- (672.84,436) -- cycle  ;
\draw (694.84,423.5) node   [align=left] {\textcolor[rgb]{0.89,0.87,0.85}{{\fontfamily{phv}\selectfont Head}}};
\draw  [fill={rgb, 255:red, 66; green, 129; blue, 164 }  ,fill opacity=1 ]  (624.03,575.28) -- (764.03,575.28) -- (764.03,600.28) -- (624.03,600.28) -- cycle  ;
\draw (694.03,587.78) node  [color={rgb, 255:red, 228; green, 223; blue, 218 }  ,opacity=1 ] [align=left] {{\fontfamily{phv}\selectfont Multi-head attention}};
\draw  [fill={rgb, 255:red, 128; green, 128; blue, 128 }  ,fill opacity=1 ]  (672.17,509.13) -- (717.17,509.13) -- (717.17,534.13) -- (672.17,534.13) -- cycle  ;
\draw (694.67,521.63) node  [color={rgb, 255:red, 228; green, 223; blue, 218 }  ,opacity=1 ] [align=left] {{\fontfamily{phv}\selectfont Norm}};
\draw  [fill={rgb, 255:red, 128; green, 128; blue, 128 }  ,fill opacity=1 ]  (675,474.13) -- (714,474.13) -- (714,499.13) -- (675,499.13) -- cycle  ;
\draw (694.5,486.63) node  [color={rgb, 255:red, 228; green, 223; blue, 218 }  ,opacity=1 ] [align=left] {{\fontfamily{phv}\selectfont MLP}};
\draw  [fill={rgb, 255:red, 128; green, 128; blue, 128 }  ,fill opacity=1 ]  (674.84,750.64) -- (715.84,750.64) -- (715.84,773.64) -- (674.84,773.64) -- cycle  ;
\draw (695.34,762.14) node  [font=\small,color={rgb, 255:red, 228; green, 223; blue, 218 }  ,opacity=1 ] [align=left] {{\fontfamily{phv}\selectfont Norm}};
\draw (890.91,458.55) node   [align=left] {$\displaystyle \times 6${\fontfamily{phv}\selectfont  \ Layers}};
\draw (695.5,334) node  [font=\footnotesize] [align=left] {\textbf{{\Large {\fontfamily{phv}\selectfont DistilGPT2}}}};
\draw (767.06,805.63) node  [font=\scriptsize] [align=left] {{\fontfamily{phv}\selectfont position embeddings}};
\draw (695,879.5) node  [font=\footnotesize] [align=left] {[BOS] the patient is status post median stern|otomy};
\draw  [fill={rgb, 255:red, 128; green, 128; blue, 128 }  ,fill opacity=1 ]  (627.94,828) -- (761.95,828) -- (761.95,853) -- (627.94,853) -- cycle  ;
\draw (694.95,840.5) node   [align=left] {{\fontfamily{phv}\selectfont \textcolor[rgb]{0.89,0.87,0.85}{Token embeddings}}};
\draw (695,394.5) node  [font=\footnotesize] [align=left] {the patient is status post median stern|otomy and};
\draw (378.59,552.31) node  [font=\scriptsize] [align=left] {Q};
\draw (422.59,552.31) node  [font=\scriptsize] [align=left] {K};
\draw (469.09,552.31) node  [font=\scriptsize] [align=left] {V};
\draw (747.59,606.31) node  [font=\scriptsize] [align=left] {Q};
\draw (748.34,737.06) node  [font=\scriptsize] [align=left] {Q};
\draw (659.09,605.81) node  [font=\scriptsize] [align=left] {K};
\draw (659.09,737.13) node  [font=\scriptsize] [align=left] {K};
\draw (703.34,606.31) node  [font=\scriptsize] [align=left] {V};
\draw (703.84,737.06) node  [font=\scriptsize] [align=left] {V};

\end{tikzpicture}

%% file: tex/table_mimic_cxr.tex
\begin{table*}
\centering
\caption{Mean NLG metric scores on the MIMIC-CXR test set with the labels of \cite{Chen_2020}. If available, the 95\% confidence intervals are reported. $n=5$ indicates the mean over five training runs. * is the training run that scored the highest validation CIDEr score.}\label{tab:mimic_cxr}
\scriptsize	
\setlength{\tabcolsep}{0pt}

\begin{tabular}{w{l}{33mm} w{r}{8.5mm} >{\tiny}w{c}{1.5mm} >{\tiny}w{l}{8.5mm} w{r}{8.5mm} >{\tiny}w{c}{1.5mm} >{\tiny}w{l}{8.5mm} w{r}{8.5mm} >{\tiny}w{c}{1.5mm} >{\tiny}w{l}{8.5mm} w{r}{8.5mm} >{\tiny}w{c}{1.5mm} >{\tiny}w{l}{8.5mm} w{r}{8.5mm} >{\tiny}w{c}{1.5mm} >{\tiny}w{l}{8.5mm} w{r}{8.5mm} >{\tiny}w{c}{1.5mm} >{\tiny}w{l}{8.5mm} w{r}{8.5mm} >{\tiny}w{c}{1.5mm} >{\tiny}w{l}{8.5mm}  }

\toprule

\multirow{2}{*}{\textbf{Model}} &\multicolumn{21}{c}{\textbf{Natural language generation metrics}} \\\cmidrule{2-22}
&\multicolumn{3}{c}{\textbf{BLEU-1}} &\multicolumn{3}{c}{\textbf{BLEU-2}} &\multicolumn{3}{c}{\textbf{BLEU-3}} &\multicolumn{3}{c}{\textbf{BLEU-4}} &\multicolumn{3}{c}{\textbf{METEOR}} &\multicolumn{3}{c}{\textbf{ROUGE-L}} &\multicolumn{3}{c}{\textbf{CIDEr}} \\\midrule
R2Gen &\cellcolor[HTML]{f97d6e}0.3530 &\cellcolor[HTML]{f97d6e} &\cellcolor[HTML]{f97d6e} &\cellcolor[HTML]{f8696b}0.2180 &\cellcolor[HTML]{f8696b} &\cellcolor[HTML]{f8696b} &\cellcolor[HTML]{f8696b}0.1450 &\cellcolor[HTML]{f8696b} &\cellcolor[HTML]{f8696b} &\cellcolor[HTML]{f8696b}0.1030 &\cellcolor[HTML]{f8696b} &\cellcolor[HTML]{f8696b} &\cellcolor[HTML]{f8696b}0.1420 &\cellcolor[HTML]{f8696b} &\cellcolor[HTML]{f8696b} &\cellcolor[HTML]{fba175}0.2770 &\cellcolor[HTML]{fba175} &\cellcolor[HTML]{fba175} &- & & \\
CMN &\cellcolor[HTML]{f97d6e}0.3530 &\cellcolor[HTML]{f97d6e} &\cellcolor[HTML]{f97d6e} &\cellcolor[HTML]{f8696b}0.2180 &\cellcolor[HTML]{f8696b} &\cellcolor[HTML]{f8696b} &\cellcolor[HTML]{fa9974}0.1480 &\cellcolor[HTML]{fa9974} &\cellcolor[HTML]{fa9974} &\cellcolor[HTML]{fcb77a}0.1060 &\cellcolor[HTML]{fcb77a} &\cellcolor[HTML]{fcb77a} &\cellcolor[HTML]{f8696b}0.1420 &\cellcolor[HTML]{f8696b} &\cellcolor[HTML]{f8696b} &\cellcolor[HTML]{fbac78}0.2780 &\cellcolor[HTML]{fbac78} &\cellcolor[HTML]{fbac78} &- & & \\
Contrastive Attention &\cellcolor[HTML]{f8696b}0.3500 &\cellcolor[HTML]{f8696b} &\cellcolor[HTML]{f8696b} &\cellcolor[HTML]{f8766d}0.2190 &\cellcolor[HTML]{f8766d} &\cellcolor[HTML]{f8766d} &\cellcolor[HTML]{feda80}0.1520 &\cellcolor[HTML]{feda80} &\cellcolor[HTML]{feda80} &\cellcolor[HTML]{f7e984}0.1090 &\cellcolor[HTML]{f7e984} &\cellcolor[HTML]{f7e984} &\cellcolor[HTML]{e1e383}0.1510 &\cellcolor[HTML]{e1e383} &\cellcolor[HTML]{e1e383} &\cellcolor[HTML]{fee582}0.2830 &\cellcolor[HTML]{fee582} &\cellcolor[HTML]{fee582} &- & & \\
PPKED &\cellcolor[HTML]{fbad78}0.3600 &\cellcolor[HTML]{fbad78} &\cellcolor[HTML]{fbad78} &\cellcolor[HTML]{fcb77a}0.2240 &\cellcolor[HTML]{fcb77a} &\cellcolor[HTML]{fcb77a} &\cellcolor[HTML]{fbaa77}0.1490 &\cellcolor[HTML]{fbaa77} &\cellcolor[HTML]{fbaa77} &\cellcolor[HTML]{fcb77a}0.1060 &\cellcolor[HTML]{fcb77a} &\cellcolor[HTML]{fcb77a} &\cellcolor[HTML]{feda80}0.1490 &\cellcolor[HTML]{feda80} &\cellcolor[HTML]{feda80} &\cellcolor[HTML]{ede683}0.2840 &\cellcolor[HTML]{ede683} &\cellcolor[HTML]{ede683} &- & & \\
$\mathcal{M}^2$ Transformer Progressive &\cellcolor[HTML]{c9dc81}0.3780 &\cellcolor[HTML]{c9dc81} &\cellcolor[HTML]{c9dc81} &\cellcolor[HTML]{e2e383}0.2320 &\cellcolor[HTML]{e2e383} &\cellcolor[HTML]{e2e383} &\cellcolor[HTML]{f7e984}0.1540 &\cellcolor[HTML]{f7e984} &\cellcolor[HTML]{f7e984} &\cellcolor[HTML]{fdd17f}0.1070 &\cellcolor[HTML]{fdd17f} &\cellcolor[HTML]{fdd17f} &\cellcolor[HTML]{fa9974}0.1450 &\cellcolor[HTML]{fa9974} &\cellcolor[HTML]{fa9974} &\cellcolor[HTML]{f8696b}0.2720 &\cellcolor[HTML]{f8696b} &\cellcolor[HTML]{f8696b} &- & & \\
CvT-21{\small 2}DistilGPT2 ($n\Equal 5$) &\cellcolor[HTML]{75c47d}0.3918 &\cellcolor[HTML]{75c47d}± &\cellcolor[HTML]{75c47d}0.00008 &\cellcolor[HTML]{7ec67d}0.2454 &\cellcolor[HTML]{7ec67d}± &\cellcolor[HTML]{7ec67d}0.00008 &\cellcolor[HTML]{81c77d}0.1685 &\cellcolor[HTML]{81c77d}± &\cellcolor[HTML]{81c77d}0.00008 &\cellcolor[HTML]{81c77d}0.1236 &\cellcolor[HTML]{81c77d}± &\cellcolor[HTML]{81c77d}0.00009 &\cellcolor[HTML]{b2d580}0.1525 &\cellcolor[HTML]{b2d580}± &\cellcolor[HTML]{b2d580}0.00004 &\cellcolor[HTML]{d7e082}0.2846 &\cellcolor[HTML]{d7e082}± &\cellcolor[HTML]{d7e082}0.00007 &\cellcolor[HTML]{f8696b}0.3614 &\cellcolor[HTML]{f8696b}± &\cellcolor[HTML]{f8696b}0.00052 \\
CvT-21{\small 2}DistilGPT2* &\cellcolor[HTML]{63be7b}0.3928 &\cellcolor[HTML]{63be7b}± &\cellcolor[HTML]{63be7b}0.00013 &\cellcolor[HTML]{63be7b}0.2478 &\cellcolor[HTML]{63be7b}± &\cellcolor[HTML]{63be7b}0.00013 &\cellcolor[HTML]{63be7b}0.1713 &\cellcolor[HTML]{63be7b}± &\cellcolor[HTML]{63be7b}0.00013 &\cellcolor[HTML]{63be7b}0.1267 &\cellcolor[HTML]{63be7b}± &\cellcolor[HTML]{63be7b}0.00013 &\cellcolor[HTML]{63be7b}0.1545 &\cellcolor[HTML]{63be7b}± &\cellcolor[HTML]{63be7b}0.00007 &\cellcolor[HTML]{63be7b}0.2863 &\cellcolor[HTML]{63be7b}± &\cellcolor[HTML]{63be7b}0.00012 &\cellcolor[HTML]{63be7b}0.3892 &\cellcolor[HTML]{63be7b}± &\cellcolor[HTML]{63be7b}0.00077 \\
\bottomrule
\end{tabular}
\end{table*}

%% file: tex/table_example_ce.tex
\begin{table}
\centering
\caption{Mean example-based CE metric scores on the MIMIC-CXR test set with the labels of \cite{Chen_2020}. If available, the 95\% confidence intervals are reported. $n=5$ indicates the mean over five training runs. * is the training run that scored the highest validation CIDEr score.}\label{tab:example_ce_metrics}
\footnotesize
\setlength{\tabcolsep}{0pt}

\begin{tabular}{w{l}{33mm} w{r}{8.5mm} >{\tiny}w{c}{1.5mm} >{\tiny}w{l}{6.5mm} w{r}{8.5mm} >{\tiny}w{c}{1.5mm} >{\tiny}w{l}{6.5mm} w{r}{8.5mm} >{\tiny}w{c}{1.5mm} >{\tiny}w{l}{6.5mm}}\toprule

\multirow{2}{*}{\textbf{Model}} &\multicolumn{9}{c}{\textbf{Example-based CE metrics}} \\\cmidrule{2-10}
&\multicolumn{3}{c}{\textbf{Precision}} &\multicolumn{3}{c}{\textbf{Recall }} &\multicolumn{3}{c}{\textbf{F-1}} \\\midrule
R2Gen &\cellcolor[HTML]{fdd47f}0.3330 &\cellcolor[HTML]{fdd47f} &\cellcolor[HTML]{fdd47f} &\cellcolor[HTML]{f8696b}0.2730 &\cellcolor[HTML]{f8696b} &\cellcolor[HTML]{f8696b} &\cellcolor[HTML]{f8696b}0.2760 &\cellcolor[HTML]{f8696b} &\cellcolor[HTML]{f8696b} \\
CMN &\cellcolor[HTML]{fdd67f}0.3340 &\cellcolor[HTML]{fdd67f} &\cellcolor[HTML]{fdd67f} &\cellcolor[HTML]{f86a6b}0.2750 &\cellcolor[HTML]{f86a6b} &\cellcolor[HTML]{f86a6b} &\cellcolor[HTML]{f8716c}0.2780 &\cellcolor[HTML]{f8716c} &\cellcolor[HTML]{f8716c} \\
Contrastive Attention &\cellcolor[HTML]{ffeb84}0.3520 &\cellcolor[HTML]{ffeb84} &\cellcolor[HTML]{ffeb84} &\cellcolor[HTML]{f9806f}0.2980 &\cellcolor[HTML]{f9806f} &\cellcolor[HTML]{f9806f} &\cellcolor[HTML]{fdd680}0.3030 &\cellcolor[HTML]{fdd680} &\cellcolor[HTML]{fdd680} \\
$\mathcal{M}^2$ Transformer Prog. &\cellcolor[HTML]{f8696b}0.2400 &\cellcolor[HTML]{f8696b} &\cellcolor[HTML]{f8696b} &\cellcolor[HTML]{63be7b}0.4280 &\cellcolor[HTML]{63be7b} &\cellcolor[HTML]{63be7b} &\cellcolor[HTML]{ffeb84}0.3080 &\cellcolor[HTML]{ffeb84} &\cellcolor[HTML]{ffeb84} \\
CvT-21{\small 2}DistilGPT2 ($n\Equal 5$) &\cellcolor[HTML]{afd480}0.3597 &\cellcolor[HTML]{afd480}± &\cellcolor[HTML]{afd480}0.0003 &\cellcolor[HTML]{ffeb84}0.4122 &\cellcolor[HTML]{ffeb84}± &\cellcolor[HTML]{ffeb84}0.0003 &\cellcolor[HTML]{70c27c}0.3842 &\cellcolor[HTML]{70c27c}± &\cellcolor[HTML]{70c27c}0.0002 \\
CvT-21{\small 2}DistilGPT2* &\cellcolor[HTML]{63be7b}0.3670 &\cellcolor[HTML]{63be7b}± &\cellcolor[HTML]{63be7b}0.0004 &\cellcolor[HTML]{c2da81}0.4184 &\cellcolor[HTML]{c2da81}± &\cellcolor[HTML]{c2da81}0.0004 &\cellcolor[HTML]{63be7b}0.3910 &\cellcolor[HTML]{63be7b}± &\cellcolor[HTML]{63be7b}0.0004 \\

\bottomrule
\end{tabular}
\end{table}

%% file: tex/table_iu_x_ray.tex
\begin{table*}
\centering
\caption{Mean NLG metric scores on the IU X-Ray test set with the labels of \cite{Chen_2020}. If available, the 95\% confidence intervals are reported. $n=5$ indicates the mean over five training runs. * is the training run that scored the highest validation CIDEr score.}\label{tab:iu_x_ray}
\scriptsize
\setlength{\tabcolsep}{0pt}


\begin{tabular}{w{l}{33mm} w{r}{8.5mm} >{\tiny}w{c}{1.5mm} >{\tiny}w{l}{8.5mm} w{r}{8.5mm} >{\tiny}w{c}{1.5mm} >{\tiny}w{l}{8.5mm} w{r}{8.5mm} >{\tiny}w{c}{1.5mm} >{\tiny}w{l}{8.5mm} w{r}{8.5mm} >{\tiny}w{c}{1.5mm} >{\tiny}w{l}{8.5mm} w{r}{8.5mm} >{\tiny}w{c}{1.5mm} >{\tiny}w{l}{8.5mm} w{r}{8.5mm} >{\tiny}w{c}{1.5mm} >{\tiny}w{l}{8.5mm} w{r}{8.5mm} >{\tiny}w{c}{1.5mm} >{\tiny}w{l}{8.5mm}  }

\toprule

\multirow{2}{*}{\textbf{Model}} &\multicolumn{21}{c}{\textbf{Natural language generation metrics}} \\\cmidrule{2-22}
&\multicolumn{3}{c}{\textbf{BLEU-1}} &\multicolumn{3}{c}{\textbf{BLEU-2}} &\multicolumn{3}{c}{\textbf{BLEU-3}} &\multicolumn{3}{c}{\textbf{BLEU-4}} &\multicolumn{3}{c}{\textbf{METEOR}} &\multicolumn{3}{c}{\textbf{ROUGE-L}} &\multicolumn{3}{c}{\textbf{CIDEr}} \\\midrule
R2Gen &\cellcolor[HTML]{fbb178}0.4700 &\cellcolor[HTML]{fbb178} &\cellcolor[HTML]{fbb178} &\cellcolor[HTML]{fcc17b}0.3040 &\cellcolor[HTML]{fcc17b} &\cellcolor[HTML]{fcc17b} &\cellcolor[HTML]{fbb078}0.2190 &\cellcolor[HTML]{fbb078} &\cellcolor[HTML]{fbb078} &\cellcolor[HTML]{f8696b}0.1650 &\cellcolor[HTML]{f8696b} &\cellcolor[HTML]{f8696b} &\cellcolor[HTML]{f8696b}0.1870 &\cellcolor[HTML]{f8696b} &\cellcolor[HTML]{f8696b} &\cellcolor[HTML]{f97c6e}0.3710 &\cellcolor[HTML]{f97c6e} &\cellcolor[HTML]{f97c6e} &- & & \\
CMN &\cellcolor[HTML]{fede81}0.4750 &\cellcolor[HTML]{fede81} &\cellcolor[HTML]{fede81} &\cellcolor[HTML]{f3e884}0.3090 &\cellcolor[HTML]{f3e884} &\cellcolor[HTML]{f3e884} &\cellcolor[HTML]{fedc81}0.2220 &\cellcolor[HTML]{fedc81} &\cellcolor[HTML]{fedc81} &\cellcolor[HTML]{ede683}0.1700 &\cellcolor[HTML]{ede683} &\cellcolor[HTML]{ede683} &\cellcolor[HTML]{fdcc7e}0.1910 &\cellcolor[HTML]{fdcc7e} &\cellcolor[HTML]{fdcc7e} &\cellcolor[HTML]{ffeb84}0.3750 &\cellcolor[HTML]{ffeb84} &\cellcolor[HTML]{ffeb84} &- & & \\
Contrastive Attention &\cellcolor[HTML]{63be7b}0.4920 &\cellcolor[HTML]{63be7b} &\cellcolor[HTML]{63be7b} &\cellcolor[HTML]{7bc57d}0.3140 &\cellcolor[HTML]{7bc57d} &\cellcolor[HTML]{7bc57d} &\cellcolor[HTML]{fedc81}0.2220 &\cellcolor[HTML]{fedc81} &\cellcolor[HTML]{fedc81} &\cellcolor[HTML]{ffeb84}0.1690 &\cellcolor[HTML]{ffeb84} &\cellcolor[HTML]{ffeb84} &\cellcolor[HTML]{f4e884}0.1930 &\cellcolor[HTML]{f4e884} &\cellcolor[HTML]{f4e884} &\cellcolor[HTML]{63be7b}0.3810 &\cellcolor[HTML]{63be7b} &\cellcolor[HTML]{63be7b} &- & & \\
PPKED &\cellcolor[HTML]{bdd881}0.4830 &\cellcolor[HTML]{bdd881} &\cellcolor[HTML]{bdd881} &\cellcolor[HTML]{63be7b}0.3150 &\cellcolor[HTML]{63be7b} &\cellcolor[HTML]{63be7b} &\cellcolor[HTML]{dce182}0.2240 &\cellcolor[HTML]{dce182} &\cellcolor[HTML]{dce182} &\cellcolor[HTML]{fdca7d}0.1680 &\cellcolor[HTML]{fdca7d} &\cellcolor[HTML]{fdca7d} &\cellcolor[HTML]{fcb379}0.1900 &\cellcolor[HTML]{fcb379} &\cellcolor[HTML]{fcb379} &- & & &\cellcolor[HTML]{f8696b}0.3510 &\cellcolor[HTML]{f8696b} &\cellcolor[HTML]{f8696b} \\
$\mathcal{M}^2$ Transformer Progressive &\cellcolor[HTML]{9fd07f}0.4860 &\cellcolor[HTML]{9fd07f} &\cellcolor[HTML]{9fd07f} &\cellcolor[HTML]{63be7b}0.3150 &\cellcolor[HTML]{63be7b} &\cellcolor[HTML]{63be7b} &\cellcolor[HTML]{dce182}0.2240 &\cellcolor[HTML]{dce182} &\cellcolor[HTML]{dce182} &\cellcolor[HTML]{ffeb84}0.1690 &\cellcolor[HTML]{ffeb84} &\cellcolor[HTML]{ffeb84} &\cellcolor[HTML]{fee582}0.1920 &\cellcolor[HTML]{fee582} &\cellcolor[HTML]{fee582} &\cellcolor[HTML]{fcb379}0.3730 &\cellcolor[HTML]{fcb379} &\cellcolor[HTML]{fcb379} &- & & \\
CvT-21{\small 2}DistilGPT2 ($n\Equal 5$) &\cellcolor[HTML]{f8696b}0.4620 &\cellcolor[HTML]{f8696b}± &\cellcolor[HTML]{f8696b}0.00038 &\cellcolor[HTML]{f8696b}0.2945 &\cellcolor[HTML]{f8696b}± &\cellcolor[HTML]{f8696b}0.00030 &\cellcolor[HTML]{f8696b}0.2141 &\cellcolor[HTML]{f8696b}± &\cellcolor[HTML]{f8696b}0.00030 &\cellcolor[HTML]{f86a6b}0.1650 &\cellcolor[HTML]{f86a6b}± &\cellcolor[HTML]{f86a6b}0.00031 &\cellcolor[HTML]{fceb84}0.1924 &\cellcolor[HTML]{fceb84}± &\cellcolor[HTML]{fceb84}0.00022 &\cellcolor[HTML]{f8696b}0.3703 &\cellcolor[HTML]{f8696b}± &\cellcolor[HTML]{f8696b}0.00024 &\cellcolor[HTML]{fdd57f}0.5868 &\cellcolor[HTML]{fdd57f}± &\cellcolor[HTML]{fdd57f}0.00335 \\
CvT-21{\small 2}DistilGPT2* &\cellcolor[HTML]{fdce7e}0.4732 &\cellcolor[HTML]{fdce7e}± &\cellcolor[HTML]{fdce7e}0.00045 &\cellcolor[HTML]{fcc07b}0.3039 &\cellcolor[HTML]{fcc07b}± &\cellcolor[HTML]{fcc07b}0.00056 &\cellcolor[HTML]{63be7b}0.2242 &\cellcolor[HTML]{63be7b}± &\cellcolor[HTML]{63be7b}0.00061 &\cellcolor[HTML]{63be7b}0.1754 &\cellcolor[HTML]{63be7b}± &\cellcolor[HTML]{63be7b}0.00065 &\cellcolor[HTML]{63be7b}0.1997 &\cellcolor[HTML]{63be7b}± &\cellcolor[HTML]{63be7b}0.00029 &\cellcolor[HTML]{eae583}0.3758 &\cellcolor[HTML]{eae583}± &\cellcolor[HTML]{eae583}0.00045 &\cellcolor[HTML]{63be7b}0.6935 &\cellcolor[HTML]{63be7b}± &\cellcolor[HTML]{63be7b}0.00495 \\

\bottomrule
\end{tabular}
\end{table*}

%% file: tex/table_label_ce.tex
\begin{table}
\centering
\caption{Label-based CE metric scores of CvT-21{\small 2}DistilGPT2 for each observation on the MIMIC-CXR test set from \citet{Chen_2020}. The 95\% confidence intervals are reported for the averaged scores. The count for each observation indicates the number of times the observation was positive over all of the ground-truth reports of the training and test sets. The positive observation count for the training set is from \citet[Table 2]{Johnson_2019a}. The positive observation count for the test set of \citet{Chen_2020} was found with CheXbert \citep{Smit_2020}.}\label{tab:label_ce_metrics}
\footnotesize
\setlength{\tabcolsep}{0pt}

\begin{tabular}{w{l}{19.5mm} w{l}{9.7mm} w{l}{8mm} w{r}{7.2mm} >{\tiny}w{c}{1.5mm} >{\tiny}w{l}{6.5mm} w{r}{7.2mm} >{\tiny}w{c}{1.5mm} >{\tiny}w{l}{6.5mm} w{r}{7.2mm} >{\tiny}w{c}{1.5mm} >{\tiny}w{l}{6.5mm}}\toprule

\multirow{2}{*}{\textbf{Observation}} &\multicolumn{2}{c}{\textbf{Count}} &\multicolumn{9}{c}{\textbf{CvT-21{\small 2}DistilGPT2}} \\\cmidrule{2-12}
&\textbf{Train} &\textbf{Test} &\multicolumn{3}{c}{\textbf{Precision}} &\multicolumn{3}{c}{\textbf{Recall}} &\multicolumn{3}{c}{\textbf{F-1}} \\\midrule
No Finding &75,163 &323 &\cellcolor[HTML]{83c77d}0.681 &\cellcolor[HTML]{83c77d} &\cellcolor[HTML]{83c77d} &\cellcolor[HTML]{f98c71}0.173 &\cellcolor[HTML]{f98c71} &\cellcolor[HTML]{f98c71} &\cellcolor[HTML]{fee482}0.276 &\cellcolor[HTML]{fee482} &\cellcolor[HTML]{fee482} \\
Support Dev. &65,637 &1,345 &\cellcolor[HTML]{63be7b}0.795 &\cellcolor[HTML]{63be7b} &\cellcolor[HTML]{63be7b} &\cellcolor[HTML]{63be7b}0.734 &\cellcolor[HTML]{63be7b} &\cellcolor[HTML]{63be7b} &\cellcolor[HTML]{63be7b}0.763 &\cellcolor[HTML]{63be7b} &\cellcolor[HTML]{63be7b} \\
Pleural Effus. &53,188 &1,056 &\cellcolor[HTML]{c1d981}0.454 &\cellcolor[HTML]{c1d981} &\cellcolor[HTML]{c1d981} &\cellcolor[HTML]{76c47d}0.692 &\cellcolor[HTML]{76c47d} &\cellcolor[HTML]{76c47d} &\cellcolor[HTML]{aad380}0.548 &\cellcolor[HTML]{aad380} &\cellcolor[HTML]{aad380} \\
Lung Opacity &50,916 &1,392 &\cellcolor[HTML]{ffeb84}0.227 &\cellcolor[HTML]{ffeb84} &\cellcolor[HTML]{ffeb84} &\cellcolor[HTML]{b5d680}0.551 &\cellcolor[HTML]{b5d680} &\cellcolor[HTML]{b5d680} &\cellcolor[HTML]{f5e884}0.321 &\cellcolor[HTML]{f5e884} &\cellcolor[HTML]{f5e884} \\
Atelectasis &45,088 &841 &\cellcolor[HTML]{eae583}0.306 &\cellcolor[HTML]{eae583} &\cellcolor[HTML]{eae583} &\cellcolor[HTML]{feeb84}0.388 &\cellcolor[HTML]{feeb84} &\cellcolor[HTML]{feeb84} &\cellcolor[HTML]{eee683}0.342 &\cellcolor[HTML]{eee683} &\cellcolor[HTML]{eee683} \\
Cardiomegaly &39,094 &1,271 &\cellcolor[HTML]{b1d580}0.512 &\cellcolor[HTML]{b1d580} &\cellcolor[HTML]{b1d580} &\cellcolor[HTML]{a3d17f}0.591 &\cellcolor[HTML]{a3d17f} &\cellcolor[HTML]{a3d17f} &\cellcolor[HTML]{aad380}0.549 &\cellcolor[HTML]{aad380} &\cellcolor[HTML]{aad380} \\
Edema &26,455 &563 &\cellcolor[HTML]{feea83}0.224 &\cellcolor[HTML]{feea83} &\cellcolor[HTML]{feea83} &\cellcolor[HTML]{dae182}0.468 &\cellcolor[HTML]{dae182} &\cellcolor[HTML]{dae182} &\cellcolor[HTML]{fbea84}0.303 &\cellcolor[HTML]{fbea84} &\cellcolor[HTML]{fbea84} \\
Pneumonia &15,769 &165 &\cellcolor[HTML]{fa9e75}0.097 &\cellcolor[HTML]{fa9e75} &\cellcolor[HTML]{fa9e75} &\cellcolor[HTML]{fcc37c}0.296 &\cellcolor[HTML]{fcc37c} &\cellcolor[HTML]{fcc37c} &\cellcolor[HTML]{fba777}0.146 &\cellcolor[HTML]{fba777} &\cellcolor[HTML]{fba777} \\
Consolidation &10,487 &176 &\cellcolor[HTML]{f98a71}0.063 &\cellcolor[HTML]{f98a71} &\cellcolor[HTML]{f98a71} &\cellcolor[HTML]{fbaa77}0.239 &\cellcolor[HTML]{fbaa77} &\cellcolor[HTML]{fbaa77} &\cellcolor[HTML]{fa9172}0.099 &\cellcolor[HTML]{fa9172} &\cellcolor[HTML]{fa9172} \\
Pneumothorax &9,317 &75 &\cellcolor[HTML]{fcb479}0.133 &\cellcolor[HTML]{fcb479} &\cellcolor[HTML]{fcb479} &\cellcolor[HTML]{e0e383}0.455 &\cellcolor[HTML]{e0e383} &\cellcolor[HTML]{e0e383} &\cellcolor[HTML]{fcc37c}0.206 &\cellcolor[HTML]{fcc37c} &\cellcolor[HTML]{fcc37c} \\
Enlarged Card. &7,004 &320 &\cellcolor[HTML]{f98c71}0.066 &\cellcolor[HTML]{f98c71} &\cellcolor[HTML]{f98c71} &\cellcolor[HTML]{f8696b}0.093 &\cellcolor[HTML]{f8696b} &\cellcolor[HTML]{f8696b} &\cellcolor[HTML]{f98670}0.077 &\cellcolor[HTML]{f98670} &\cellcolor[HTML]{f98670} \\
Lung Lesion &6,129 &199 &\cellcolor[HTML]{f86a6b}0.010 &\cellcolor[HTML]{f86a6b} &\cellcolor[HTML]{f86a6b} &\cellcolor[HTML]{f98971}0.167 &\cellcolor[HTML]{f98971} &\cellcolor[HTML]{f98971} &\cellcolor[HTML]{f86b6b}0.019 &\cellcolor[HTML]{f86b6b} &\cellcolor[HTML]{f86b6b} \\
Fracture &3,768 &148 &\cellcolor[HTML]{f8696b}0.007 &\cellcolor[HTML]{f8696b} &\cellcolor[HTML]{f8696b} &\cellcolor[HTML]{fdd37f}0.333 &\cellcolor[HTML]{fdd37f} &\cellcolor[HTML]{fdd37f} &\cellcolor[HTML]{f8696b}0.013 &\cellcolor[HTML]{f8696b} &\cellcolor[HTML]{f8696b} \\
Pleural Other &1,961 &122 &\cellcolor[HTML]{f86e6c}0.016 &\cellcolor[HTML]{f86e6c} &\cellcolor[HTML]{f86e6c} &\cellcolor[HTML]{f98971}0.167 &\cellcolor[HTML]{f98971} &\cellcolor[HTML]{f98971} &\cellcolor[HTML]{f8706c}0.030 &\cellcolor[HTML]{f8706c} &\cellcolor[HTML]{f8706c} \\

\midrule

Macro-average & \multicolumn{1}{c}{-} & \multicolumn{1}{c}{-} &\cellcolor[HTML]{f7e984}0.256 &\cellcolor[HTML]{f7e984}± &\cellcolor[HTML]{f7e984}0.0013 &\cellcolor[HTML]{fee983}0.382 &\cellcolor[HTML]{fee983}± &\cellcolor[HTML]{fee983}0.0003 &\cellcolor[HTML]{faea84}0.307 &\cellcolor[HTML]{faea84}± &\cellcolor[HTML]{faea84}0.0006 \\
Micro-average &\multicolumn{1}{c}{-} & \multicolumn{1}{c}{-} &\cellcolor[HTML]{d0de82}0.398 &\cellcolor[HTML]{d0de82}± &\cellcolor[HTML]{d0de82}0.0004 &\cellcolor[HTML]{cedd82}0.497 &\cellcolor[HTML]{cedd82}± &\cellcolor[HTML]{cedd82}0.0004 &\cellcolor[HTML]{cddd82}0.442 &\cellcolor[HTML]{cddd82}± &\cellcolor[HTML]{cddd82}0.0003 \\
\bottomrule
\end{tabular}
\end{table}

%% file: tex/case_study.tex
\tikzset{every picture/.style={line width=0.75pt}} 

\begin{tikzpicture}[x=0.75pt,y=0.75pt,yscale=-1,xscale=1]

\draw (69,230) node  {\includegraphics[width=52.5pt,height=52.5pt]{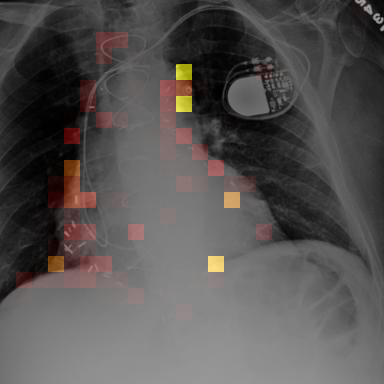}};
\draw (151,230) node  {\includegraphics[width=52.5pt,height=52.5pt]{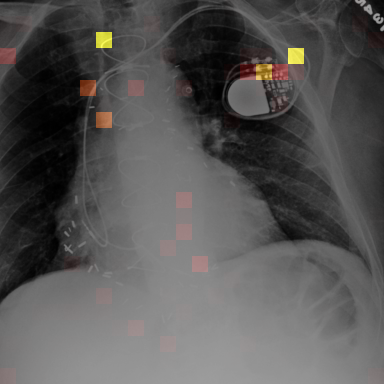}};
\draw (233,230) node  {\includegraphics[width=52.5pt,height=52.5pt]{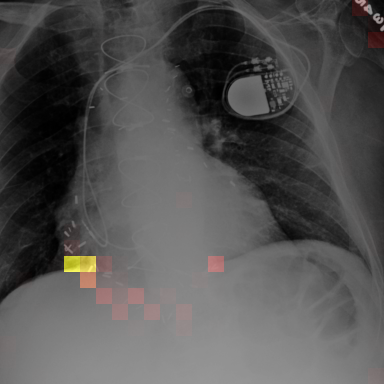}};
\draw (315,230) node  {\includegraphics[width=52.5pt,height=52.5pt]{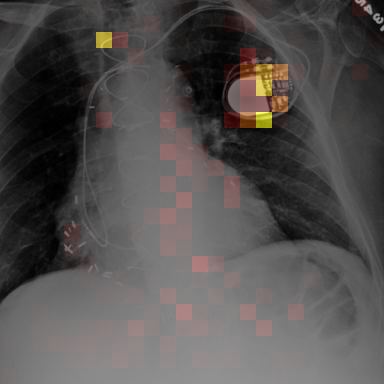}};
\draw (397,230) node  {\includegraphics[width=52.5pt,height=52.5pt]{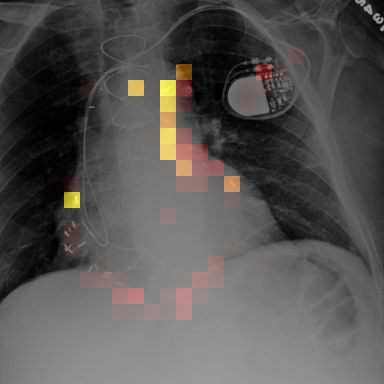}};
\draw (643,230) node  {\includegraphics[width=52.5pt,height=52.5pt]{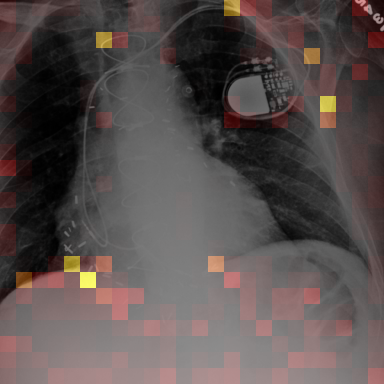}};
\draw (69,312) node  {\includegraphics[width=52.5pt,height=52.5pt]{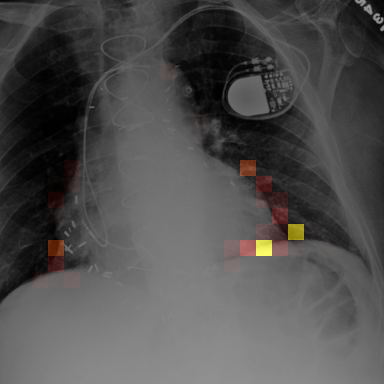}};
\draw (151,312) node  {\includegraphics[width=52.5pt,height=52.5pt]{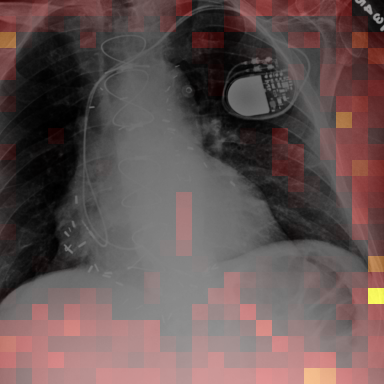}};
\draw (233,312) node  {\includegraphics[width=52.5pt,height=52.5pt]{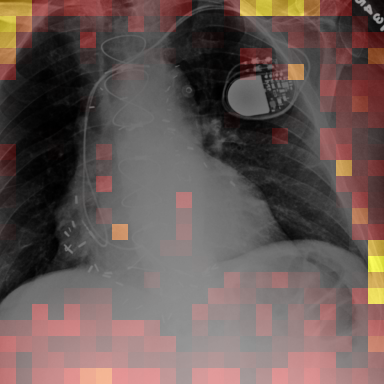}};
\draw (315,312) node  {\includegraphics[width=52.5pt,height=52.5pt]{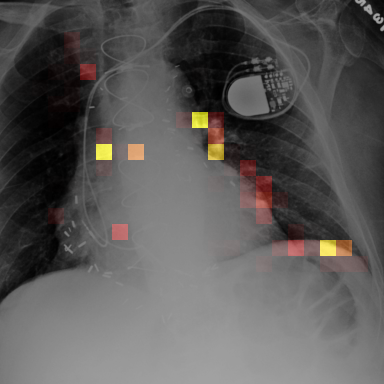}};
\draw (397,312) node  {\includegraphics[width=52.5pt,height=52.5pt]{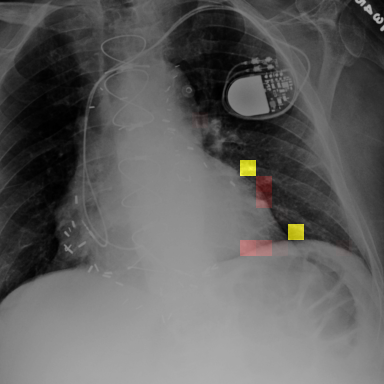}};
\draw (643,312) node  {\includegraphics[width=52.5pt,height=52.5pt]{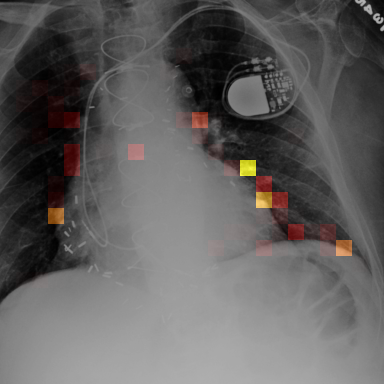}};
\draw (561,230) node  {\includegraphics[width=52.5pt,height=52.5pt]{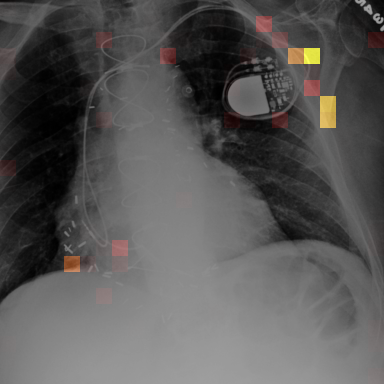}};
\draw (561,312) node  {\includegraphics[width=52.5pt,height=52.5pt]{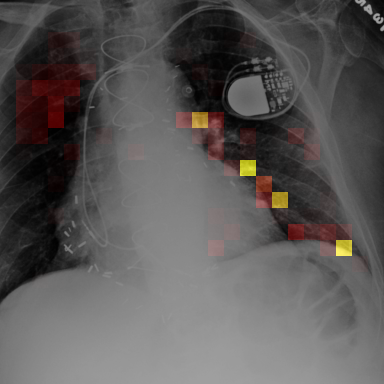}};
\draw (151,394) node  {\includegraphics[width=52.5pt,height=52.5pt]{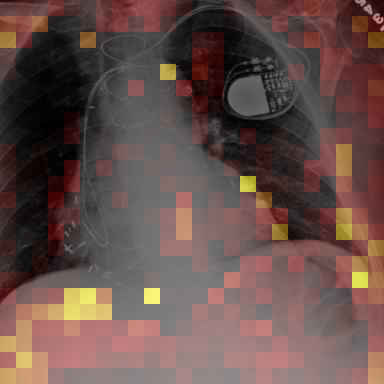}};
\draw (69,394) node  {\includegraphics[width=52.5pt,height=52.5pt]{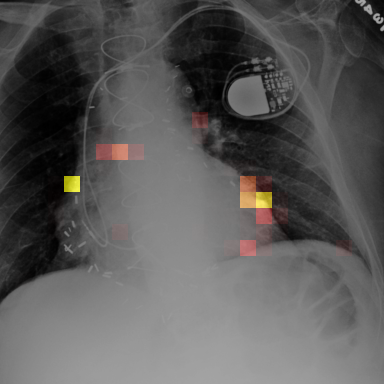}};
\draw (233,394) node  {\includegraphics[width=52.5pt,height=52.5pt]{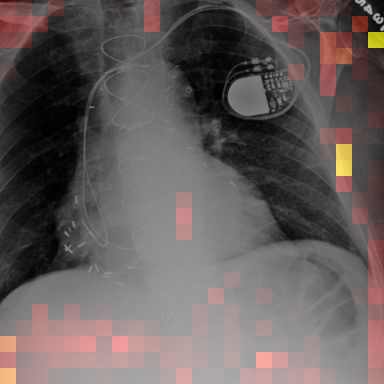}};
\draw (315,394) node  {\includegraphics[width=52.5pt,height=52.5pt]{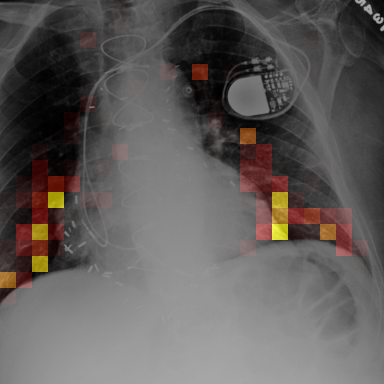}};
\draw (397,394) node  {\includegraphics[width=52.5pt,height=52.5pt]{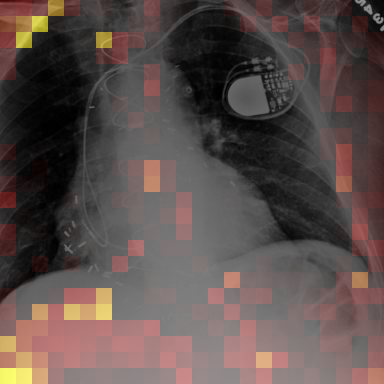}};
\draw (479,230) node  {\includegraphics[width=52.5pt,height=52.5pt]{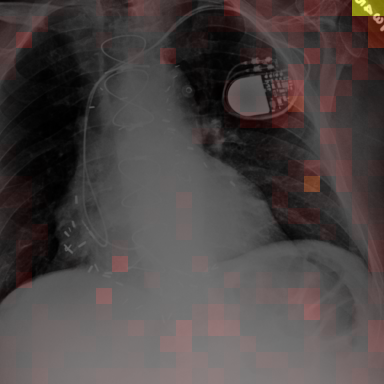}};
\draw (479,312) node  {\includegraphics[width=52.5pt,height=52.5pt]{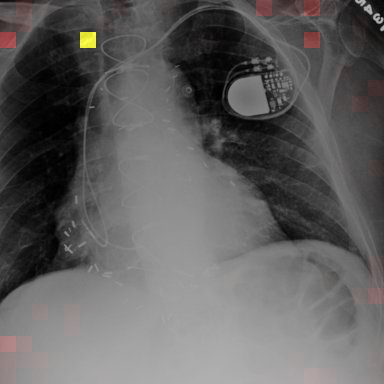}};
\draw (479,394) node  {\includegraphics[width=52.5pt,height=52.5pt]{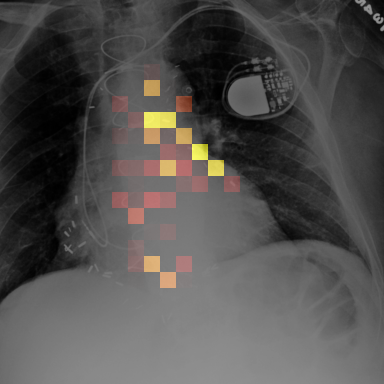}};
\draw (561,394) node  {\includegraphics[width=52.5pt,height=52.5pt]{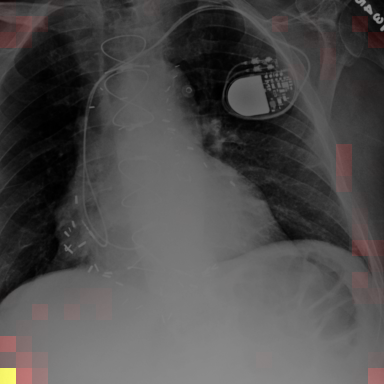}};
\draw (643,394) node  {\includegraphics[width=52.5pt,height=52.5pt]{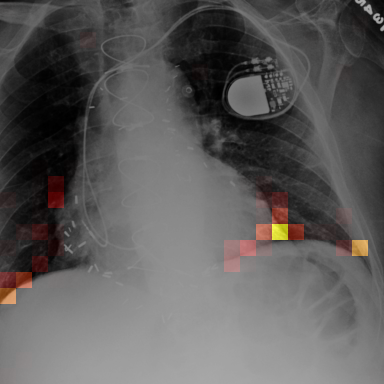}};
\draw (69,476) node  {\includegraphics[width=52.5pt,height=52.5pt]{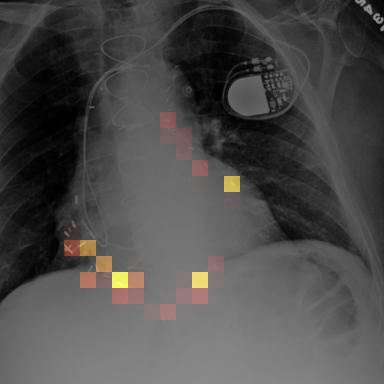}};
\draw (151,476) node  {\includegraphics[width=52.5pt,height=52.5pt]{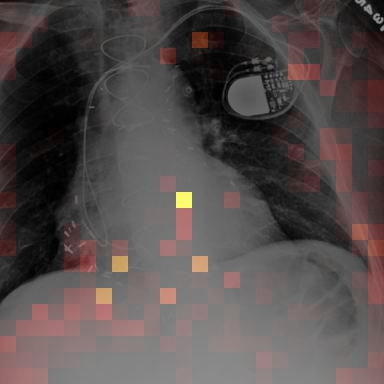}};
\draw (233,476) node  {\includegraphics[width=52.5pt,height=52.5pt]{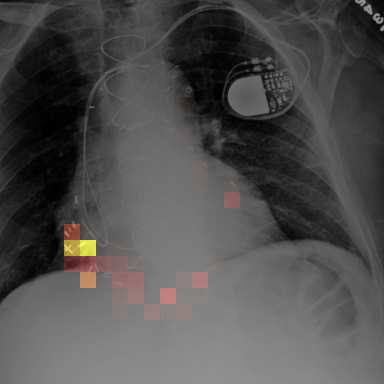}};
\draw (315,476) node  {\includegraphics[width=52.5pt,height=52.5pt]{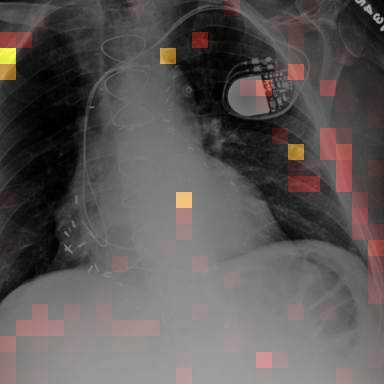}};
\draw (397,476) node  {\includegraphics[width=52.5pt,height=52.5pt]{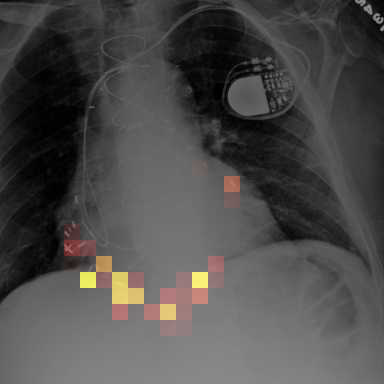}};
\draw (479,476) node  {\includegraphics[width=52.5pt,height=52.5pt]{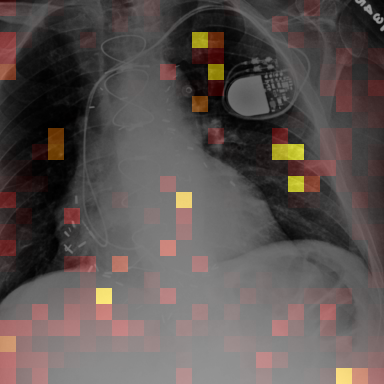}};
\draw (561,476) node  {\includegraphics[width=52.5pt,height=52.5pt]{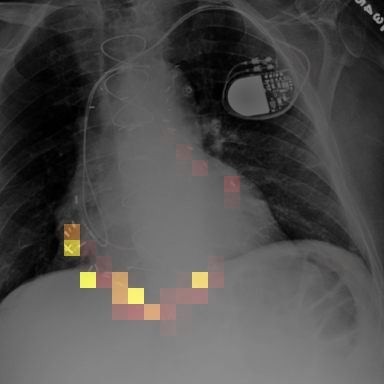}};
\draw (643,476) node  {\includegraphics[width=52.5pt,height=52.5pt]{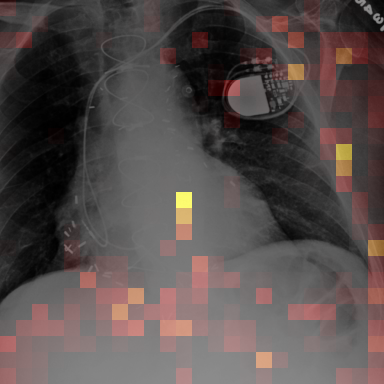}};
\draw (69,558) node  {\includegraphics[width=52.5pt,height=52.5pt]{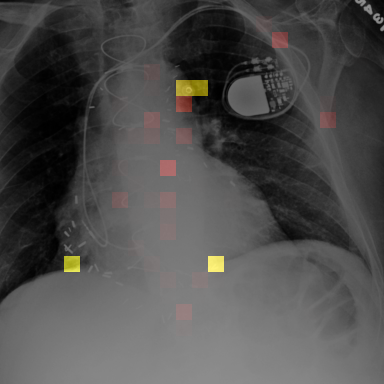}};
\draw (151,558) node  {\includegraphics[width=52.5pt,height=52.5pt]{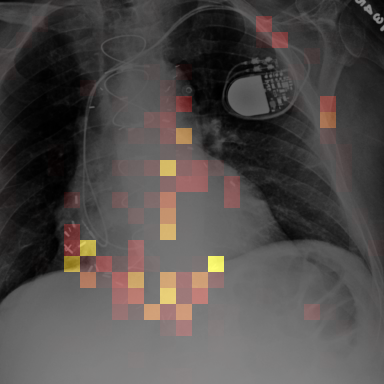}};
\draw (233,558) node  {\includegraphics[width=52.5pt,height=52.5pt]{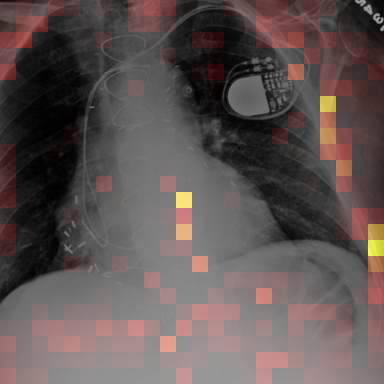}};
\draw (315,558) node  {\includegraphics[width=52.5pt,height=52.5pt]{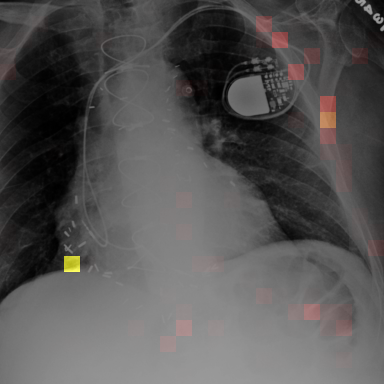}};
\draw (397,558) node  {\includegraphics[width=52.5pt,height=52.5pt]{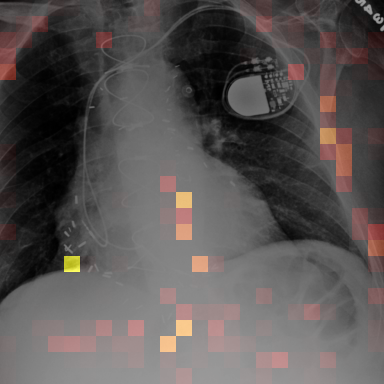}};
\draw (479,558) node  {\includegraphics[width=52.5pt,height=52.5pt]{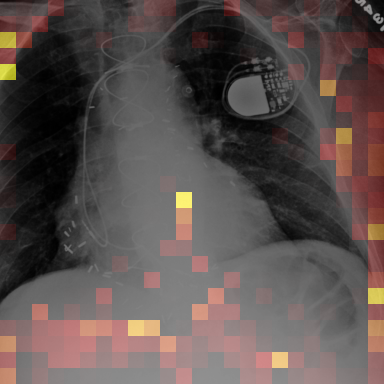}};
\draw (561,558) node  {\includegraphics[width=52.5pt,height=52.5pt]{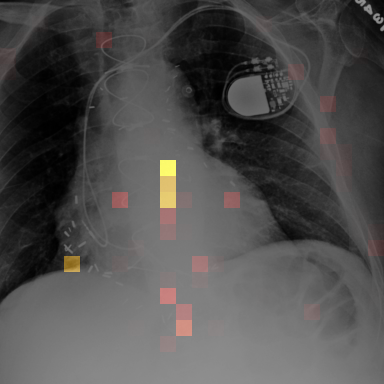}};
\draw (643,558) node  {\includegraphics[width=52.5pt,height=52.5pt]{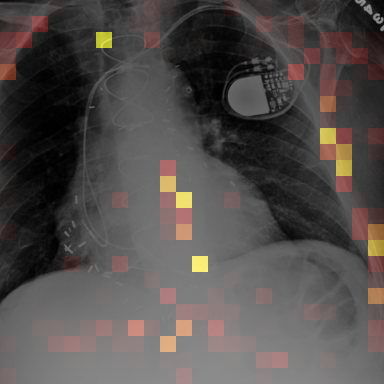}};
\draw (69,640) node  {\includegraphics[width=52.5pt,height=52.5pt]{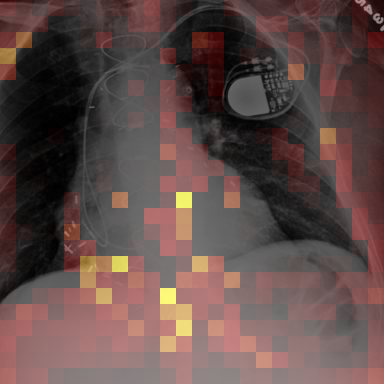}};
\draw (151,640) node  {\includegraphics[width=52.5pt,height=52.5pt]{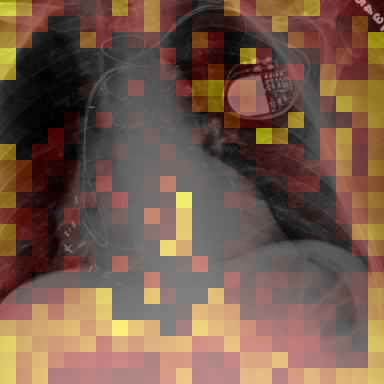}};
\draw (233,640) node  {\includegraphics[width=52.5pt,height=52.5pt]{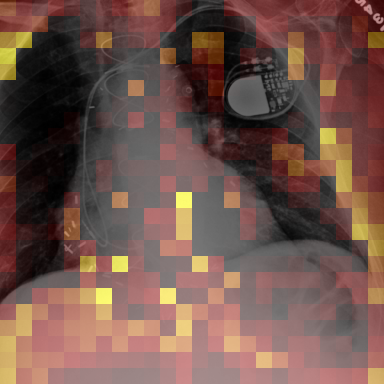}};
\draw (315,640) node  {\includegraphics[width=52.5pt,height=52.5pt]{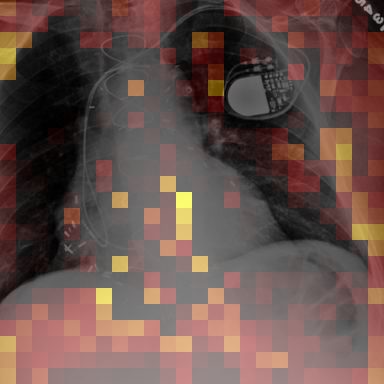}};
\draw (397,640) node  {\includegraphics[width=52.5pt,height=52.5pt]{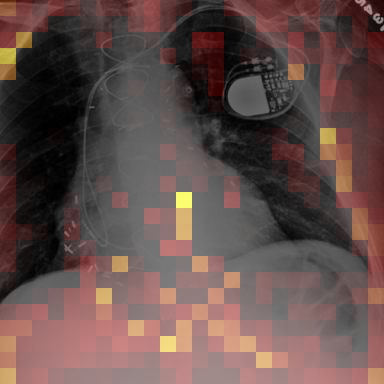}};
\draw (479,640) node  {\includegraphics[width=52.5pt,height=52.5pt]{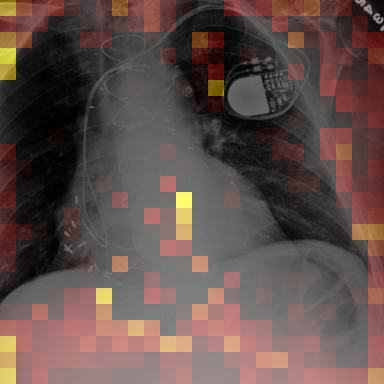}};
\draw (561,640) node  {\includegraphics[width=52.5pt,height=52.5pt]{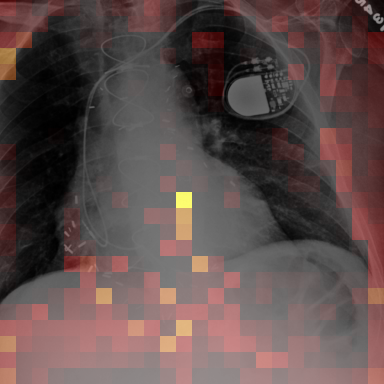}};
\draw (643,640) node  {\includegraphics[width=52.5pt,height=52.5pt]{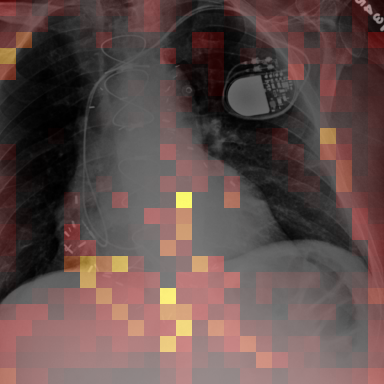}};
\draw (607,87) node  {\includegraphics[width=106.5pt,height=106.5pt]{images/cxr.png}};
\draw (356,696) node  {\includegraphics[width=483pt,height=13.5pt]{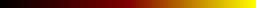}};

\draw  [draw opacity=0][fill={rgb, 255:red, 228; green, 223; blue, 218 }  ,fill opacity=1 ]  (0,0) -- (520,0) -- (520,78) -- (0,78) -- cycle  ;

\draw (3,4) node [anchor=north west][inner sep=0.75pt]   [align=left] {{\fontfamily{phv}\selectfont Ground truth}\\{\footnotesize \textcolor[rgb]{0,0,0}{patient} is status post median sternotomy and cabg. left-sided pacemaker device is noted with leads}\\{\footnotesize terminating in the right atrium and right ventricle unchanged. the heart remains mildly enlarged but }\\{\footnotesize stable. the aorta is unfolded. there is mild pulmonary vascular congestion which is improved when }\\{\footnotesize compared to the prior exam. no new focal consolidation.}};

\draw  [draw opacity=0][fill={rgb, 255:red, 228; green, 223; blue, 218 }  ,fill opacity=1 ]  (0,82.5) -- (520,82.5) -- (520,175.5) -- (0,175.5) -- cycle  ;
\draw (3.2,86.5) node [anchor=north west][inner sep=0.75pt]   [align=left] {{\fontfamily{phv}\selectfont Generated}\\{\footnotesize \textcolor[rgb]{0,0,0}{the patient} is status post median stern|otomy and \textcolor[rgb]{0.8,0.64,0.08}{\textbf{cab}}\textcolor[rgb]{0,0,0}{|}g. left|-|sided dual|-|ch|amber \textcolor[rgb]{0.84,0.13,0.18}{\textbf{pac}}\textcolor[rgb]{0,0,0}{|}emaker \ \ \ \ }\\{\footnotesize device is noted with leads terminating in the right at|\textbf{\textcolor[rgb]{0.18,0.75,0.44}{rium}} and right vent|ric|le. moderate }\\{\footnotesize enlarg|ement of the cardiac \textbf{\textcolor[rgb]{0.79,0.49,0.98}{silhouette}} is unchanged. the medi|ast|inal and \textbf{\textcolor[rgb]{0.1,0.45,0.47}{hilar}} cont|ours are }\\{\footnotesize similar. \textcolor[rgb]{0,0,0}{pulmonary} vas|\textbf{\textcolor[rgb]{0.12,0.59,0.99}{cul}}\textcolor[rgb]{0,0,0}{|}ature is not eng|or|ged. no focal consolidation ple|ural eff|\textbf{\textcolor[rgb]{0.38,0.33,0.8}{usion}} or }\\{\footnotesize \textcolor[rgb]{0.95,0.6,0.33}{\textbf{pneum}}\textcolor[rgb]{0,0,0}{|}oth|or|ax is present.}};
\draw (18.99,229.65) node  [rotate=-270] [align=left] {{\fontfamily{phv}\selectfont Layer 6 }\\{\fontfamily{phv}\selectfont Head 12}};
\draw (18.99,310.99) node  [rotate=-270] [align=left] {{\fontfamily{phv}\selectfont Layer 5}\\{\fontfamily{phv}\selectfont Head 6}};
\draw (18.99,392.99) node  [rotate=-270] [align=left] {{\fontfamily{phv}\selectfont Layer 4}\\{\fontfamily{phv}\selectfont Head 6}};
\draw (18.99,474.99) node  [rotate=-270] [align=left] {{\fontfamily{phv}\selectfont Layer 3}\\{\fontfamily{phv}\selectfont Head 7}};
\draw (18.99,557.65) node  [rotate=-270] [align=left] {{\fontfamily{phv}\selectfont Layer 2}\\{\fontfamily{phv}\selectfont Head 10}};
\draw (18.99,637.99) node  [rotate=-270] [align=left] {{\fontfamily{phv}\selectfont Layer 1}\\{\fontfamily{phv}\selectfont Head 1}};
\draw (69.87,186.49) node   [align=left] {{\fontfamily{phv}\selectfont \textcolor[rgb]{0.8,0.64,0.08}{\textbf{cab}}\textcolor[rgb]{0,0,0}{|}}};
\draw (153.5,186.5) node   [align=left] {{\fontfamily{phv}\selectfont \textcolor[rgb]{0.84,0.13,0.18}{\textbf{pac}}\textcolor[rgb]{0,0,0}{|}}};
\draw (232.2,186.49) node   [align=left] {{\fontfamily{phv}\selectfont \textcolor[rgb]{0,0,0}{|}\textcolor[rgb]{0.18,0.75,0.44}{\textbf{rium}}}};
\draw (315.41,186.49) node   [align=left] {\textcolor[rgb]{0.79,0.49,0.98}{\textbf{{\fontfamily{phv}\selectfont silhouette}}}};
\draw (642.53,186.49) node   [align=left] {{\fontfamily{phv}\selectfont \textbf{\textcolor[rgb]{0.95,0.6,0.33}{pneum}}\textcolor[rgb]{0,0,0}{|}}};
\draw (478.74,186.49) node   [align=left] {{\fontfamily{phv}\selectfont \textcolor[rgb]{0,0,0}{|}\textbf{\textcolor[rgb]{0.12,0.59,0.99}{cul}}\textcolor[rgb]{0,0,0}{|}}};
\draw (395.83,186.49) node   [align=left] {\textbf{\textcolor[rgb]{0.1,0.45,0.47}{{\fontfamily{phv}\selectfont hilar}}}};
\draw (560.94,186.49) node   [align=left] {{\fontfamily{phv}\selectfont \textcolor[rgb]{0,0,0}{|}\textbf{\textcolor[rgb]{0.38,0.33,0.8}{usion}}}};
\draw (534,9.49) node [anchor=west] [inner sep=0.75pt]   [align=left] {{\fontfamily{phv}\selectfont CXR}};
\draw (104.03,681.49) node   [align=left] {{\footnotesize {\fontfamily{phv}\selectfont Low cross-attention weight}}};
\draw (605.5,681.5) node   [align=left] {{\footnotesize {\fontfamily{phv}\selectfont High cross-attention weight}}};

\end{tikzpicture}